\documentclass[numbers]{article}

\usepackage{amsmath}
\usepackage{graphicx}
\usepackage{multirow}
\usepackage{booktabs}   
\usepackage{graphicx}   
\usepackage{colortbl}   
\usepackage[table]{xcolor} 
\usepackage{amsmath}    
\usepackage{makecell}   
\usepackage{siunitx}  
\usepackage{pifont}
\usepackage{tcolorbox}
\usepackage{tikz}
\usetikzlibrary{arrows.meta, positioning, calc}

\newcommand{\cmark}{\ding{51}}
\newcommand{\xmark}{\ding{55}}
\usepackage[preprint]{neurips_2026}
\usepackage[utf8]{inputenc} 
\usepackage[T1]{fontenc}    
\usepackage{hyperref}       
\usepackage{url}            
\usepackage{booktabs}       
\usepackage{amsfonts}       
\usepackage{nicefrac}       
\usepackage{microtype}      
\usepackage{xcolor}         
\usepackage{placeins}       
\usepackage{booktabs}
\usepackage[table]{xcolor} 
\usepackage{array}          
\usepackage{enumitem}

\title{\textsc{MiraBench}: Evaluating Action-Conditioned Reliability in Robotic World Models}

\author{%
    \normalfont
    Tianzhuo Yang$^{1}$\;\;
    Zihan Shen$^{1}$\;\;
    Zirui Mi$^{1}$\;\;
    Zhaoyi Zhang$^{1}$\;\;
    Jiayi Zhou$^{1}$ \\
    Jiaming Ji$^{1,2}$\;\;
    Juntao Dai$^{1,2}$\;\;
    Jiawei Chen$^{1}$\;\;
    Boyuan Chen$^{1,2,\dagger}$\;\;
    Yaodong Yang$^{1,\dagger}$ \\
    \vspace{-0.5em} \\
    \textnormal{$^{1}$Institute for Artificial Intelligence, Peking University}\\
    \textnormal{$^{2}$Physis Lab}
}

\begin{document}

\maketitle

\begingroup
\renewcommand{\thefootnote}{\fnsymbol{footnote}}
\footnotetext[2]{Corresponding authors: Boyuan Chen and Yaodong Yang.}
\endgroup

\begin{abstract}
Action-conditioned world models are increasingly used as scalable simulators for robot learning, yet current evaluations provide limited evidence that their predictions are reliable under the actions they condition on. Existing benchmarks largely emphasize visual fidelity, leaving unclear whether predicted futures are physically plausible, faithful to commanded actions, and calibrated to failure when actions should not succeed. We introduce \textsc{MiraBench}, a hierarchical benchmark that defines \emph{action-conditioned reliability} as a core evaluation target for robotic world models. MiraBench decomposes this target into three progressively demanding levels: \emph{Physics Adherence}, which evaluates reference-free physical consistency; \emph{Action-Following Fidelity}, which measures whether predictions respect task-relevant action inputs; and \emph{Optimism Bias Detection}, which probes the tendency to predict successful outcomes under failure-inducing actions. To support this evaluation, we curate a human-annotated corpus with over 16,000 judgments across tasks, failure categories, and leading world models. We evaluate 12 representative model configurations spanning vector-conditioned robotic world models, text-conditioned generative world models, open-weight systems, closed-source systems, and multiple model scales. Across this broad model landscape, MiraBench reveals three central findings: visual fidelity is a poor proxy for action fidelity; increasing model scale does not reliably improve action following; and optimism bias is pervasive across current systems. By shifting evaluation from appearance to action-conditioned reliability, MiraBench provides a diagnostic foundation for assessing and improving robotic world models as faithful simulators.
\end{abstract}

\section{Introduction}
\label{sec:intro}

Embodied intelligence is advancing rapidly, with autonomous robots increasingly deployed in manufacturing, healthcare, and everyday assistance~\citep{brohan2023rt2visionlanguageactionmodelstransfer, octomodelteam2024octoopensourcegeneralistrobot, embodimentcollaboration2025openxembodimentroboticlearning}.
Yet a central bottleneck remains: acquiring diverse, high-quality interaction data at scale.
Real-world data collection is expensive, slow, and difficult to generalize across environments, while rare failures, long-horizon tasks, and safety-critical interactions are especially hard to capture in sufficient quantity~\citep{khazatsky2025droidlargescaleinthewildrobot, embodimentcollaboration2025openxembodimentroboticlearning}.
This mismatch between what embodied agents must learn and what physical data collection can provide has made learned simulators increasingly attractive.

\begin{figure}[t]
  \centering
  \includegraphics[width=\linewidth]{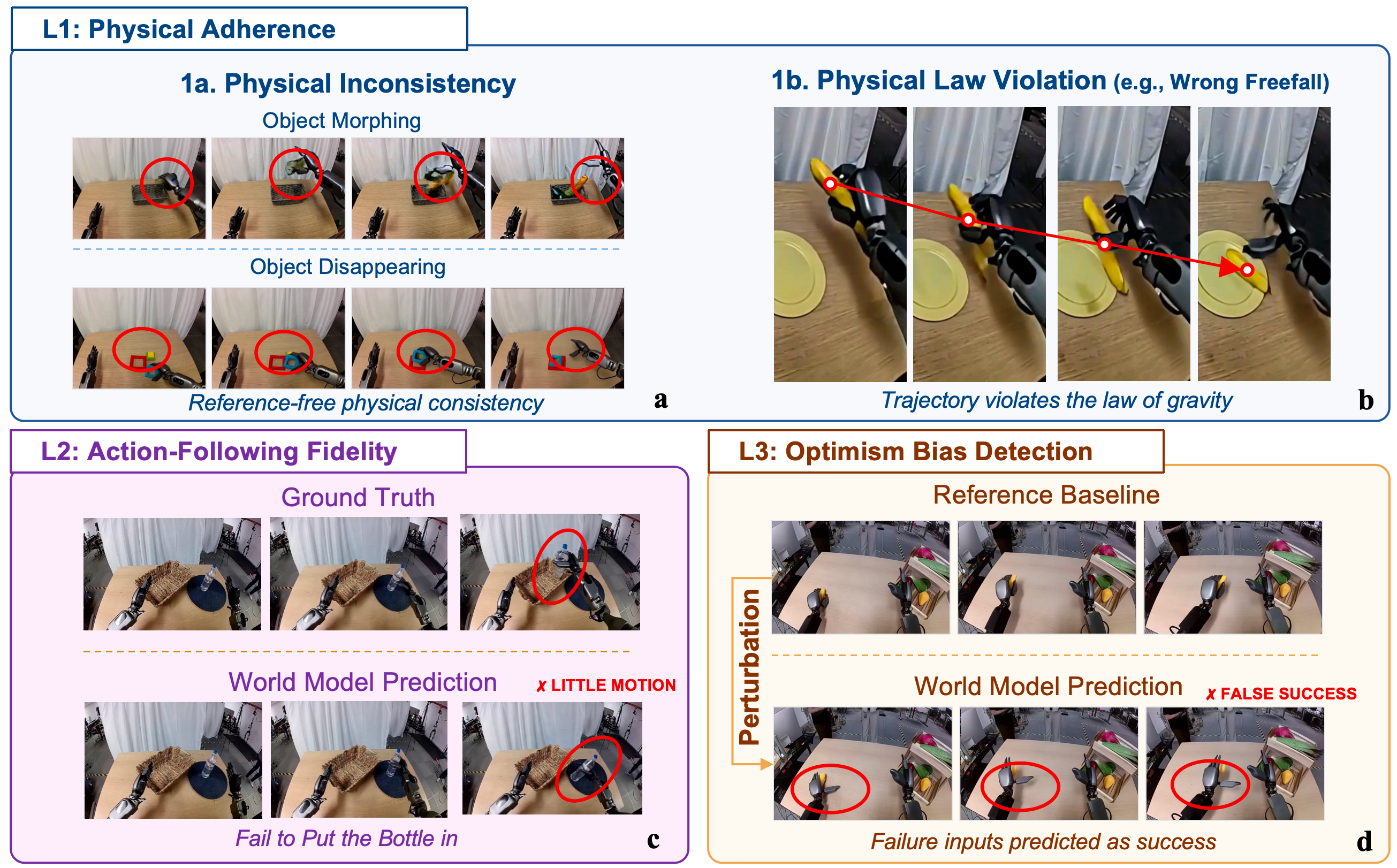}
\caption{
Representative failure modes motivating MiraBench.
(a--b) Physics Adherence failures include object morphing, disappearance, and implausible free-fall dynamics.
(c) Action-Following failures occur when predicted motion is incomplete or mismatched with the commanded action.
(d) Optimism Bias occurs when failure actions are overwritten by successful predictions.
}
  \label{fig:case}
\end{figure}

World models offer a promising solution.
By predicting future observations conditioned on actions, they can function as scalable simulators that generate trajectories beyond what can be practically collected in the real world~\citep{yang2024learninginteractiverealworldsimulators}.
This promise has led to a rapid expansion of action-conditioned world models for robotics~\citep{gao2026dreamdojogeneralistrobotworld, nvidia2025cosmosworldfoundationmodel, zhu2025irasimfinegrainedworldmodel, guo2026ctrlworldcontrollablegenerativeworld} and related domains such as driving~\citep{hu2023gaia1generativeworldmodel, gao2024vistageneralizabledrivingworld}.
However, current evaluation practice remains incomplete.

Existing benchmarks largely measure how generated videos \emph{look}, rather than whether they preserve the action-conditioned consequences required of a faithful simulator~\citep{shang2026worldarenaunifiedbenchmarkevaluating, li2025worldmodelbenchjudgingvideogeneration, qin2024worldsimbenchvideogenerationmodels, huang2023vbenchcomprehensivebenchmarksuite}.
For robot learning, this distinction is critical.
A world model can produce visually plausible futures while still violating physics, ignoring the commanded action, or overwriting a failure-inducing action with a hallucinated success.
These errors are especially plausible because robot learning datasets are dominated by successful demonstrations~\citep{mandlekar2021matterslearningofflinehuman, walke2024bridgedatav2datasetrobot, embodimentcollaboration2025openxembodimentroboticlearning}, while failures are sparse, filtered out, or entirely absent.
Therefore, a model may acquire a strong prior toward successful completion that suppresses contradictory action evidence.
We refer to this systematic failure mode as \emph{Optimism Bias} (Fig.\ref{fig:case}).

To evaluate this problem, we introduce \textit{MiraBench}, a hierarchical benchmark for \emph{action-conditioned reliability} in robotic world models.
MiraBench decomposes reliability into three nested levels: \textit{Physics Adherence}, which tests whether predicted futures remain physically coherent; \textit{Action-Following Fidelity}, which tests whether generated outcomes reflect the commanded action and task intent; and \textit{Optimism Bias Detection}, which tests whether models preserve failure outcomes when conditioned on realistic failure-inducing actions.
This hierarchy is diagnostic: lower-level failures indicate physical incoherence, while higher-level failures reveal action insensitivity or success-biased prediction.

A key component of MiraBench is its human-grounded evaluator construction.
Rather than relying on uncalibrated visual preference scores, we collect a fine-grained annotation corpus on representative world-model outputs and use it to validate module-specific VLM evaluators.
The corpus contains 906 generated videos and 16,704 structured annotation decisions across four modules: 16-indicator physical consistency, physics-law grading with anomaly tracks, structured action-following judgments, and 18-question optimism-bias diagnosis.
These annotations provide supervision for evaluators that are sensitive to object persistence, motion plausibility, occlusion behavior, task completion, failure causes, and success-overwriting cues.
The validated evaluators are then applied to a broader model suite, allowing MiraBench to combine human-level diagnostic granularity with scalable multi-model evaluation.

Using this pipeline, we evaluate 12 representative model configurations spanning vector-conditioned robotic world models and text-conditioned generative world models, including both open-weight and closed-source systems released by major industrial and academic groups such as NVIDIA, Alibaba, and Kuaishou.
Across this model landscape, we find three consistent patterns: visual fidelity is a poor proxy for action fidelity; model scale does not reliably improve action following; and optimism bias is pervasive across current systems.
Together, these findings suggest that appearance-centric evaluation substantially under-characterizes the reliability of current robotic world models as simulators.

Our contributions are as follows:
\begin{itemize}[leftmargin=*]
    \item We introduce \emph{action-conditioned reliability} as an evaluation target for robotic world models, distinguishing simulator faithfulness from generic visual fidelity.

    \item We formalize \emph{Optimism Bias}, a measurable failure mode in which world models hallucinate successful outcomes under failure-inducing actions.

    \item We present \textit{MiraBench}, a hierarchical benchmark that evaluates Physics Adherence, Action-Following Fidelity, and Optimism Bias Detection through reference-free physics checks, action-conditioned task evaluation, and targeted failure-preservation tests.

    \item We release a fine-grained human annotation corpus with 906 videos and 16,704 structured annotation decisions across four evaluation modules, providing per-indicator supervision for human-grounded VLM evaluators and reusable failure-mode analysis.

    \item We conduct a broad automated evaluation of 12 representative world-model configurations using the validated MiraBench evaluators, revealing systematic gaps between visual fidelity, nominal action-following, scale, and failure preservation.
\end{itemize}

\section{Related Work}
\label{sec:related}

\paragraph{World models for embodied AI.}
Learning to predict future states from actions has been studied from early model-based RL~\cite{sutton1990dyna, https://doi.org/10.5281/zenodo.1207631} through the DreamerV1/V2/V3 series~\cite{hafner2020dreamcontrollearningbehaviors, hafner2022masteringataridiscreteworld, hafner2024masteringdiversedomainsworld} and pixel-space models such as IRIS~\cite{micheli2023transformerssampleefficientworldmodels} and DIAMOND~\cite{alonso2024diffusionworldmodelingvisual}.
For robotic manipulation specifically, UniSim~\cite{yang2024learninginteractiverealworldsimulators}, IRASim~\cite{zhu2025irasimfinegrainedworldmodel}, CtrlWorld~\cite{guo2026ctrlworldcontrollablegenerativeworld}, and DreamDojo~\cite{gao2026dreamdojogeneralistrobotworld} demonstrate that action-conditioned video diffusion can achieve high visual fidelity on manipulation benchmarks.
Parallel developments in autonomous driving~\cite{hu2023gaia1generativeworldmodel, wang2023drivedreamerrealworlddrivenworldmodels, gao2024vistageneralizabledrivingworld} and open-world settings~\cite{bruce2024geniegenerativeinteractiveenvironments, zhou2024robodreamerlearningcompositionalworld} broaden the scope further.
Despite this progress, all of these works evaluate their models on visual quality metrics (FVD, PSNR, SSIM, or human preference) and none systematically measures whether predictions are faithful to the specific actions especially failure ones.
The near-universal exclusion of failure trajectories from training corpora~\cite{mandlekar2021matterslearningofflinehuman, walke2024bridgedatav2datasetrobot, embodimentcollaboration2025openxembodimentroboticlearning} is treated as a data collection norm rather than an evaluation problem.
MiraBench is the first benchmark to treat it as one.

\paragraph{Evaluation benchmarks for video world models.}
FVD~\cite{unterthiner2019accurategenerativemodelsvideo}, VBench~\cite{huang2023vbenchcomprehensivebenchmarksuite}, EvalCrafter~\cite{liu2024evalcrafterbenchmarkingevaluatinglarge}, and T2V-CompBench~\cite{sun2025t2vcompbenchcomprehensivebenchmarkcompositional} establish strong foundations for measuring perceptual quality and compositional text alignment.
World-model-specific benchmarks, including WorldSimBench~\cite{qin2024worldsimbenchvideogenerationmodels}, WorldModelBench~\cite{li2025worldmodelbenchjudgingvideogeneration}, WorldScore~\cite{duan2025worldscoreunifiedevaluationbenchmark}, and WorldArena~\cite{shang2026worldarenaunifiedbenchmarkevaluating}, add physics probes and instruction-following dimensions.
However, all of these frameworks evaluate what the model generates, not whether its outputs are \emph{faithful to the conditioning input}: a model that produces plausible-looking videos through a success prior scores identically to one that follows actions precisely, because the two are indistinguishable under any visual quality metric.

\paragraph{Physical reasoning in video models.}
IntPhys~\cite{riochet2020intphysframeworkbenchmarkvisual}, CLEVRER~\cite{yi2020clevrercollisioneventsvideo}, ComPhy~\cite{chen2022comphycompositionalphysicalreasoning}, Physion~\cite{bear2022physionevaluatingphysicalprediction}, Physion++~\cite{tung2023physionevaluatingphysicalscene}, and PhyWorldBench~\cite{gu2026phyworldbenchcomprehensiveevaluationphysical}collectively establish that physics violations are common in generative models and that standard metrics fail to detect them.
MiraBench's Level~1 builds on this finding, but shifts the question from abstract physics understanding to physics \emph{conditioned on a specific action}: not whether the model knows that unsupported objects fall, but whether it correctly predicts the fall when commanded to release an object.

\paragraph{Action following, policy learning, and synthetic data.}
Model-based RL methods~\cite{Schrittwieser_2020, hansen2022temporaldifferencelearningmodel, janner2021trustmodelmodelbasedpolicy} rely on world model fidelity to support planning in imagination; when fidelity is poor, policies optimized in simulation fail to transfer~\cite{janner2021trustmodelmodelbasedpolicy}.
Works on synthetic data generation for robot learning, including ROSIE~\cite{yu2023scalingrobotlearningsemantically}, RoboGen~\cite{wang2024robogenunleashinginfinitedata}, GenSim~\cite{wang2024gensimgeneratingroboticsimulation}, establish that action-state mapping accuracy matters more than visual realism for downstream policy quality.
Domain randomization~\cite{tobin2017domainrandomizationtransferringdeep} and sim-to-real transfer~\cite{openai2019learningdexterousinhandmanipulation} address fidelity gaps in physics simulators; our work identifies the analogous gap in learned world models.
WorldArena~\cite{shang2026worldarenaunifiedbenchmarkevaluating} and CtrlWorld~\cite{guo2026ctrlworldcontrollablegenerativeworld} are the closest precursors in measuring action-following in manipulation settings, but both measure average-case performance without isolating the failure regime where optimism bias is strongest.

\section{Problem Formulation}
\label{sec:motivation}


\paragraph{World model as a conditional generator.}
Let $\mathcal{W}$ denote a robotic world model and $\mathcal{E}$ the true environment dynamics.
Given an initial observation $o_0$, an action sequence $\mathbf{a}_{1:T}$:
\begin{equation}
  \hat{\mathbf{v}}_{1:T} \sim p_{\mathcal{W}}(\cdot \mid o_0, \mathbf{a}_{1:T}),
  \qquad
  \mathbf{v}^*_{1:T} \sim p_{\mathcal{E}}(\cdot \mid o_0, \mathbf{a}_{1:T}),
  \label{eq:wm_def}
\end{equation}
where $\hat{\mathbf{v}}_{1:T}$ is the predicted trajectory and $\mathbf{v}^*_{1:T}$ is the environment trajectory.
A faithful world model should preserve the consequences of the conditioning signal: its rollout should be physically admissible, semantically consistent with the specified action, and calibrated to failure when the action implies failure.
We refer to this property as \textit{action-conditioned reliability}.

\paragraph{Physical Adherence.}
Let $\Phi=\{\phi_1,\ldots,\phi_K\}$ denote a set of physical invariants that should hold for any realizable trajectory, including object persistence, coherent motion, contact causality, occlusion continuity, and simple physical laws.
Rather than treating each invariant as a binary predicate, we associate each $\phi_k$ with a normalized violation degree
$\delta_k(\hat{\mathbf{v}})\in[0,1]$, where $0$ means the invariant is satisfied and $1$ indicates a severe violation.
We define \textit{Physical Adherence} as
\begin{equation}
  \mathrm{PA}(\hat{\mathbf{v}})
  =
  1 -
  \frac{1}{K}\sum_{k=1}^{K}\delta_k(\hat{\mathbf{v}}).
  \label{eq:pa}
\end{equation}
It remains \textit{reference-free}: physical validity is assessed from the generated rollout itself, without a ground-truth video.
Thus, a prediction can appear visually realistic while still receiving low score if it violates persistence, contact dynamics, or basic kinematic constraints.

\paragraph{Optimism Bias.}
Let $\mathbf{a}^+$ denote a nominal action sequence that succeeds in the environment, and let $\mathbf{a}^-=\mathrm{Perturb}(\mathbf{a}^+,\tau,s)$ be a perturbation of type $\tau$ and severity $s$ that should induce failure under true dynamics.
Let $Y(\mathbf{v},g)\in\{0,1\}$ denote whether trajectory $\mathbf{v}$ depicts task success under goal $g$.
By construction, $Y(\mathbf{v}^{*-},g)=0$ for $\mathbf{v}^{*-}\sim p_{\mathcal{E}}(\cdot\mid o_0,\mathbf{a}^-,g)$.
We define \textit{Optimism Bias} as:
\begin{equation}
  \mathrm{OB}(\mathcal{W})
  =
  \mathbb{E}_{o_0,g,\tau,s}
  \left[
    Y(\hat{\mathbf{v}}^{-},g)
    \;\middle|\;
    \hat{\mathbf{v}}^{-}\sim p_{\mathcal{W}}(\cdot\mid o_0,\mathbf{a}^{-},g),
    \ Y(\mathbf{v}^{*-},g)=0
  \right].
  \label{eq:ob}
\end{equation}
A model with $\mathrm{OB}(\mathcal{W})=0$ preserves failure outcomes, whereas $\mathrm{OB}(\mathcal{W})=1$ always predicts success despite failure-inducing actions.
Equivalently, we report \textit{Failure Preservation} as $1-\mathrm{OB}(\mathcal{W})$, so higher values indicate better reliability.

\section{Motivating Study}
\label{sec:verification}

\noindent\textbf{The decoupling hypothesis.}
We begin with a small-scale motivating study to test whether failures in action-conditioned reliability are already observable in current world models.
Our central hypothesis is that success-dominated robot data can induce a decoupling between visual plausibility and action faithfulness.
Let $\mathcal{D}^+ = \{(o_0^{(i)}, \mathbf{a}^{+(i)}, \mathbf{v}^{*(i)})\}$ denote a training distribution consisting primarily of successful demonstrations.
Since counterfactual failures are rarely observed, a learned world model may explain future videos using a strong scene-conditioned success prior and only weak residual dependence on the action:
\begin{equation}
  p_{\mathcal{W}}(\hat{\mathbf{v}} \mid o_0, \mathbf{a})
  \propto
  p_{\text{prior}}(\hat{\mathbf{v}} \mid o_0)
  \cdot
  p_{\text{residual}}(\hat{\mathbf{v}} \mid o_0, \mathbf{a}),
  \label{eq:decouple}
\end{equation}
where $p_{\text{prior}}$ captures likely successful futures implied by the initial observation, and $p_{\text{residual}}$ captures the remaining action-conditioned variation.
Under this view, a failure-inducing action $\mathbf{a}^-$ can be overwhelmed by the success prior, causing the model to generate a plausible successful trajectory rather than the correct failure outcome. This hypothesis leads to two empirical predictions.
First, models should exhibit measurable differences in physical consistency, since physical incoherence is a lower-level failure that can mask action sensitivity.
Second, visual quality should not be sufficient evidence of reliability: a model may produce plausible videos while still predicting success under actions that should fail.

\begin{figure}[t]
  \centering
  \includegraphics[width=\linewidth]{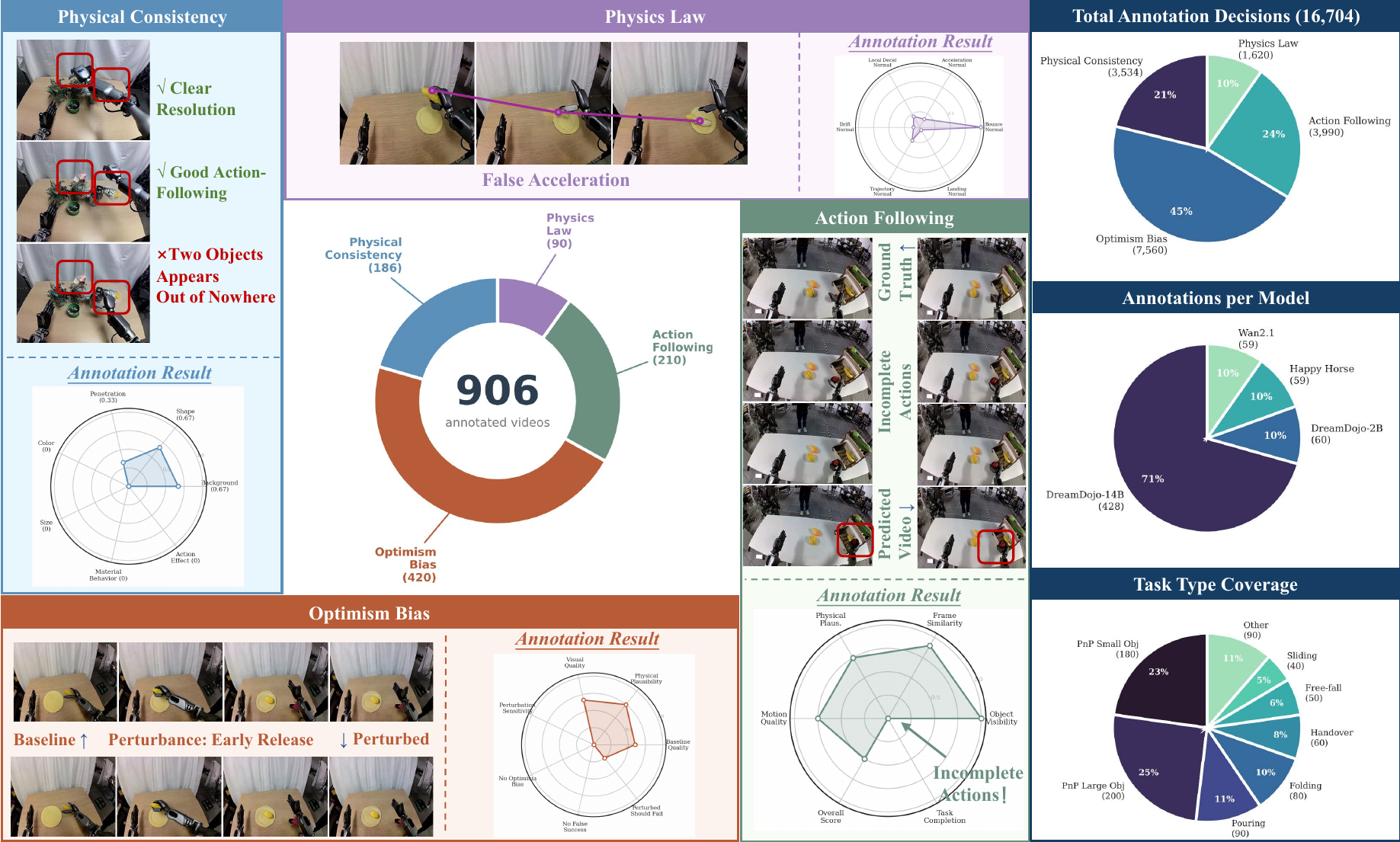}
  \caption{
    Overview of MiraBench's human annotation corpus.
    The corpus contains 906 generated videos and 16,704 structured human annotation decisions across three modules: (a) \textit{Physical Adherence}, which includes \textit{Physical Consistency} and \textit{Physics Law Compliance}, (b) \textit{Action Following}, and (c) \textit{Optimism Bias Detection}.
    These annotations are collected on representative model outputs and provide per-indicator supervision for validating MiraBench evaluators, which are then used to assess a broader set of world models.
  }
  \label{fig:dataset_composition}
\end{figure}

\noindent\textbf{Study setting.}
We test these predictions through a controlled human annotation study on four representative models: DreamDojo-2B~\citep{gao2026dreamdojogeneralistrobotworld}, DreamDojo-14B~\citep{gao2026dreamdojogeneralistrobotworld}, Wan2.1-14B~\citep{wan2025wanopenadvancedlargescale}, and Happy Horse~\citep{happyhorse2026}.
This set covers different model scales, access types, and generation regimes, allowing us to examine whether the hypothesized failures are tied to a single model family or appear across heterogeneous systems.
Trained annotators evaluated generated videos using structured rubrics rather than overall preference scores.
For physical consistency, annotators scored fine-grained dimensions such as object shape, size, material behavior, background stability, and action-effect coherence.
For optimism bias, annotators compared baseline and failure-perturbed predictions using multi-item judgments covering perturbation sensitivity, task-success prediction, false-success behavior, and failure preservation.
This design allows the pilot study to distinguish low-level physical incoherence from higher-level success-biased prediction, and to verify whether visually plausible generations can still be action-conditionally unreliable. An overview is provided in Fig.\ref{fig:dataset_composition}.
Full questionnaires, grading scales, and annotation instructions are provided in Appendix~\ref{app:annotation}.

\noindent\textbf{(a) Physical consistency varies sharply across models.}
The left panel of Table~\ref{tab:verification} reports physical-consistency pass rates across four dimensions on 30 GR-1 episodes.
Even the strongest model, Happy Horse, does not achieve perfect consistency, while DreamDojo-2B passes only 17--30\% of cases across dimensions.
The stable ranking across dimensions, Happy Horse $>$ Wan2.1 $>$ DreamDojo-14B $>$ DreamDojo-2B, suggests that these annotations capture model-level differences in physical fidelity rather than isolated artifacts.

\begin{table}[t]
\centering
\footnotesize
\setlength{\tabcolsep}{3.2pt}
\renewcommand{\arraystretch}{1.05}
\caption{
Pilot human evaluation on GR-1 episodes.
Physical consistency reports pass rates; the failure perturbation probe reports visual quality, physical plausibility, and optimism bias.
}
\label{tab:verification}
\begin{tabular*}{\linewidth}{@{\extracolsep{\fill}}lccccc ccc@{}}
\toprule
& \multicolumn{5}{c}{\textbf{Physical Consistency}} 
& \multicolumn{3}{c}{\textbf{Failure Perturbation Probe}} \\
\cmidrule(lr){2-6} \cmidrule(l){7-9}
\textbf{Model}
& \makecell{Shape\\(\%)$\uparrow$}
& \makecell{Size\\(\%)$\uparrow$}
& \makecell{Material\\(\%)$\uparrow$}
& \makecell{Background\\(\%)$\uparrow$}
& \makecell{Average\\(\%)$\uparrow$}
& \makecell{Visual\\Quality$\uparrow$}
& \makecell{Physical\\Plausibility\\(\%)$\uparrow$}
& \makecell{Optimism\\Bias\\(\%)$\uparrow$} \\
\midrule
\textit{Happy Horse} \cite{happyhorse2026}    & 90.0 & 90.0 & 93.3 & 100.0 & 93.3 & 2.24 & 96.6 & 35.7 \\
\textit{Wan2.1-14B} \cite{wan2025wanopenadvancedlargescale}     & 46.7 & 53.3 & 50.0 &  76.7 & 56.7 & 0.48 & 6.9 & 50.0 \\
\textit{DreamDojo-14B} \cite{gao2026dreamdojogeneralistrobotworld}  & 36.7 & 43.3 & 34.5 &  53.3 & 42.0 & 2.83 & 93.1 & 15.0 \\
\textit{DreamDojo-2B} \cite{gao2026dreamdojogeneralistrobotworld}   & 16.7 & 30.0 & 23.3 &  26.7 & 24.2 & 1.80 & 40.0 & 23.3 \\
\bottomrule
\end{tabular*}
\vspace{-0.1in}
\end{table}

\noindent\textbf{(b) Optimism bias is prevalent and decoupled from visual quality.}
The right panel of Table~\ref{tab:verification} reports per-model optimism bias rates under six implicit failure perturbations.
Happy Horse achieves high physical plausibility, yet still exhibits a 62.1\% bias rate, indicating that strong generation quality does not guarantee faithful failure prediction.
\textit{DreamDojo-14B} shows a similar pattern, combining relatively high visual quality with substantial optimism bias.
Conversely, \textit{Wan2.1} has a lower detected bias rate, but this should not be interpreted as reliability: its low visual quality and physical plausibility limit its ability to produce recognizable success predictions.
These results support the need to evaluate action-conditioned reliability separately from visual realism.

\noindent\textbf{Implications for benchmark design.}
The motivating study suggests three design requirements for MiraBench.
First, physical consistency should be evaluated reference-free, since counterfactual actions often lack ground-truth videos.
Second, optimism bias should be probed with failure-inducing perturbations whose correct outcomes are known under basic task physics, rather than with perturbations that merely degrade visual quality.
Third, evaluation should be hierarchical: models with poor physical coherence cannot be meaningfully diagnosed for action sensitivity or optimism bias without first localizing the lower-level failure.

\section{Design of \textsc{MIRABENCH}}
\label{sec:framework}


\subsection{Construction of Test Cases}
\label{sec:benchmark:datagen}

\begin{figure*}[t]
  \centering
  \includegraphics[width=0.9\linewidth]{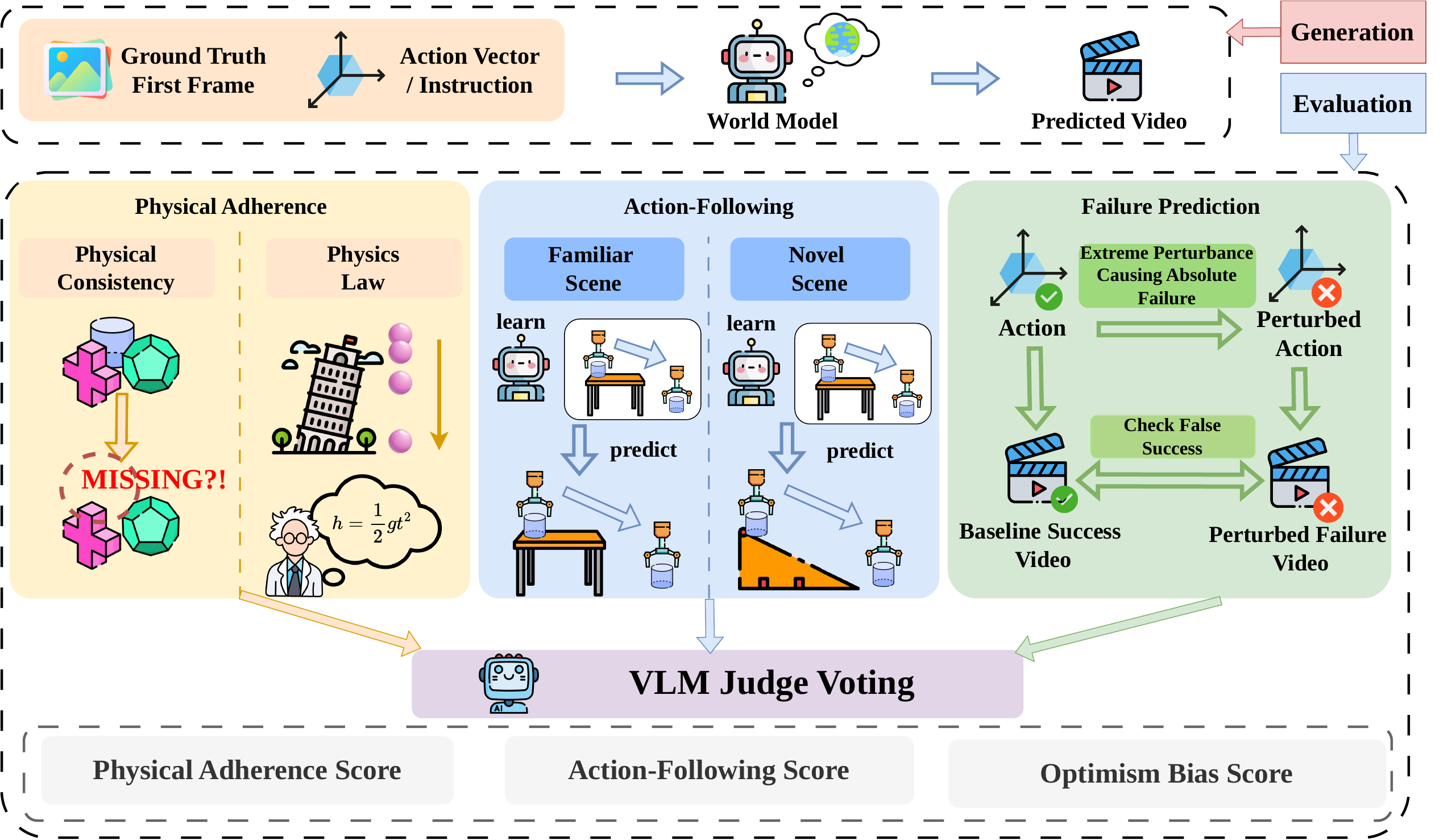}
  \caption{
    Overview of \textsc{MIRABench}.
    From robotic manipulation episodes, \textsc{MIRABench} constructs nominal and failure-inducing action inputs, evaluates generated rollouts through three levels of action-conditioned reliability, and use VLM evaluators from structured annotation data.
    Rule-based kinematic checks complement VLM for physics-law compliance.
  }
  \label{fig:framework}
\end{figure*}

MiraBench operationalizes \emph{action-conditioned reliability} through the pipeline in Fig.~\ref{fig:framework}. MiraBench is built from manipulation episodes in the GR-1 humanoid corpus~\citep{gao2026dreamdojogeneralistrobotworld} and the Lingchu bimanual dataset, covering contact-rich tasks such as pick-and-place, pouring, folding, bimanual handover, sliding, and free-fall interactions.
For nominal evaluation, models receive an initial observation together with either a raw motor command sequence or a natural-language task instruction.
For failure evaluation, we construct paired counterfactual inputs by perturbing a nominal successful action into a physically interpretable failure-inducing action while keeping the initial scene fixed.
For descriptive-action models, the same failure modes are expressed as natural-language instructions.
This design supports two complementary action modalities, \textit{precise actions} and \textit{descriptive actions}, allowing MiraBench to localize failures in motor grounding, semantic task understanding, or success-biased prediction.
The resulting corpus contains 906 generated videos and 16,704 human annotation decisions across four annotation modules.

\noindent\textbf{Level 1: Physics Adherence.}
Physics Adherence evaluates whether a generated rollout is physically meaningful before action fidelity is assessed.
It consists of two modules.
\textit{Physical Consistency} measures whether the manipulated object remains coherent over time, including its shape, material, size, and colour, as well as identity preservation through occlusion.
\textit{Physics Law Compliance} targets cases where videos look coherent but encode implausible motion.
For object-motion episodes, we extract trajectories with \textit{SAM2.1}~\citep{ravi2024sam2segmentimages} and evaluate simple kinematic cues such as uniformly-accelerated motion, impact behavior, post-contact drift, and bounce decay, complemented by a 10\% VLM-based video-quality term that flags rollouts in which no scorable translational motion including free-fall, slide and push occurs.

\noindent\textbf{Level 2: Action-Following Fidelity.}
Action-Following Fidelity tests whether a physically plausible rollout actually reflects the conditioning action.
A model may generate a coherent video while completing the wrong task, manipulating the wrong object, or producing motion that is smooth but semantically unrelated to the command.
\textsc{MIRABench} therefore evaluates three dimensions: task completion, target-object grounding, and motion quality.
These dimensions assess whether the predicted outcome matches the action and task intent rather than merely forming a plausible continuation.
To avoid overfitting the evaluation to a single embodiment, \textsc{MIRABench} includes cross-embodiment settings spanning \textit{GR-1}, \textit{Unitree G1}, \textit{DROID}~\citep{khazatsky2025droidlargescaleinthewildrobot}, and bimanual manipulation.

\noindent\textbf{Level 3: Optimism Bias Detection.}
Optimism Bias Detection probes whether a model preserves failure when the action implies failure.
We define six failure-inducing perturbation families grounded in common manipulation failures: insufficient grip force, premature release, carry slip, contact oscillation, wrist tilt, and approach overshoot.
Each perturbation keeps the visual scene fixed while changing the action-implied outcome.
For each episode, the model generates a nominal rollout and a counterfactual rollout.
A faithful model should predict the expected failure under the perturbed action, while an optimism-biased model produces a counterfactual rollout that remains semantically aligned with the successful baseline.
The evaluation focuses on task outcome and object state, not incidental rendering differences.

\subsection{Human-Grounded Evaluator}
\label{sec:benchmark:vlm}

\textsc{MIRABench} uses human annotation as supervision for automated evaluators. We obtain structured human annotations on physical consistency, physics-law compliance, action-following outcomes, and optimism-bias indicators. For automated evaluation, we deploy frontier-scale VLM models as zero-shot judges and stabilize their outputs through multi-frame voting: uniformly sampling frames from generated videos and aggregating their independent judgments. Specifically, Level 1 samples 20 evenly spaced frames for frame difference calculation and consistency comparison between the i-th and (i+10)-th frames, Level 2 aggregates across 16 sampled frames with different weights, and Level 3 uses 7-frame majority voting on late-phase frames where perturbation effects are most salient. This multi-frame approach reduces sensitivity to individual frame noise while preserving temporal context. On held-out expert-labeled samples, our evaluators achieve agreement above 85\% with human annotations and outperform Gemini-2.5-pro by 30\%. Full details are in Appendix~\ref{app:scoring}.

\section{Experiments and Analysis}
\label{sec:experiments}

We evaluate \textsc{MIRABench} on 12 representative model configurations spanning vector-conditioned robotic world models and text-conditioned generative world models.
The vector-conditioned group includes \textit{DreamDojo}, \textit{DreamDojo-GR1}, and \textit{Cosmos-GR1} at 2B and 14B scales, all conditioned on raw action vectors and an initial frame.
The text-conditioned group includes \textit{Cosmos-GR1} in text mode, \textit{Wan2.1/2.2}, \textit{WanX}, \textit{Happy Horse}, and \textit{Kling}, conditioned on natural-language task instructions and an initial frame.
This suite covers conditioning paradigms, model scales, access types, and systems released by major industrial and academic groups.
All scores are produced by \textsc{MIRABench}'s human-grounded evaluators, and Table~\ref{tab:main_results_aligned} reports the complete scores.


\definecolor{GroupGray}{HTML}{F4F4F4}
\definecolor{PendingGray}{HTML}{777777}
\newcommand{\tbd}{\textcolor{PendingGray}{\textsc{TBD}}}

\begin{table*}[t]
  \centering
  \caption{
    Main results on MiraBench across representative vector-conditioned and text-conditioned world models.
    Results are grouped by the three reliability levels.
    Higher is better for all scores.
  }
  \label{tab:main_results_aligned}
  \scriptsize
  \setlength{\tabcolsep}{4.0pt}
  \renewcommand{\arraystretch}{1.08}

  \begin{tabular*}{1.0\textwidth}{@{\extracolsep{\fill}}l c ccc ccc c@{}}
    \toprule
    \multirow{2}{*}{\textbf{Model}}
    & \multirow{2}{*}{\makecell{\textbf{Scale}\\\textbf{/ Access}}}
    & \multicolumn{3}{c}{\textbf{L1: Physics Adherence}}
    & \multicolumn{3}{c}{\textbf{L2: Action-Following}}
    & \multicolumn{1}{c}{\textbf{L3: Failure Preservation}} \\
    \cmidrule(lr){3-5} \cmidrule(lr){6-8} \cmidrule(l){9-9}
    &
    & Obj.$\uparrow$
    & Occ.$\uparrow$
    & Rule$\uparrow$
    & TCR$\uparrow$
    & OPS$\uparrow$
    & Gen.$\uparrow$
    & Bias Res.$\uparrow$ \\
    \midrule

    \rowcolor{GroupGray}
    \multicolumn{9}{@{}l}{\textit{Vector-conditioned models}} \\
    DreamDojo        & 2B  & 21.3 & 58.7 &  7.0 & 14.0 & 90.0 & 100.0 & 48.7 \\
    DreamDojo-GR1    & 2B  & 43.3 & 64.0 & 25.5 & 92.0 & 98.0 &  42.6 & 23.1 \\
    DreamDojo        & 14B & 38.7 & 79.0 &  7.8 & 18.0 & 92.0 & 100.0 & 12.8 \\
    DreamDojo-GR1    & 14B & 53.0 & 68.3 & 22.7 & 92.0 & 96.0 &  42.6 & 12.8 \\
    Cosmos-GR1       & 2B  & 30.5 & 51.0 & 17.0 & 68.0 & 98.0 &  54.2 & 17.5 \\
    Cosmos-GR1       & 14B & 30.3 & 53.7 & 14.0 & 54.0 & 96.0 &  71.2 & 15.4 \\
    \midrule

    \rowcolor{GroupGray}
    \multicolumn{9}{@{}l}{\textit{Text-conditioned models}} \\
    Cosmos-GR1        & 14B    & 10.3 & 42.3  & 10.9 & 76.0 & 56.0 &  97.4 & 30.8 \\
    WAN 2.1           & 14B    & 51.7 & 67.3  & 18.1 & 58.0 & 74.0 & 100.0 & 87.2 \\
    WAN 2.2           & 5B     & 65.2 & 81.4  & 10.6 & 92.0 & 90.0 &  72.6 & 56.4 \\
    Happy Horse       & Closed & 92.0 & 94.0  & 12.7 & 96.0 & 94.0 &  97.4 & 56.4 \\
    WanX              & Closed & 77.2 & 87.9  & 10.3 & 70.0 & 90.0 & 100.0 & 46.2 \\
    Kling 3.0 Omni    & Closed & 85.7 & 100.0 & 14.6 & 100.0 & 94.0 & 100.0 & 35.9 \\
    \bottomrule
  \end{tabular*}

  \vspace{0.5mm}
  \begin{minipage}{1.0\textwidth}
    \scriptsize
    Obj. = Object Consistency; Occ. = Occlusion Consistency; Rule = Physical Rule Adherence;
    TCR = Task Completion Rate; OPS = Object Preservation Score; Gen. = Generalization;
    Bias Res. = Bias Resistance.
    Vector-conditioned Cosmos-GR1 entries use 6000 fine-tuning steps.
  \end{minipage}
  \vspace{-0.25in}
\end{table*}

\begin{figure*}[t]
  \centering
  \includegraphics[width=0.98\textwidth]{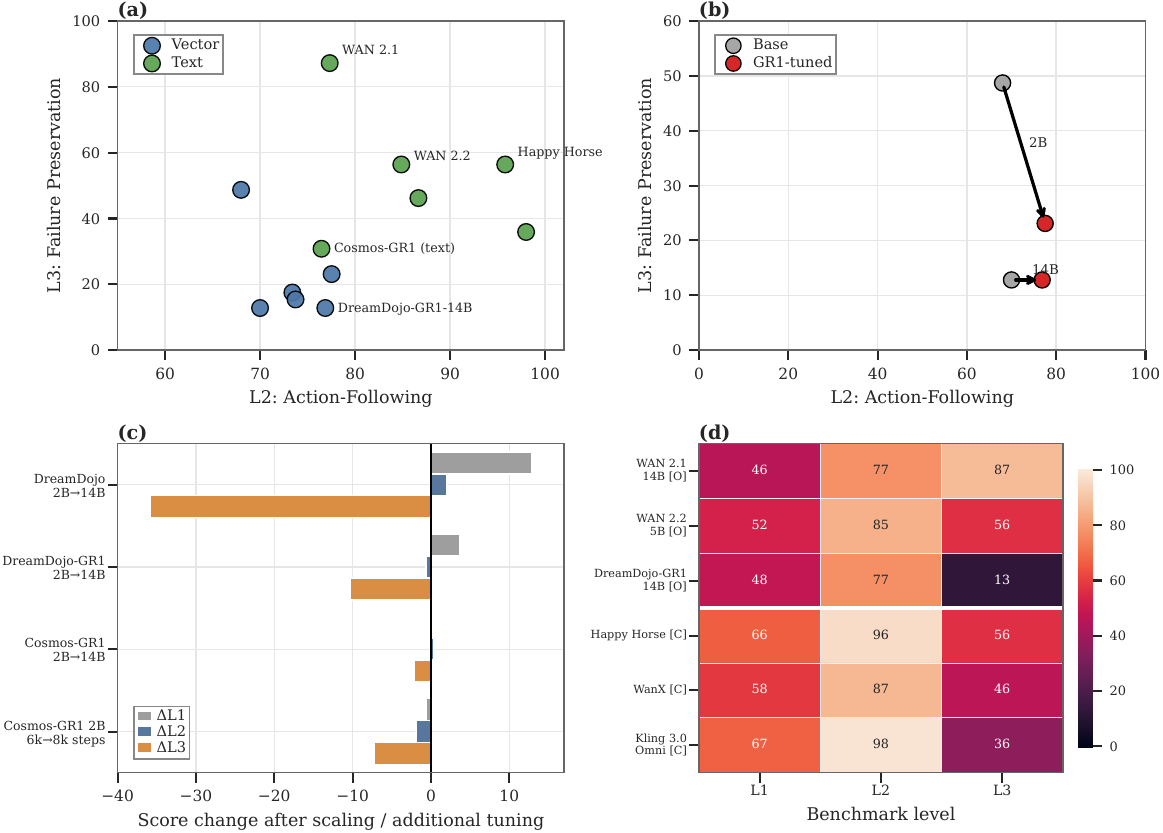}
  \caption{
    (a) Action-following does not guarantee failure preservation.
    (b) GR1 post-training improves task execution while weakening failure preservation at fixed scale.
    (c) Scaling and additional fine-tuning produce non-uniform changes across all three benchmark levels.
    (d) Open-weight [O] and closed-source [C] systems show complementary strengths, but no model dominates all levels.
  }
  \label{fig:summary_findings}
\end{figure*}

\noindent\textbf{Finding 1: Action-following is decoupled from failure preservation.}
As shown in Fig.~\ref{fig:summary_findings}(a), \textsc{MIRABench} reveals a clear gap between performance under nominal actions and reliability under failure-inducing actions.
Models with strong action-following scores can still fail to preserve failure outcomes.
For example, \textit{DreamDojo-GR1-14B} achieves high task completion and object preservation, yet its failure-preservation score remains among the lowest.
Happy Horse similarly performs strongly on physics and action following, but moderately on failure preservation.
Thus, successful-looking rollouts do not imply a world model preserves the causal relationship between actions and outcomes.

\noindent\textbf{Finding 2: Success-only post-training improves the wrong capability.}
Fig.\ref{fig:summary_findings}(b) shows GR1 post-training substantially improves nominal action-following but not failure preservation.
For \textit{DreamDojo-2B}, task completion increases from 14.0 to 92.0 and object preservation from 90.0 to 98.0 after post-training, while failure preservation drops from 48.7 to 23.1.
At 14B scale, task completion rises from 18.0 to 92.0, but failure preservation remains low.
Success-dominated post-training strengthens normal task execution without teaching the model to represent failed outcomes.

\noindent\textbf{Finding 3: Scaling and extra fine-tuning are not reliable fixes.}
Fig.\ref{fig:summary_findings}(c) compares changes under scaling and additional fine-tuning.
Within the \textit{DreamDojo} family, the 14B model improves some physical-consistency metrics over the 2B model, but failure preservation drops from 48.7 to 12.8.
\textit{DreamDojo-GR1} shows nearly unchanged nominal action-following from 2B to 14B, while failure preservation again decreases.
\textit{Cosmos-GR1} also shows limited evidence that scale alone improves failure preservation.
The \textit{Cosmos-GR1} step ablation further shows that increasing fine-tuning from 6000 to 8000 steps changes the action-following profile without producing a monotonic gain in action-conditioned reliability.
These results indicate that the observed failures are not merely capacity or optimization deficits, but reflect missing supervision for action-outcome contingency.

\noindent\textbf{Finding 4: Model families differ, but no current system solves all levels.}
As summarized in Fig.~\ref{fig:summary_findings}(d), text-conditioned, vector-conditioned, open-weight, and closed-source models exhibit distinct failure profiles.
Some text-conditioned systems achieve stronger failure preservation, while several vector-conditioned models achieve high nominal action-following under raw action inputs.
However, no evaluated model dominates Physics Adherence, Action-Following Fidelity, and Failure Preservation simultaneously.
\textit{Wan2.1} attains the strongest failure-preservation score but is not uniformly strongest in physics or nominal action-following, while Happy Horse performs strongly on physics and action-following but still leaves a gap in failure preservation.
This heterogeneity motivates \textsc{MIRABench}'s hierarchical design: a single aggregate visual-quality score would obscure whether a model fails through physical incoherence, action insensitivity, or success-biased prediction.

\noindent\textbf{Further Implications.}
These findings show that generating successful-looking videos is not equivalent to learning a faithful action-conditioned simulator.
Mitigating optimism bias will likely require failure-aware training data, objectives that contrast nominal and failure-inducing actions, and physical priors that rule out impossible trajectories.
\textsc{MIRABench} provides a diagnostic testbed for measuring whether such interventions improve the specific level at which a model fails.

\noindent\textbf{Case study and human validation.}
Figure~\ref{fig:case} shows representative failures corresponding to the quantitative patterns above, including object persistence errors, physics-law violations, action mismatch, and success-biased failure suppression.
Additional galleries are provided in Appendix~\ref{app:physics_gallery}, Appendix~\ref{app:gallery:action}, and Appendix~\ref{app:gallery}.
All automated scores are grounded in the human annotation corpus described in Section~\ref{sec:framework}, which contains 906 generated videos and 16,704 structured annotation decisions across \textit{Physical Adherence} (\textit{Physical Consistency} and \textit{Physics Law Compliance}), \textit{Action Following}, and \textit{Optimism Bias Detection}.
Dataset examples and corpus statistics are summarized in Fig.~\ref{fig:dataset_composition}, with full rubrics and subset breakdowns in Appendix~\ref{app:annotation} and Appendix~\ref{app:appendix_h}.

\section{Conclusion}
\label{sec:conclusion}

We introduced \textsc{MiraBench}, a hierarchical benchmark for evaluating \emph{action-conditioned reliability} in robotic world models. By decomposing reliability into Physics Adherence, Action-Following Fidelity, and Optimism Bias Detection, MiraBench shifts evaluation from visual plausibility to simulator faithfulness. Its fine-grained human annotation corpus of 906 videos and 16,704 structured decisions provides supervision for human-grounded VLM evaluators, enabling scalable assessment of 12 representative model configurations. Our results show that current world models can appear visually plausible while failing to preserve action-outcome contingency: visual quality is a weak proxy for action fidelity, scaling and post-training do not reliably improve failure preservation, and optimism bias remains widespread. We hope MiraBench provides a diagnostic foundation for building world models that are not only realistic to watch, but reliable as simulators for embodied AI.

\noindent\textbf{Limitations and Future Work.}
\textsc{MIRABench} currently focuses on tabletop manipulation with short-horizon contact dynamics. Future extensions should cover broader embodied domains such as navigation, locomotion, deformable-object interaction, and long-horizon planning. Its perturbation taxonomy targets representative failure-inducing actions and is designed to be extensible rather than exhaustive. Future work can expand the perturbation space, collect larger-scale failure demonstrations, and develop objectives that preserve action-outcome contingency, such as contrastive action-outcome learning, failure-aware curricula, and physics-informed modeling.

\bibliographystyle{plainnat} 
\bibliography{references}

\appendix
\section{Official Dataset Introduction}
\label{appendix:datasets}

\begin{figure}[!t]
\centering
\begin{tikzpicture}[scale=0.78, every node/.style={transform shape},
    node distance = 0.82cm and 2.0cm,
    bblue/.style  = {rectangle, rounded corners=3pt, draw=blue!55,
                     fill=blue!7, text width=3.0cm, align=center,
                     font=\small, inner sep=5pt, minimum height=0.70cm},
    bgreen/.style = {rectangle, rounded corners=3pt, draw=green!55!black,
                     fill=green!7, text width=3.0cm, align=center,
                     font=\small, inner sep=5pt, minimum height=0.70cm},
    bblue2/.style = {bblue,  fill=blue!18,  font=\small\bfseries},
    bgreen2/.style= {bgreen, fill=green!18, font=\small\bfseries},
    borange/.style= {rectangle, rounded corners=3pt, draw=orange!65,
                     fill=orange!12, text width=3.4cm, align=center,
                     font=\small\bfseries, inner sep=5pt, minimum height=0.70cm},
    bred/.style   = {rectangle, rounded corners=3pt, draw=red!55,
                     fill=red!8, text width=3.4cm, align=center,
                     font=\small\bfseries, inner sep=5pt, minimum height=0.70cm},
    arr/.style    = {-{Stealth[length=4pt,width=3pt]}, thick, black!50},
    head/.style   = {font=\bfseries\small, align=center},
]

\node[bblue]  (sim)   {Isaac Lab\\Simulation};
\node[bblue]  (tele)  [below=of sim]   {10 Human Tele-\\operated Demos};
\node[bblue]  (mimic) [below=of tele]  {Mimic: $\times$5\\Augmentation};
\node[bblue]  (hdf5)  [below=of mimic] {HDF5 (50 demos)\\36-DOF / 54-DOF};
\node[bblue]  (lero)  [below=of hdf5]  {GR00T-LeRobot\\Converter};
\node[bblue2] (adset) [below=of lero]  {Arena-GR1 Dataset\\(JSONL + MP4)};

\draw[arr] (sim)   -- (tele);
\draw[arr] (tele)  -- (mimic);
\draw[arr] (mimic) -- (hdf5);
\draw[arr] (hdf5)  -- (lero);
\draw[arr] (lero)  -- (adset);

\node[head, above=0.28cm of sim] {NVIDIA Arena-GR1 Pipeline};

\node[bgreen]  (human) [right=4.5cm of sim]    {Human Operator\\(bare hand or glove)};
\node[bgreen]  (glove) [below=of human]         {Exoskeleton Glove\\+ Head Camera};
\node[bgreen]  (raw)   [below=of glove]         {RGB, Depth, Pose\\qpos, Fingertips};
\node[bgreen]  (align) [below=of raw]           {Temporal Align\\(10\,FPS unified)};
\node[bgreen]  (annot) [below=of align]         {Clip \& Step\\Annotation};
\node[bgreen2] (ldset) [below=of annot]         {SynData\\(Parquet + Zarr v3)};

\draw[arr] (human) -- (glove);
\draw[arr] (glove) -- (raw);
\draw[arr] (raw)   -- (align);
\draw[arr] (align) -- (annot);
\draw[arr] (annot) -- (ldset);

\node[head, above=0.28cm of human] {PsiBot SynData Pipeline};

\coordinate (mid) at ($(adset.south)!0.5!(ldset.south)$);
\node[borange] (pretrain) [below=1.4cm of mid]
    {Policy Pre-training\\(behavior cloning /\\world-model learning)};

\draw[arr] (adset.south) -- ++(0,-0.5) -| (pretrain.north);
\draw[arr] (ldset.south) -- ++(0,-0.5) -| (pretrain.north);

\node[bred] (deploy) [below=0.82cm of pretrain]
    {Robot Deployment\\(fine-tuning + RL)};

\draw[arr] (pretrain) -- (deploy);

\end{tikzpicture}
\caption{Data collection and processing pipelines.
  \textbf{Left (blue):} Arena-GR1 starts from 10 teleoperated demonstrations in Isaac Lab,
  expands to 50 via Mimic, and is stored as JSONL metadata and MP4 videos in GR00T-LeRobot format.
  \textbf{Right (green):} SynData captures real-world bimanual demonstrations with a
  full-modality exoskeleton rig, temporally aligns all streams to 10\,FPS, and stores
  trajectory data in Zarr v3 with Parquet-based indexing and clip-/step-level annotations.
  Both datasets feed a shared policy pre-training stage followed by robot-specific
  fine-tuning.}
\label{fig:dataset_pipeline}
\end{figure}
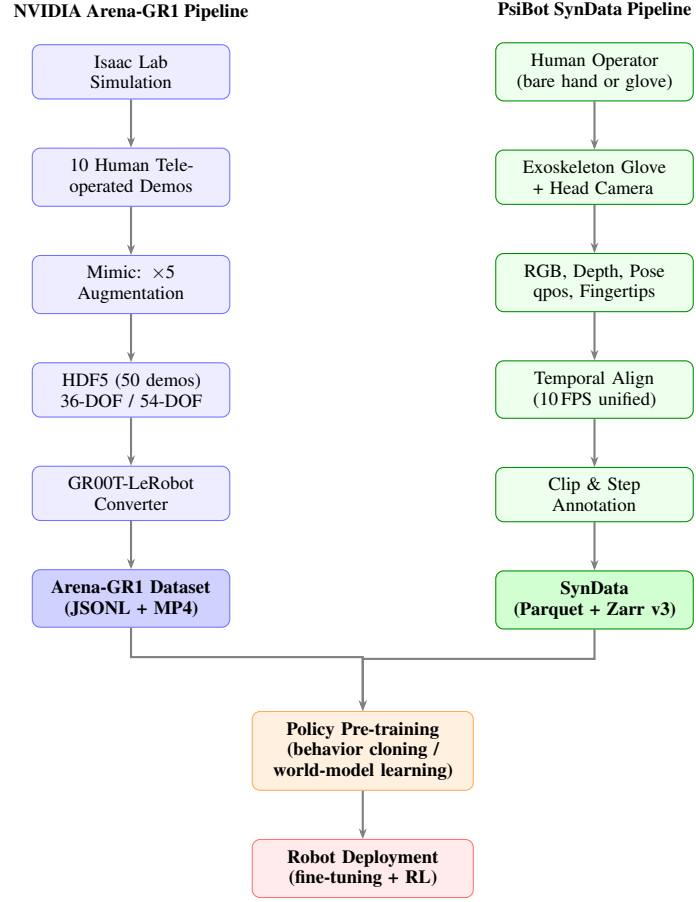

We leverage two large-scale robot manipulation datasets for training and evaluation:
the \textbf{NVIDIA Arena-GR1-Manipulation-Task} dataset and the \textbf{PsiBot SynData}
dataset.  Together, they provide complementary coverage of embodied manipulation
settings: the former supplies simulation-generated humanoid robot trajectories with
precise joint-space control signals, while the latter contributes real-world full-modality
bimanual hand demonstration data at scale.
Figure~\ref{fig:dataset_pipeline} illustrates the respective data collection pipelines.

\subsection{NVIDIA Arena-GR1-Manipulation-Task Dataset}
\label{appendix:datasets:arena}

\paragraph{Overview.}
The Arena-GR1-Manipulation-Task dataset~\cite{nvidia2025gr00tn1openfoundation} is an
open-source robot manipulation benchmark released by NVIDIA under the CC-BY-4.0
license as part of the Isaac GR00T initiative.  It is designed for training and
evaluating generalist humanoid manipulation policies via behavior cloning and
simulation-to-reality transfer research. The dataset conforms to the GR00T-LeRobot v2.0 specification~\cite{lerobot}, making it
directly compatible with GR00T N1.5 post-training pipelines.

\paragraph{Robot Platform and Task.}
Data are collected in the \textbf{IsaacLab-Arena} simulation environment using the
\textbf{Fourier GR-1} humanoid robot~\cite{nvidia2025gr00tn1openfoundation}. The dataset focuses on a
single contact-rich manipulation task: \emph{opening a microwave door}.  Visual
input is captured at 512$\times$512 resolution from a first-person-view
RGB camera.  All demonstrations are generated at 50\,Hz.

\paragraph{Data Collection.}
The dataset combines two complementary acquisition strategies:
\begin{itemize}[leftmargin=*]
    \item \textbf{Human teleoperation.}  10 demonstrations are collected by a human
          operator controlling the GR-1 within Isaac Lab via a depth camera and
          keyboard interface.
    \item \textbf{Synthetic augmentation.}  50 demonstrations are generated
          automatically from the 10 seed trajectories using
          MimicGen(5 per seed), forming the
          complete training corpus.  The 10 human-annotated seeds and the 50
          MimicGen-generated demonstrations are stored as two separate HDF5 files.
\end{itemize}
All demonstrations are generated at 50\,Hz and subsequently converted from HDF5 to
the GR00T-LeRobot format for downstream training.

\paragraph{Data Format and Structure.}
The released dataset provides three artefacts:
\begin{itemize}[leftmargin=*]
    \item \texttt{arena\_gr1\_manipulation\_dataset\_annotated.hdf5} — the 10
          human-annotated seed demonstrations.
    \item \texttt{arena\_gr1\_manipulation\_dataset\_generated.hdf5} — the full 50
          Mimic-generated demonstrations in raw HDF5.
    \item \texttt{lerobot/} — the GR00T-LeRobot formatted version of the 50
          demonstrations, organised as \texttt{data/chunk-\{id\}/episode\_\{id\}.parquet}
          and \texttt{videos/chunk-\{id\}/}.
\end{itemize}
The per-frame schema of the LeRobot artefact is detailed in
Table~\ref{tab:arena_schema}.

\begin{table}[h]
\centering
\small
\caption{Per-frame data schema of the Arena-GR1-Manipulation-Task (LeRobot format).}
\label{tab:arena_schema}
\setlength{\tabcolsep}{4pt}
\resizebox{\textwidth}{!}{%
\begin{tabular}{p{7.0cm}p{1.5cm}p{6.0cm}}
\toprule
\textbf{Field} & \textbf{Type} & \textbf{Description} \\
\midrule
\texttt{action}              & \texttt{float64}   & Desired joint positions (36 DoF) \\
\texttt{observation.state}   & \texttt{float64}   & Measured joint positions (54 DoF) \\
\texttt{timestamp}           & \texttt{float64}   & Simulation time in seconds \\
\texttt{episode\_index}      & \texttt{int64}     & Episode (demo) identifier \\
\texttt{task\_index}         & \texttt{int64}     & Multi-task loader index (always 0) \\
\texttt{annotation.human.action.task\_description} & \texttt{int64} & Index to language instruction in metadata \\
\texttt{annotation.human.action.valid} & \texttt{int64} & Annotation validity flag \\
\bottomrule
\end{tabular}}
\end{table}

\paragraph{Dataset Statistics.}
The full dataset comprises 50 demonstrations (50 RGB videos) with a total
storage footprint of 5.16\,GB.  All trajectories are generated at 50\,Hz.  Metadata is provided through four files:
\texttt{episodes.jsonl} (episode list and lengths), \texttt{tasks.jsonl} (task
list), \texttt{modality.json} (modality configuration), and \texttt{info.json}
(dataset-level statistics).

\paragraph{Train / Validation Split.}
Following standard practice, we reserve 5 demonstrations (10\%) as a held-out
validation set and use the remaining 45 for training.

\subsection{PsiBot SynData Dataset}
\label{appendix:datasets:syndata}

\paragraph{Overview.}
SynData is a large-scale real-world multimodal dataset released by
\textbf{PsiBot}, publicly available on Hugging
Face.\footnote{\url{https://huggingface.co/datasets/PsiBotAI/SynData}}
Despite its name, SynData consists entirely of \emph{real-world human
demonstrations} — ``Syn'' refers to its \emph{synchronized} full-modality capture
pipeline, not synthetic generation.  It is designed to serve as a foundational
pre-training corpus for embodied intelligence models, with a focus on data quality,
multimodal consistency, and temporal alignment.

\paragraph{Data Collection and Sensing Platform.}
Demonstrations are collected from human operators performing everyday bimanual
manipulation tasks using two complementary capture modes:
\begin{itemize}[leftmargin=*]
    \item \textbf{Exoskeleton glove mode.}  A proprietary exoskeleton glove
          achieves millimeter-level positioning accuracy and captures the full degrees
          of freedom of both hands and arms while preserving natural operator behavior.
    \item \textbf{Bare-hand mode.}  Unencumbered natural interaction data is collected
          to complement the structured glove recordings and enrich behavioral diversity.
\end{itemize}

\paragraph{Modalities.}
Every clip simultaneously records the following sensor streams, all temporally
aligned to a unified 10\,FPS timeline:

\begin{table}[h]
\centering
\caption{Modalities in the SynData dataset.}
\label{tab:syndata_modalities}
\setlength{\tabcolsep}{6pt}
\begin{tabular}{ll}
\toprule
\textbf{Modality} & \textbf{Description} \\
\midrule
\texttt{head\_rgb}               & RGB image sequence from the head camera \\
\texttt{head\_depth}             & Depth image sequence from the head camera \\
\texttt{head\_camera\_intrinsics} & Intrinsic parameters of the head camera \\
\texttt{head\_tracker2head\_camera} & Extrinsic: head tracker $\to$ head camera \\
\texttt{head\_pose}              & Head 6-DOF pose in world frame \\
\texttt{left\_wrist\_pose} / \texttt{right\_wrist\_pose} & Left/right wrist 6-DOF pose \\
\texttt{left\_qpos} / \texttt{right\_qpos}               & Left/right hand joint states \\
\texttt{left\_fingertip} / \texttt{right\_fingertip}     & Fingertip keypoints (left/right hand) \\
\bottomrule
\end{tabular}
\end{table}

\paragraph{Data Format and Organisation.}
SynData adopts a clip-based organisation stored in Zarr v3 format.
Each clip represents one complete task execution and is the minimum training unit.
The directory layout is:
\begin{itemize}[leftmargin=*]
    \item \texttt{task.json} — task ID-to-name mapping.
    \item \texttt{index/clips.parquet} — global clip index with fields
          \texttt{clip\_id}, \texttt{task\_key}, \texttt{volume\_id}, \texttt{rel\_path},
          \texttt{start\_idx}, \texttt{end\_idx}, and \texttt{num\_frames}.
    \item \texttt{annotations/clip\_annotations.parquet} — clip-level semantic
          descriptions.
    \item \texttt{annotations/clip\_steps.parquet} — step-level annotations for
          long-horizon task decomposition and hierarchical policy learning.
    \item \texttt{tasks/<task\_id>/<volume\_id>.zarr} — multimodal data volumes
          distributed as \texttt{.zarr.tar} packages, each storing modalities as
          independent arrays with the time dimension first.
\end{itemize}

\paragraph{Task Coverage.}
The dataset covers a wide spectrum of everyday bimanual manipulation tasks,
including clothes sorting, box sorting, fragile-item packing, and further categories
spanning contact-rich household and industrial scenarios.  Step-level annotations
support long-horizon task modeling and hierarchical policy learning.

\paragraph{Train / Validation Split.}
We apply stratified sampling across task categories, reserving 10\% of clips per
task as a held-out validation set and using the remaining 90\% for pre-training.

\section{Implicit Failure Perturbation Taxonomy}
\label{app:perturbations}

We define six implicit failure perturbations that modify the GR-1 action vector to encode physically interpretable manipulation failures.
The GR-1 action space is a 384D vector whose first 29 dimensions are active, organized as: left arm (7 joints), right arm (7 joints), left hand (6 joints), right hand (6 joints), and waist (3 joints); the remaining 355 dimensions are zero-padded.
All perturbations target specific joint groups within this 29D active subspace while preserving overall trajectory structure and numerical plausibility.
Each perturbation uses severity $s = 0.5$ in all experiments.

\subsection{Perturbation Definitions}

We denote the action at time $t$ restricted to a joint group $g$ as $a_{t,g}$, where $g \in \{\text{L-arm}, \text{R-arm}, \text{L-hand}, \text{R-hand}, \text{L-wrist}, \text{R-wrist}\}$.

\paragraph{1. Grip Force Insufficient (\texttt{grip\_force\_weak}).}
Left hand joint commands scaled to $(1-s)$ from $t_0 = 0.40T$ onward.
At $s=0.5$, grip force is halved; the object should slip during transport.
\begin{equation}
  a'_{t,\,\text{L-hand}} = (1 - s) \cdot a_{t,\,\text{L-hand}}, \quad t \geq \lfloor 0.40T \rfloor
\end{equation}

\paragraph{2. Premature Release (\texttt{premature\_release}).}
Left hand joints reduced to 2\% during carry phase ($0.40T$ to $0.80T$), before reaching placement target.
Object should fall mid-transport.
\begin{equation}
  a'_{t,\,\text{L-hand}} = 0.02 \cdot a_{t,\,\text{L-hand}}, \quad \lfloor 0.40T \rfloor \leq t \leq \lfloor 0.80T \rfloor
\end{equation}

\paragraph{3. Grip Carry Slip (\texttt{grip\_carry\_slip}).}
Left hand timing advanced by $\Delta = \lfloor T(0.15 + 0.20s) \rfloor$ frames; arm trajectory unchanged.
Gripper opens before arm reaches target.
\begin{equation}
  a'_{t,\,\text{L-hand}} = a_{\min(t + \Delta,\, T{-}1),\,\text{L-hand}}
\end{equation}

\paragraph{4. Contact Oscillation (\texttt{contact\_oscillation}).}
3-cycle sinusoidal injection on both left and right arm joints during contact phase ($0.25T$ to $0.70T$), amplitude $A = 0.4 \cdot \text{std}(a_{:,\,\text{L-arm}})$.
Prevents stable grasp formation.
\begin{equation}
  a'_{t,\,\text{L/R-arm}} = a_{t,\,\text{L/R-arm}} + A \cdot \sin\!\left(\tfrac{6\pi(t - t_0)}{t_1 - t_0}\right)
\end{equation}

\paragraph{5. Wrist Tilt During Grasp (\texttt{wrist\_tilt\_grasp}).}
Both left and right wrist joints (2 per wrist) offset by $+0.8$ rad from $0.15T$ to $0.85T$, causing incorrect contact geometry.
\begin{equation}
  a'_{t,\,\text{L/R-wrist}} = a_{t,\,\text{L/R-wrist}} + 0.8
\end{equation}

\paragraph{6. Approach Overshoot (\texttt{approach\_overshoot}).}
Left arm joint trajectory scaled $\times 1.30$ during approach ($0.10T$ to $0.75T$); end-effector overshoots object before gripper closes.
\begin{equation}
  a'_{t,\,\text{L-arm}} = 1.30 \cdot a_{t,\,\text{L-arm}}, \quad \lfloor 0.10T \rfloor \leq t \leq \lfloor 0.75T \rfloor
\end{equation}

\subsection{Per-Task Perturbation Assignment}

Each task receives 3 perturbations (plus baseline), with 2 mandatory types applied universally and 1 task-specific type selected based on task affordance:

\begin{table}[h]
  \caption{Perturbation schedule per task. All tasks share \texttt{grip\_force\_weak} and \texttt{premature\_release}; the third varies.}
  \label{tab:perturbation_schedule}
  \centering\small
  \begin{tabular}{ll}
    \toprule
    Task & Third Perturbation \\
    \midrule
    gr1\_pnp\_apple, fold\_cloth & \texttt{wrist\_tilt\_grasp} \\
    gr1\_pnp\_mango, gr1\_egodex, pnp\_corn, pnp\_dragonfruit & \texttt{contact\_oscillation} \\
    gr1\_pnp\_pear, pour\_items & \texttt{approach\_overshoot} \\
    pnp\_cucumber & \texttt{grip\_carry\_slip} \\
    \bottomrule
  \end{tabular}
\end{table}

\subsection{Translation to Natural Language (Descriptive Action Modality)}
\label{app:perturbation_prompts}

For the text-instruction-conditioned evaluation pipeline, each vector-level perturbation is translated into a natural-language prompt that describes the same failure mode semantically.
Prompts are generated by Gemini-2.5-Pro given the task context and perturbation specification, then \textbf{manually verified by human annotators} against the corresponding ground-truth video to ensure factual accuracy and physical plausibility.

Each task--episode pair has a \texttt{prompts.json} file containing one baseline prompt and one prompt per perturbation condition.
Below are representative examples for the \texttt{gr1\_pnp\_apple} task:

\begin{table}[h]
  \caption{Example natural-language prompts for the descriptive action modality (task: pick-and-place apple).}
  \label{tab:prompt_examples}
  \centering\small
  \begin{tabular}{p{3.2cm}p{8cm}}
    \toprule
    Condition & Prompt \\
    \midrule
    Baseline &
    ``From a first-person perspective, picks up a red apple from the center of a wooden table and carefully places it into the bottom shelf of a two-tiered wooden crate on the right.'' \\[4pt]
    \texttt{grip\_force\_weak} &
    ``From a first-person perspective, a robot with black hands attempts to pick up a red apple from a wooden table using only 0.1 newton of grip force.'' \\[4pt]
    \texttt{premature\_release} &
    ``From a first-person perspective, a robot with black hands picks up a red apple from a wooden table but releases its grip prematurely while moving it towards the wooden crate on the right, releasing it before reaching the destination.'' \\[4pt]
    \texttt{wrist\_tilt\_grasp} &
    ``From a first-person perspective, awkwardly grasps a red apple from a wooden table, its wrist bent sideways at an unnatural angle, before moving the unstably held apple to place it into the bottom shelf of the wooden crate.'' \\
    \bottomrule
  \end{tabular}
\end{table}

\paragraph{Prompt design principles.}
\begin{enumerate}[leftmargin=*]
  \item \textbf{Failure mode explicit}: the prompt explicitly states the physical failure (e.g., ``0.1 newton of grip force'', ``releases prematurely'') so that any text-conditioned model with adequate language understanding should generate the corresponding failure outcome.
  \item \textbf{Task context preserved}: the prompt retains full scene description (objects, spatial layout, robot morphology) so the model is not confused about what task is being attempted.
  \item \textbf{Human-verified}: all prompts are reviewed by annotators who confirm (a)~the described failure matches the vector-level perturbation effect, and (b)~the scene description matches the first frame the model will receive as conditioning.
\end{enumerate}

\subsection{Summary Table}

\begin{table}[h]
  \caption{Consolidated perturbation reference.}
  \label{tab:perturbation_summary}
  \centering\small
  \setlength{\tabcolsep}{4pt}
  \begin{tabular}{llccc}
    \toprule
    Type & Target Joints & Phase (\%$T$) & Operation & Expected Failure \\
    \midrule
    \texttt{grip\_force\_weak}     & L-hand (6D)          & 40--100 & $\times(1{-}s)$ & Object slip \\
    \texttt{premature\_release}    & L-hand (6D)          & 40--80  & $\times 0.02$   & Mid-air drop \\
    \texttt{grip\_carry\_slip}     & L-hand (6D)          & all     & shift $+\Delta$ & Early release \\
    \texttt{contact\_oscillation}  & L/R-arm (7D+7D)      & 25--70  & $+A\sin(\cdot)$ & Unstable contact \\
    \texttt{wrist\_tilt\_grasp}    & L/R-wrist (2D+2D)    & 15--85  & $+0.8$ rad      & Misaligned grasp \\
    \texttt{approach\_overshoot}   & L-arm (7D)           & 10--75  & $\times 1.30$   & Missed object \\
    \bottomrule
  \end{tabular}
\end{table}

\paragraph{Design principles.}
The perturbations satisfy three criteria:
(1)~\textit{Physical interpretability}: each corresponds to a named failure mode in manipulation literature;
(2)~\textit{Numerical plausibility}: perturbed vectors remain within or near the normalized $[-1, 1]$ range;
(3)~\textit{Guaranteed failure}: under correct physics, each perturbation necessarily prevents task completion.
A world model that produces success under these conditions is exhibiting optimism bias by definition.


\section{Natural Language Prompt Generation}
\label{app:prompts}

MiraBench uses natural-language prompts in two distinct roles:
(1)~as \textit{evaluation prompts} for VLM-based scoring across all three levels, and
(2)~as \textit{conditioning prompts} for the descriptive action modality in Levels~2 and~3.
Both are generated by Gemini-2.5-Pro and human-verified; their design differs in language, length, and purpose.

\subsection{Evaluation Prompts (Levels 1--3)}

The VLM evaluators at all three levels receive composite videos alongside structured natural-language queries.
Rather than hand-crafting these queries, we adopt a \textit{video-to-prompt inversion} pipeline: Gemini-2.5-Pro watches the video (uniformly sampled to 16 frames) and generates a factual description of the observed content, which then serves as context for the evaluation query.

\paragraph{Generation protocol.}
Gemini receives a system prompt enforcing strict writing constraints:
\begin{itemize}[leftmargin=*]
  \item \textbf{Positive-only language}: no negations or evaluation-style phrases (e.g., ``no penetration'', ``no deformation'').
  \item \textbf{Verb-driven structure}: ``who does what to which object, in what direction, along what path.''
  \item \textbf{Precise visual grounding}: colors, materials, and spatial relations must be explicitly stated (e.g., ``dark green transparent glass mug'' rather than ``cup'').
  \item \textbf{Observable-only}: no subjective intent, no unobservable quantities (forces, internal states).
\end{itemize}

The output is a prompt in Wan2.1-native style (for Level~1 evaluation on Chinese-prompt models) structured as: scene $+$ subject $+$ object (with color/position) $+$ action (with direction) $+$ static constraints.
A companion \texttt{negative\_prompt} lists both generic quality negatives and video-specific physical violation terms.

\paragraph{Few-shot calibration.}
Each Gemini call includes 3 randomly sampled seed examples (drawn from 5 human-written references spanning different task categories) to enforce consistent style, length, and structure across the full dataset.

\paragraph{Human verification.}
All generated prompts are reviewed by annotators who verify (a)~factual accuracy against the video content, and (b)~absence of evaluation-biasing language.
Prompts failing verification are regenerated with additional constraints.

\subsection{Conditioning Prompts for Descriptive Action Modality (Levels 2--3)}

For the text-instruction-conditioned pipeline, each task--episode pair requires a set of prompts: one baseline prompt describing normal task execution, plus one prompt per perturbation condition describing the corresponding failure mode.
These prompts serve as the model's \textit{input} (conditioning signal), not as evaluation queries.

\paragraph{Baseline prompts.}
Generated by Gemini-2.5-Pro in English, describing what the robot does in the ground-truth video from a first-person perspective.
Structure: viewpoint $+$ robot description $+$ object identification $+$ action sequence $+$ placement target.
Example: ``From a first-person perspective, picks up a red apple from the center of a wooden table and carefully places it into the bottom shelf of a two-tiered wooden crate on the right.''

\paragraph{Perturbation prompts.}
Each perturbation type is translated from its vector-level specification into a semantically equivalent natural-language description.
The prompt explicitly states the failure mechanism while preserving full scene context:

\begin{itemize}[leftmargin=*]
  \item \texttt{grip\_force\_weak} $\rightarrow$ ``\ldots attempts to pick up [object] using only 0.1 newton of grip force.''
  \item \texttt{premature\_release} $\rightarrow$ ``\ldots picks up [object] but releases its grip prematurely \ldots releasing it before reaching the destination.''
  \item \texttt{grip\_carry\_slip} $\rightarrow$ ``\ldots the [object] slips in its grasp \ldots''
  \item \texttt{contact\_oscillation} $\rightarrow$ ``\ldots its fingers oscillate and tremble with unsteady pressure before attempting to lift \ldots''
  \item \texttt{wrist\_tilt\_grasp} $\rightarrow$ ``\ldots its wrist bent sideways at an unnatural angle \ldots''
  \item \texttt{approach\_overshoot} $\rightarrow$ ``\ldots overshoots past the [object] before correcting its position to grasp it \ldots''
\end{itemize}

\paragraph{Generation and verification.}
Perturbation prompts are generated by Gemini-2.5-Pro given: (1)~the baseline prompt, (2)~the perturbation type name and its physical description, and (3)~the first frame of the episode for visual grounding.
All generated prompts are then \textbf{manually reviewed by human annotators} who verify:
\begin{enumerate}[leftmargin=*]
  \item The described failure mode is physically consistent with the vector-level perturbation (e.g., ``0.1 newton'' correctly reflects the $\times 0.5$ grip force reduction).
  \item The scene description (objects, colors, spatial layout) matches the first frame the model will receive.
  \item The prompt does not inadvertently reveal evaluation criteria or contain ambiguous language.
\end{enumerate}

\paragraph{Cross-task consistency.}
Prompts across all 9 tasks follow a uniform structure (viewpoint $+$ agent $+$ failure description $+$ object $+$ intended action), ensuring that differences in model performance across tasks reflect task difficulty rather than prompt quality variation.
The full set of prompts is released as part of the benchmark data.

\section{Framework Scoring}
\label{app:scoring}
Each MiraBench evaluation level employs a dedicated scoring pipeline that combines a VLM-based perceptual judge with task-specific inference strategies.
All three pipelines share a common design principle: rather than fine-tuning a domain-specific model, we leverage a frontier-scale VLM as a zero-shot judge and stabilize its outputs through structured prompting and majority voting.
This appendix documents the complete scoring logic, prompt templates, and validation results for each level.

\subsection{Physics (Level 1)}

The Physics level forms the diagnostic core of MiraBench. Before a generated
rollout can be assessed for action-following or optimism bias, it must remain
physically valid. We split Level~1 into two sub-levels:
\emph{Physical Consistency Detection} (behavioural-level coherence;
Section~\ref{app:scoring:phys:pc}) and \emph{Physics-Law Compliance}
(quantitative kinematic validity under a single conservative force;
Section~\ref{app:scoring:phys:pl}). Both operate \emph{reference-free}: the
score depends only on the predicted rollout $\hat{v}$, since physical laws
hold unconditionally and the level must be evaluable on counterfactual
action sequences for which no reference exists.

\subsubsection{Physical Consistency Detection (Level 1a)}
\label{app:scoring:phys:pc}

We measure behavioural-level physical consistency through two orthogonal
indicators, both implemented as zero-shot pairframe binary judgements on
InternVL3-78B \citep{internvl3}:
\begin{itemize}[leftmargin=*]
    \item \textbf{D1 — Object Consistency.} The manipulated object should
    preserve plausible shape, material, size, and colour during continuous
    visibility \citep{vbench2}.
    \item \textbf{D2 — Occlusion Consistency.} When the object is occluded
    and reappears, it should remain the same object.

\end{itemize}
The two are by design orthogonal --- object physics vs.\ object identity.
The combined \textbf{PCS} score is the equal-weight mean of the two
sub-scores. Since both indicators share the same frame-extraction
front-end, their N/A flags are perfectly correlated; PCS is therefore
either a real number in $[0,1]$ or \texttt{None} when both indicators
abstain.

\paragraph{Pipeline (shared by D1 and D2).}
We extract 20 evenly-spaced frames and form 10 \emph{midcut} pairs
(frame $i$ paired with frame $i+10$, $i=0,\dots,9$). Each pair is
vertically stacked into a single image and submitted to InternVL3-78B,
which returns a binary verdict on that pair (\texttt{A}=consistent,
\texttt{B}=inconsistent). The video-level score on a $0$--$100$ scale rewards each pair the
VLM judges consistent:
\begin{equation}
s_d = 100\cdot\Bigl(1 - \tfrac{1}{n}\sum_{i=1}^{n} \mathbb{1}[\mathrm{vote}_i = \texttt{B}]\Bigr)
\;\in\; [0, 100], \qquad n=10,
\label{eq:vlm_score}
\end{equation}
i.e.\ each \texttt{B} vote subtracts $100/n=10$ points from a perfect
score of $100$. A categorical label \texttt{B}, set
whenever any pair is judged \texttt{B}, is also reported but does
\emph{not} enter the PCS aggregation.

\paragraph{D1: Object Consistency prompt.}
\begin{quote}\small
These two frames are from a robot manipulation video. Frame 1 is on
top, Frame 2 is on bottom.

Identify the MAIN OBJECT --- the item being grasped or moved by the
robot arm (not the arm itself).

Carefully compare the main object between the two frames across ALL of
the following:
\begin{itemize}
\item \textbf{SHAPE:} Does it keep the same overall form? Any unexpected
deformation, collapse, bending, or structural change $\to$ inconsistent.
\item \textbf{MATERIAL:} Does it behave like the same material type?
A rigid object that looks soft / rubbery, or a flexible object that
looks stiff $\to$ inconsistent. Even subtle changes in how the object
deforms or holds its shape count.
\item \textbf{SIZE:} Is it approximately the same size? Any unexplained
sudden change in scale $\to$ inconsistent.
\item \textbf{COLOUR:} Is the dominant colour stable? Sudden hue change
unexplained by lighting $\to$ inconsistent.
\end{itemize}

Judge whether the object remains PHYSICALLY PLAUSIBLE across the two
frames:
\begin{itemize}
\item \textbf{A. Consistent:} shape, material, size, and colour are
stable --- differences are clearly explained by viewpoint, grip angle,
or distance only.
\item \textbf{B. Inconsistent:} something physically implausible changed.
\textbf{Do NOT default to A} --- if you notice any suspicious change
that cannot be explained by viewpoint alone, choose B.
\end{itemize}

Output only \texttt{A} or \texttt{B} on the first line, then one
sentence describing what you observed.
\end{quote}

\paragraph{D2: Occlusion Consistency prompt.}
The Occlusion Consistency indicator uses the same pipeline as D1 but a
different prompt that explicitly handles occlusion and asks the VLM to
judge \emph{identity} rather than physical attributes. Instructing the
VLM to reason over the visible portions rather than abstaining when one
frame is partially hidden removes the up-front
``occlusion-event present?'' gate that earlier two-stage
partial-vs-full designs \citep{vbench2} required and reduces the N/A
rate from $\sim$70\% to $<$5\%.

\begin{quote}\small
These two frames are from a robot manipulation video. Frame 1 is on
top, Frame 2 is on bottom.

\textbf{Context:} one or both frames may involve occlusion --- the
object may be partially or fully hidden by the robot arm or another
object.

Focus on the main object (the item being grasped or moved, not the arm
itself). If one frame shows it partially hidden, judge only the visible
portions.

Judge whether the object looks like the SAME object in both frames ---
same identity, colour, shape, and appearance:
\begin{itemize}
\item \textbf{A. Consistent:} the visible portions look the same ---
any differences are explained by occlusion angle, viewpoint, or grip
only.
\item \textbf{B. Inconsistent:} the visible portions look genuinely
different in a way that occlusion alone cannot explain --- changed
colour, shape, texture, or appears to be a different object.
\end{itemize}

\begin{itemize}
\item Do NOT default to A. If you are unsure, choose B.
\item Err on the side of flagging inconsistency.
\end{itemize}

Output only \texttt{A} or \texttt{B} on the first line, then one
sentence about what you observed.
\end{quote}

\paragraph{Combined PCS score.}
\begin{equation}
\mathrm{PCS}(\hat{v}) = \tfrac{1}{2}\bigl(s_\text{obj}(\hat{v}) + s_\text{occ}(\hat{v})\bigr)
\;\in\; [0, 100], \qquad s_d \in [0, 100].
\label{eq:pcs}
\end{equation}
Equal weighting reflects the orthogonality of the two properties.
Table~\ref{fig:pcs_perf} reports per-indicator pairwise accuracy against
human annotations on the 68-pair occ68 head-to-head set.

\begin{table}[t]

\centering\small
\caption{PCS(Physical Consistency Score) per-indicator alignment with human annotations on 128 videos.
Pairwise accuracy is calculated via consistency matching between model outputs and human judgements.}
\begin{tabular}{lcc}
\toprule
Indicator & Backbone & Pairwise acc.\ (\%) \\
\midrule
Obj.\ Cons. & InternVL3-78B & 87.3 \\
Occ.\ Cons. & InternVL3-78B & 85.1 \\
\bottomrule
\end{tabular}
\label{fig:pcs_perf}
\end{table}

\paragraph{Key design decisions (Level 1a).}
\textbf{(1) Orthogonal two-dimensional decomposition.}\quad
Decomposing ``physical consistency'' into object physics and object
identity lets us trace any low PCS score back to a specific failure
dimension, each individually verifiable in our released human-annotation
corpus, and maps cleanly onto cognitive-science principles --- object
permanence under occlusion is grounded in infant-physics studies.
\noindent\textbf{(2) Pairframe binary judgement.}\quad
Earlier four-tier (A/B/C/D) VLM indicators \citep{vbench2}
disproportionately answer ``A'' on diffusion content, collapsing
four-tier signal to binary anyway. We use binary judgement directly with
an adversarial ``do not default to A'' prompt, giving a cleaner threshold
and lower agreement noise.
\noindent\textbf{(3) Midcut pairing with threshold~1.}\quad
Random pairing biases toward A by leaving most pairs at adjacent frames
where the object has barely moved; midcut spans half the video duration
to expose accumulated changes. Threshold $=$ 1 (any inconsistent pair
flags the whole video) reflects the strict requirement that any local
violation invalidates the entire generation.

\subsubsection{Physics-Law Compliance (Level 1b)}
\label{app:scoring:phys:pl}

Where Level 1a tests behavioural-level consistency, Level 1b stress-tests
\emph{quantitative kinematic compliance} under a single conservative
force --- gravity (free fall) or friction (horizontal push). Level 1b
uses an explicit kinematic-computation pipeline: \emph{no language-model
judgement enters the kinematic score}. A 10\% Video Quality Score (VQS)
contribution gates whether the video's physical premise (any object
exhibits a clear translational motion) is satisfied at all.

\paragraph{Task definition.}
Two motion regimes are dispatched automatically: \textbf{vertical
free-fall} ($a=g$, $y(t) = \tfrac{1}{2}gt^2 + v_0 t + y_0$) and
\textbf{horizontal push} ($a=-\mu g$,
$x(t) = -\tfrac{1}{2}\mu g\,t^2 + v_0 t + x_0$).
Expected motion is therefore quadratic on the appropriate axis; the
evaluator scores how well each scoring segment fits this quadratic form
plus three discrete event features.

\paragraph{Pipeline.}
\textbf{Stage~0 (Video Quality Score).} A VLM inspects six
evenly-spaced frames horizontally tiled into a single image and answers
two binary questions in JSON: \verb|video_ok| (recognisable robot scene)
and \verb|has_motion| (any object exhibits a clear translational motion
--- free-fall, slide, push, roll, etc. --- at some point in the clip).
The two answers map to a discrete $\mathrm{VQS}\in\{0,5,10\}$
(both False / video\_ok only / both True), reported on the same
$0$--$100$ scale as the final PhysLawScore. When VQS$=0$ or VQS$=5$,
the kinematic component is forced to $0$; only when VQS$=10$ does the
kinematic score enter the final aggregation. This prevents SAM2 +
polyfit from producing spuriously high scores on videos where no
scorable motion actually occurs (e.g.\ a static held object yields a
degenerate but smooth trajectory). Because the gate asks about
\emph{any} translational motion --- not free-fall specifically --- the
same Stage~0 check applies uniformly to both the vertical (free-fall)
and horizontal (push / slide) sub-pipelines.
\textbf{Stages 1--5 (kinematic).} A VLM (or any open-vocabulary
detector) returns the object bounding box in frame 0; SAM2.1 forward propagation yields a mask-centroid
trajectory; an axis classifier dispatches to vertical / horizontal /
mixed; the active axis is segmented into fall / rise / rest (vertical)
or push / slide / rest (horizontal); each scorable segment is fit with
a degree-2 polynomial.

\paragraph{Per-segment scoring.}
Each scorable segment yields a fitted acceleration $\hat{a}$. The
segment score is the product of three independent physical-validity
factors, each in $[0,1]$:
\begin{equation}
\mathrm{seg}\_\mathrm{score} = \mathrm{sign\_ok} \cdot \mathrm{magnitude\_ok} \cdot \mathrm{uniformity\_ok}.
\end{equation}
$\mathrm{sign\_ok}$ verifies that the conservative force has the correct
direction (positive $\hat{a}$ for fall, negative for slide).
$\mathrm{magnitude\_ok}$ is $1.0$ when $|\hat{a}|$ is within
$[0.3, 3]\times \frac{2|\Delta y|}{\Delta t^2}$ of the kinematic
expectation, with linear decay outside this range.
$\mathrm{uniformity\_ok}$ halves the segment, fits each half, and
computes the coefficient of variation between the two fitted
accelerations: $\mathrm{half\_cv}\le 0.15 \to 1.0$, with linear decay
to $0.0$ at $\mathrm{half\_cv}=0.80$. The polynomial $R^2$ is recorded
as a diagnostic but does \emph{not} enter the score.

The trajectory-level curve score is the length-weighted mean of segment
scores with a coverage factor and a hard-violation override:
\begin{equation}
\mathrm{curve} =
\begin{cases}
0 & \text{if any fall/rise segment has } \mathrm{sign\_ok}=0 \\
\mathrm{cov}\cdot \dfrac{\sum_s n_s \cdot \mathrm{seg}\_\mathrm{score}_s}{\sum_s n_s} & \text{otherwise,}
\end{cases}
\end{equation}
where $\mathrm{cov} = \min(1, \mathrm{coverage}/0.3)$.

\paragraph{Event features and gated fusion.}
Four discrete features --- velocity drop at impact
$(v_{\text{before}} - v_{\text{after}})/v_{\text{before}}$,
post-landing drift $\Delta y_{\text{post}}/0.10$, a binary
$\mathrm{has\_impact}$ indicator, and bounce decay $h_2/h_1$
(physical $<1$) --- are combined with weights $(0.30, 0.20, 0.30, 0.20)$
to yield $\mathrm{event}$. The kinematic score gates curve against
event:
\begin{equation}
\mathrm{kinematic\_score} =
\begin{cases}
0.30\cdot\mathrm{curve} + 0.70\cdot\mathrm{event} & \text{if } \mathrm{curve} \ge 0.30 \\
0.70\cdot\mathrm{curve} + 0.30\cdot\mathrm{event} & \text{if } \mathrm{curve} < 0.30,
\end{cases}
\end{equation}
preferring curve when it has flagged a clear violation and event
otherwise --- avoiding the case where a noise-sensitive curve fit drags
down a video whose event-level physics is in fact correct.

\paragraph{Final score.}
On the $0$--$100$ scale, the final PhysLaw score combines the kinematic
component (on $[0,1]$) with VQS:
\begin{equation}
\mathrm{PhysLaw}(\hat{v}) \;=\; 0.9\cdot\mathrm{effective\_physics} + \mathrm{VQS},
\label{eq:phys_law_final}
\end{equation}
where
\begin{equation}
\mathrm{effective\_physics} =
\begin{cases}
100\cdot\mathrm{kinematic\_score} & \text{if }\verb|has_motion|=\text{True},\\
0 & \text{otherwise.}
\end{cases}
\label{eq:phys_law_effective}
\end{equation}
A video the VLM judges to lack any scorable motion therefore cannot
exceed $5/100$, regardless of trajectory smoothness.

\paragraph{Physics-Law VQS prompt.}
\begin{quote}\small
You are evaluating a generated video that should show a
physics-scorable scene (free-fall, slide / push along a surface, or
any other clear, sustained translational motion). The image shows
several equally-spaced frames tiled left to right. Answer ONLY with
valid JSON:

\begin{verbatim}
{ "video_ok": true/false,
  "has_motion": true/false,
  "reason": "one short sentence" }
\end{verbatim}

\textbf{video\_ok:} the video shows a recognisable robot-arm scene
(not all-black, not corrupted, not static).
\textbf{has\_motion:} ANY object visible in the scene undergoes a
clear translational motion at some point --- free-fall, sliding,
being pushed, rolling, or otherwise traversing space. Be LENIENT;
set to \texttt{false} ONLY when the entire clip is essentially
static.
\end{quote}

\paragraph{Test-set performance and key design decisions (Level 1b).}
On the 89 graded videos in the released Physics-Law set,
$\mathrm{PhysLaw}$ matches well with human Grade A/B/C/D; per-object breakdown in
Section~\ref{sec:appendix_h_pl}.
\textbf{(1)~Explicit kinematic computation, no LLM as judge:} the 90\%
component is a quadratic fit and three explicit factors --- an auditor
can trace any low score back to a specific segment, axis, and factor.
\textbf{(2)~VLM as quality gate, not as judge:} a pure SAM2 + polyfit
pipeline can score highly on videos where no scorable motion occurs (a
static held object yields a degenerate but smooth trajectory); the
Stage~0 VQS detects this premise mismatch and caps Physics Law at
$5/100$ irrespective of trajectory smoothness.
\textbf{(3)~Multiplicative factor decomposition:}
$\mathrm{seg}\_\mathrm{score}$ is the product of three independent
checks rather than a single $R^2$, which would conflate three distinct
failure modes (anti-gravity, wrong magnitude, non-uniform acceleration).
\textbf{(4)~Length-weighted segment averaging instead of $\min$:} a
naive $\min$ lets a short, noise-sensitive rebound segment force the
trajectory to 0; the length-weighted mean is robust while still flagging
hard violations through the explicit ``any sign\_ok = 0 $\to$ 0''
override.

\subsection{Action Following (Level 2)}


\subsubsection{Task Completion Rate (TCR)}
\label{app:scoring:tcr}

TCR is a binary per-episode metric that asks whether the world model's predicted video demonstrates successful completion of the manipulation task.

\paragraph{Task definition.}
Given a predicted video $\hat{\mathbf{v}}$ and the corresponding natural-language task instruction $\ell$, determine whether the task goal is achieved in $\hat{\mathbf{v}}$ (TCR = 1) or not (TCR = 0).

\paragraph{Model.}
We use \textbf{InternVL3-78B} as the automated judge, deployed across 3 GPUs with tensor-parallel split.
No fine-tuning is applied; the model is used zero-shot with a task-specific prompt.

\paragraph{Inference: 16-frame whole-sequence judge.}
We uniformly sample 16 frames from the predicted video and present all 16 frames to the VLM in a single call, along with the task instruction.
The VLM outputs a single binary judgment for the entire video---no per-frame voting, no aggregation:
\begin{equation}
  \text{TCR} = v \in \{0, 1\}
\end{equation}
This holistic approach lets the VLM reason over the full temporal context, catching cases where task completion occurs mid-video or where motion progress across frames disambiguates single-frame uncertainty.

Importantly, \textbf{no ground-truth frames are shown to the judge}. Removing the GT reference eliminates a systematic bias where the VLM anchors on arm-pose similarity between predicted and ground-truth frames, which unfairly penalises models whose motion trajectories diverge from the GT while still achieving the task goal.

\paragraph{Prompt design.}\mbox{}\\[-0.6em]

\begin{tcolorbox}

\small\ttfamily
You are evaluating a robot manipulation video.\\
Here are 16 frames sampled uniformly from the predicted video, in chronological order.\\[4pt]
Task instruction: "\{instruction\}"\\[4pt]
Looking at the entire video: did the robot complete the task instruction?\\
\quad\textbullet\ Track object positions across the full sequence to confirm goal achievement.\\
\quad\textbullet\ Do NOT require any specific arm pose --- focus on whether the target object reaches the correct final state.\\[4pt]
Respond ONLY with 0 or 1.\\
1 = task completed in this video\\
0 = task not completed
\end{tcolorbox}

\paragraph{Model-level score.}
The TCR score for a model on a given split is the fraction of episodes with TCR\,=\,1, reported as a percentage:
\begin{equation}
  \text{TCR score} = \frac{1}{N}\sum_{n=1}^{N} \text{TCR}_n \times 100
\end{equation}

\paragraph{Key design decisions.}
\begin{enumerate}[leftmargin=*]
  \item \textbf{Whole-sequence over per-frame}: Presenting all 16 frames at once allows the VLM to reason about temporal progression---e.g.\ recognising that an object was successfully placed mid-video even if the arm subsequently moves away. Per-frame voting schemes risk losing this context and introduce sensitivity to aggregation hyper-parameters (tail weighting, vote threshold).
  \item \textbf{No ground-truth reference}: Showing GT frames alongside predicted frames creates an implicit arm-pose anchor that biases the judge against models whose trajectories differ from GT, even when the task goal is achieved. Removing GT frames eliminates this bias and evaluates the predicted video on its own merits.
  \item \textbf{Single binary output}: A straightforward 0/1 judgment avoids the need for confidence calibration, vote aggregation, or threshold tuning, making the metric simple and reproducible.
\end{enumerate}

\paragraph{Human consistency validation (GR1 split, 48 episodes).}
We validate the automated evaluator against human-verified labels on 48 GR1 episodes.
The evaluator achieves an accuracy of \textbf{87.5\%} (42/48) and a Spearman rank correlation of $\boldsymbol{\rho = 0.545}$ ($p < 0.001$) between the per-episode confidence score and the ground-truth binary label.


\subsubsection{Object Preservation Score (OPS)}
\label{app:scoring:ops}

OPS is a binary per-episode metric that assesses whether the world model maintains physically plausible object appearance throughout the predicted video, independent of whether the task is completed.

\paragraph{Task definition.}
Given a predicted video $\hat{\mathbf{v}}$ and the corresponding ground-truth video $\mathbf{v}^*$, along with the task instruction $\ell$, determine whether the objects in $\hat{\mathbf{v}}$ are visually coherent throughout the sequence (OPS = \textit{preserved}) or exhibit artefacts such as deformation, disappearance, or unnatural pop-in/pop-out events (OPS = \textit{flawed}).

\paragraph{Model.}
We use \textbf{InternVL3-78B} as the automated judge, deployed across 3 GPUs with tensor-parallel split.
No fine-tuning is applied; the model is used zero-shot with a task-specific prompt.

\paragraph{Inference: 16-frame uniform sampling with mean aggregation.}
We uniformly sample 16 frames from both the predicted and ground-truth video, producing 16 temporally aligned frame pairs.
Each pair is independently judged by the VLM, yielding a binary quality vote $q_i \in \{0, 1\}$.
Unlike TCR, temporal position is not expected to affect object quality, so votes are aggregated by simple mean:
\begin{equation}
  \text{confidence} = \frac{1}{16}\sum_{i=1}^{16} q_i
\end{equation}
The per-episode label is then:
\begin{equation}
  \text{OPS} = \begin{cases}
    \textit{preserved} & \text{if confidence} \geq 0.70 \\
    \textit{flawed}    & \text{if confidence} < 0.70
  \end{cases}
\end{equation}

\paragraph{Prompt design.} \mbox{}\\[-0.6em]

\begin{tcolorbox}
\small\ttfamily

You are evaluating the visual quality of a single predicted robot video frame.\\[4pt]
\quad\textbullet\ Frame1 = frame from the PREDICTED video (world model output --- evaluate this)\\
\quad\textbullet\ Frame2 = frame from the GROUND TRUTH video at the same timestamp (reference only)\\[4pt]
Task instruction: "\{instruction\}"\\[4pt]
Check Frame1 against Frame2 for the following issues:\\
\quad 1. Is the target object (the object being manipulated) clearly visible in Frame1,\\
\quad\quad without unexpected blurring, occlusion, or disappearance?\\
\quad 2. Are all objects in Frame1 free of distortion or unnatural deformation?\\
\quad 3. Are there no objects that pop in or pop out unnaturally between frames\\
\quad\quad (appearing or vanishing without physical cause)?\\[4pt]
Respond ONLY with 0 or 1. No explanation.\\
1 = Frame1 passes all checks (high quality, matches GT object presence)\\
0 = Frame1 fails at least one check (object issue, distortion, or pop artifact)
\end{tcolorbox}

\paragraph{Model-level score.}
The OPS score for a model on a given split is the fraction of \textit{preserved} episodes:
\begin{equation}
  \text{OPS score} = \frac{|\{n : \text{OPS}_n = \textit{preserved}\}|}{N} \times 100
\end{equation}

\paragraph{Human consistency validation (GR1 split, 26 human-annotated episodes).}
We validate the automated evaluator against human annotations using the binary \textit{preserved} vs.\ \textit{flawed} split.
The evaluator achieves a \textbf{92.3\%} accuracy (24 of 26 episodes agree with human judgment), confirming that the 0.70 threshold reliably separates episodes that humans judge as object-coherent from those they do not.

\paragraph{Key design decisions.}
\begin{enumerate}[leftmargin=*]
  \item \textbf{Uniform temporal weighting}: Object preservation is a frame-level property that should hold throughout the video, not just at the end.
  Simple mean aggregation avoids biasing the score towards any particular phase of the manipulation.
  \item \textbf{Binary label with conservative threshold}: A single \textit{preserved}/\textit{flawed} dichotomy reflects the fact that object artefacts are both highly salient and disqualifying for downstream use.
  The 0.70 confidence threshold requires a clear majority of frames to pass, catching videos where artefacts appear intermittently.
  \item \textbf{Object-focused prompt with three explicit checks}: Decomposing the quality judgment into three distinct failure modes (visibility, deformation, pop artefacts) guides the VLM towards consistent and interpretable binary decisions, reducing false negatives that arise from a holistic ``looks bad'' judgment.
\end{enumerate}


\subsubsection{Generalizability (GEN)}
\label{app:scoring:gen}

GEN measures how well a model generalises from its training split to unseen episodes, quantifying the gap between in-distribution and out-of-distribution TCR performance.

\paragraph{Definition.}
Let $\text{TCR}_{\text{GR1}}$ and $\text{TCR}_{\text{Gen}}$ denote a model's Task Completion Rate on the GR1 (in-distribution) and generalizability (out-of-distribution) splits, respectively. The raw generalisation gap is:
\begin{equation}
  \Delta = \text{TCR}_{\text{GR1}} - \text{TCR}_{\text{Gen}}
\end{equation}
A positive $\Delta$ indicates overfitting to the training distribution. To convert this gap into a score where higher is better and the range is bounded on $[0, 100]$, we apply an exponential decay with clipping:
\begin{equation}
  \text{GEN} = \min\!\Big(100,\; 100 \cdot e^{-\Delta / 100}\Big)
\end{equation}

\paragraph{Interpretation.}
\begin{itemize}[leftmargin=*]
  \item $\Delta \leq 0$ (no overfitting): $\text{GEN} \geq 100$, clipped to 100\%. The model generalises at least as well as it performs in-distribution.
  \item $\Delta > 0$ (overfitting): GEN decays exponentially, penalising larger gaps while remaining non-zero. For example, $\Delta = 2.7$ yields GEN $= 97.4\%$; $\Delta = 85.3$ yields GEN $= 42.6\%$.
\end{itemize}

\paragraph{Why exponential rather than linear?}
A linear transform ($100 - \Delta$) would assign GEN $= 0$ for $\Delta \geq 100$, creating a hard floor. The exponential decay ensures that even severely overfit models retain a non-zero score proportional to their gap, while small gaps remain close to 100\%. The divisor of 100 scales the decay so that a 10-point gap causes only ${\sim}1\%$ loss, matching the intuition that moderate overfitting should not be penalised as severely as extreme overfitting.

\paragraph{Current results.}
Table~\ref{tab:gen-results} reports TCR and GEN for all evaluated models.

\begin{table}[h]
\centering
\small
\begin{tabular}{lcc}
\toprule
\textbf{Model} & \textbf{TCR (GR1)} & \textbf{GEN} \\
\midrule
happyhorse              & 96.0\% & 97.4\% \\
dreamdojo\_14b          & 92.0\% & 42.6\% \\
dreamdojo\_2b           & 92.0\% & 42.6\% \\
wan2.2                  & 92.0\% & 72.6\% \\
cosmos\_14b             & 76.0\% & 97.4\% \\
wanx21\_i2v\_plus       & 70.0\% & 100.0\% \\
wan2.1                  & 58.0\% & 100.0\% \\
cosmos\_a2v\_14b        & 54.0\% & 71.2\% \\
cosmos\_a2v\_2b         & 40.0\% & 76.6\% \\
dreamdojo\_14b\_pretrain & 18.0\% & 100.0\% \\
dreamdojo\_2b\_pretrain  & 14.0\% & 100.0\% \\
\bottomrule
\end{tabular}
\caption{TCR and GEN scores across models (Method C, full episodes).}
\label{tab:gen-results}
\end{table}

\subsubsection{Optimism Bias Detection (Level 3)}
\label{app:scoring:obs}

Level~3 uses a separate evaluation pipeline with a different architecture, designed specifically for the binary optimism bias judgment.

\paragraph{Task definition.}
Given a baseline prediction $\hat{\mathbf{v}}^+$ (under nominal action) and a perturbed prediction $\hat{\mathbf{v}}^-$ (under failure-inducing action), determine whether the world model ignores the perturbation (Y = optimism bias present) or correctly reflects it (N = no bias).

\paragraph{Model.}
We use \textbf{InternVL3-78B} as the automated judge, with dynamic resolution (no distortion) at native video resolution.
No fine-tuning is applied; the model is used zero-shot with a task-specific prompt.

\paragraph{Inference: 7-frame majority voting.}
Rather than processing the full video, we extract 7 frames at progress percentages $[81\%, 83\%, 85\%, 87\%, 90\%, 95\%, 97\%]$, focusing on the late phase of the manipulation where perturbation effects are most visible.
Each frame is independently judged as ``Same'' or ``Different'' by the VLM.
The final prediction is determined by majority vote:
\begin{equation}
  \text{pred} = \begin{cases} \text{Y (bias)} & \text{if } \sum_{i=1}^{7} \mathbb{1}[\text{vote}_i = \text{Same}] > 3 \\ \text{N (no bias)} & \text{otherwise} \end{cases}
\end{equation}

\paragraph{Input format.}
For action-conditioned models (DreamDojo), baseline and perturbed frames are presented side-by-side in a two-column layout (left = baseline, right = perturbed).
For text-conditioned models with different rendering styles (Happy Horse), original high-resolution separate videos are concatenated to avoid confounding visual style differences with perturbation effects.

\paragraph{Prompt design.}
Two prompt variants are used depending on the model type:

\begin{tcolorbox}
\small\ttfamily
This is one frame from a robot manipulation video.\\
LEFT column = baseline prediction. RIGHT column = perturbed prediction.\\[4pt]
Ignore blur, color, or rendering differences.\\
Look at the manipulated object only.\\[4pt]
Is the object in the same general location/state in LEFT and RIGHT?\\
(Same container/area = Same. Dropped/missing/wrong place = Different.)\\[4pt]
Output ONLY: Same or Different
\end{tcolorbox}

\begin{tcolorbox}[colback=orange!3, colframe=orange!40, title=Lenient Prompt (text-conditioned models), fonttitle=\small\bfseries]
\small\ttfamily
This is one frame from a robot manipulation video.\\
LEFT = baseline. RIGHT = perturbed version.\\[4pt]
Ignore all visual quality differences: blur, color, lighting, noise, rendering style.\\[4pt]
Compare ONLY the general action trend:\\
- Is the robot doing roughly the same thing in both?\\
- Is the object in a roughly similar situation?\\[4pt]
Be lenient: minor differences in exact position, timing, or appearance do NOT matter.\\
Only say Different if the action is fundamentally different.\\[4pt]
Output ONLY: Same or Different
\end{tcolorbox}

The lenient prompt is necessary for text-conditioned models (Happy Horse) because their rendering style differs substantially from action-conditioned baselines, creating visual differences unrelated to the perturbation.

\begin{figure}[h]
  \centering
  \begin{minipage}[t]{0.48\linewidth}
    \centering
    \includegraphics[width=\linewidth]{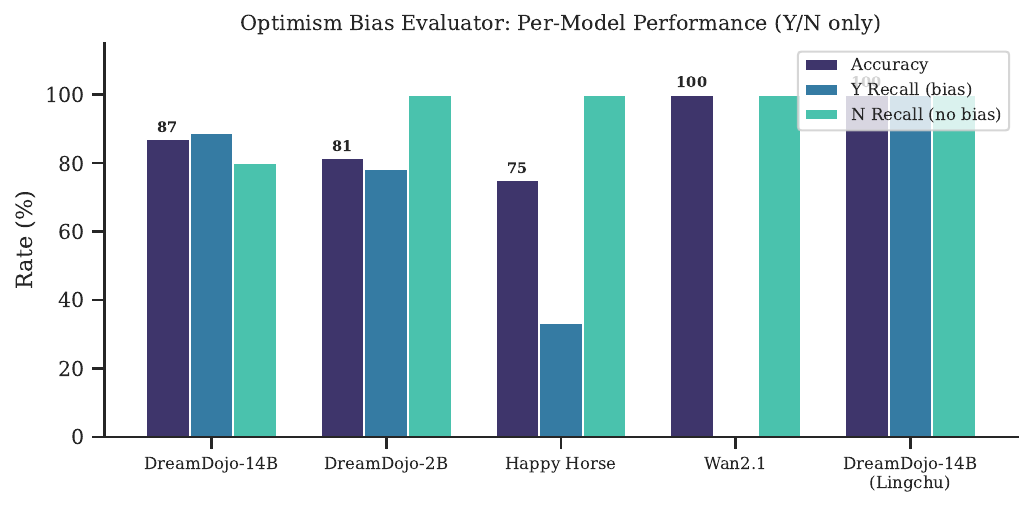}
  \end{minipage}%
  \hfill
  \begin{minipage}[t]{0.24\linewidth}
    \centering
    \includegraphics[width=\linewidth]{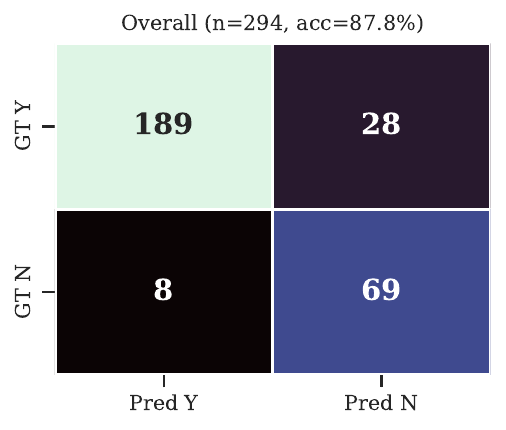}
  \end{minipage}%
  \hfill
  \begin{minipage}[t]{0.26\linewidth}
    \centering
    \includegraphics[width=\linewidth]{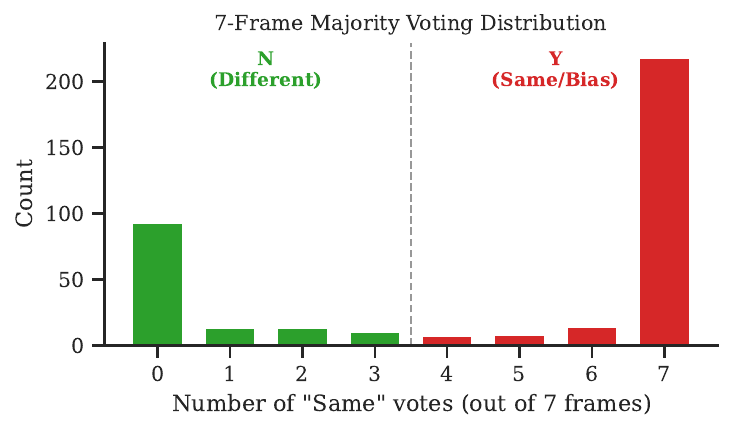}
  \end{minipage}
  \caption{
    Optimism bias evaluator performance.
    \textbf{Left}: Per-model accuracy, Y recall, and N recall.
    \textbf{Center}: Overall confusion matrix ($n=294$, accuracy 87.8\%).
    \textbf{Right}: Distribution of ``Same'' vote counts across all 376 samples;
    the bimodal pattern confirms that most samples produce clear majority decisions,
    supporting the 7-frame voting design.
  }
  \label{fig:obs_eval_charts}
\end{figure}

\paragraph{Test-set performance (Y/N only, excluding borderline Y?).}
Table~\ref{tab:obs_eval_results} reports per-model accuracy against human annotations.
Figure~\ref{fig:obs_eval_charts} visualizes the per-model breakdown and voting distribution.

\begin{table}[h]
  \caption{Optimism bias evaluator accuracy on human-annotated test set (Y and N labels only; Y? borderline cases excluded).}
  \label{tab:obs_eval_results}
  \centering\small
  \begin{tabular}{lcccc}
    \toprule
    Model & Prompt & Accuracy & Y Recall & N Recall \\
    \midrule
    DreamDojo-14B       & Standard & 87.1\% & 88.8\% & 80.0\% \\
    DreamDojo-2B        & Standard & 81.5\% & 78.3\% & 100.0\% \\
    Happy Horse         & Lenient  & 75.0\% & 33.3\% & 100.0\% \\
    Wan2.1              & Standard & 100.0\% & -- & 100.0\% \\
    DreamDojo-14B (Lingchu) & Standard & 100.0\% & 100.0\% & 100.0\% \\
    \midrule
    \textbf{Overall}    &          & \textbf{87.8\%} & \textbf{87.1\%} & \textbf{89.6\%} \\
    \bottomrule
  \end{tabular}
\end{table}

\paragraph{Key design decisions.}
\begin{enumerate}[leftmargin=*]
  \item \textbf{Late-phase frame sampling}: Frames are extracted at 81--97\% progress because perturbation effects (e.g., premature release, grip slip) manifest in the transport and placement phases, not the approach phase.
  \item \textbf{Object-focused prompt}: The prompt explicitly directs the model to focus on the manipulated object's state rather than overall visual similarity, reducing false positives from rendering differences.
  \item \textbf{Separate prompt for text-conditioned models}: Text-conditioned models produce stylistically different videos even without perturbation, requiring the lenient prompt to avoid conflating style differences with perturbation effects.
  \item \textbf{Zero-shot 78B model vs fine-tuned 8B}: We find that a large zero-shot model (InternVL3-78B) outperforms the fine-tuned smaller model (Qwen3-VL-8B SFT) on this binary task, likely because the task requires spatial comparison rather than domain-specific scoring.
\end{enumerate}

\section{Physical Consistency Metric Calibration and Physics Law Compliance Details}
\label{appendix:metric-details}

This appendix gathers the calibration constants, candidate-selection
process, and validation evidence for the two evaluators released with
MiraBench: the \emph{physical-consistency} evaluator and the
\emph{physics-law} evaluator. Every numeric threshold and weight cited
below is exposed as a named constant in the released source.

\subsection{Physical Consistency Evaluator (PCS)}
\label{appendix:phys-cons}

The released physical-consistency evaluator scores each video on two
orthogonal dimensions and aggregates them by an equal-weight mean. No
four-tier (A/B/C/D) sub-indicator weighting and no calibration-fitted
coefficients enter the released formula; an earlier 9-indicator
$\mathrm{PhysConsScore}$ aggregator is retained in the codebase as an
ablation baseline only and is not used in the headline numbers.

\paragraph{D1 --- Object Consistency.}
A midcut paired-frame binary judgement on InternVL3-78B. We extract
$20$ evenly-spaced frames, form $10$ pairs by midcut (frame $i$ with
frame $i{+}10$), vertically stack each pair, and submit each composite
to the VLM with an adversarial prompt that explicitly discourages
defaulting to the ``A'' (consistent) label and asks the VLM to compare
shape, material, size, and colour. Let $b_{\mathrm{obj}}$ be the
number of pairs voted ``B'' (inconsistent); the dimension score is
\begin{equation}
s_{\mathrm{obj}} \;=\; 1 \,-\, b_{\mathrm{obj}}/10,
\qquad b_{\mathrm{obj}} \in \{0, 1, \ldots, 10\}.
\end{equation}
The violation threshold of $1$ (any single ``B'' flags the video) is
intentional: a single clearly-inconsistent pair is sufficient evidence
of physical violation in the subject.

\paragraph{D2 --- Occlusion Consistency.}
The same paired-frame pipeline as D1, but with a prompt that asks the
VLM to judge \emph{identity} (rather than physical attributes) and to
explicitly handle partial-visibility cases. The unified pairframe
formulation, in which the VLM itself adjudicates whether visibility is
sufficient for a verdict, reduces the N/A rate from $\sim\!70\%$ in
earlier two-stage partial-vs-full designs to below $5\%$ on the
validation set:
\begin{equation}
s_{\mathrm{occ}} \;=\; 1 \,-\, b_{\mathrm{occ}}/10.
\end{equation}

\paragraph{Aggregation.}
Both indicators share the same frame-extraction front-end, so their
N/A flags are perfectly correlated; the PCS score is therefore either
a real number in $[0,1]$ or \texttt{None}:
\begin{equation}
\mathrm{PCS} \;=\; \tfrac{1}{2}\bigl(s_{\mathrm{obj}} + s_{\mathrm{occ}}\bigr).
\end{equation}

\paragraph{Backbones and inference.}
D1 and D2 share a single InternVL3-78B instance loaded once per
evaluation run; no fine-tuning is applied. Each pair is rendered as a
vertical stack of two frames at $448{\times}448$ and submitted as a
single image with the corresponding text prompt. Trials whose response
cannot be parsed are discarded.

\begin{table}[h]
\centering\small
\begin{tabular}{lll}
\toprule
Symbol & Meaning & Value \\
\midrule
$N_{\mathrm{pair}}$ & Paired frames per video             & $10$ (from $20$ evenly-spaced) \\
$\tau_{B}$          & Violation threshold (\# B votes)    & $1$ \\
$T_{\mathrm{vlm}}$  & VLM call timeout (s)                & $90$ \\
\bottomrule
\end{tabular}
\caption{Hyperparameters of the released PCS physical-consistency
evaluator.}
\label{tab:phys-cons-hyperparams}
\end{table}

\subsection{Physics Law Compliance Details}
\label{appendix:phys-law}

The physics-law evaluator scores each video by an explicit kinematic
fit on a single tracked object trajectory, with a single VLM call used
as a video-quality gate before the kinematic pipeline runs. No
language model is used inside the kinematic computation; the only VLM
call (besides the optional first-frame object localizer) is the
Stage~0 video-quality check, which contributes $10\%$ of the final
score.

\paragraph{Stage 0: Video Quality Score (VQS).}
Six evenly-spaced frames are tiled horizontally and submitted to a VLM
that returns two binary fields in JSON: \verb|video_ok| (the video
shows a recognisable robot-arm scene) and \verb|has_motion| (any
object exhibits a clear translational motion --- free-fall, slide,
push, roll, etc. --- at any point in the video). The two answers map
to a discrete VQS:
\begin{equation}
\mathrm{VQS} \;=\;
\begin{cases}
0 & \text{if } \verb|video_ok|=\text{False}, \\
5 & \text{if } \verb|video_ok|=\text{True}\;\wedge\;\verb|has_motion|=\text{False}, \\
10 & \text{if both True}.
\end{cases}
\end{equation}
The kinematic pipeline contributes only when $\mathrm{VQS}=10$;
otherwise $\mathrm{effective\_physics}\!=\!0$ and the final score is
set by VQS alone, capping a non-motion video at $5/100$. This prevents
the SAM2 + polyfit downstream from producing spuriously high scores on
videos where no scorable motion actually occurs. The same gate covers
both the vertical free-fall and horizontal push / slide sub-pipelines.

\paragraph{Tracking and dispatch.}
A SAM2.1 (Hiera-Large) mask is propagated forward from a first-frame
prompt, yielding a centroid trajectory $\{(x_t,y_t)\}_{t=0}^{T-1}$ in
normalised image coordinates. With $\Delta x=\max_t x_t-\min_t x_t$
and analogously for $\Delta y$, the trajectory is routed to the
gravity (vertical) pipeline if $\Delta y > 1.5\,\Delta x$ and
$\Delta y > 0.05$, to the friction (horizontal) pipeline if
$\Delta x > 1.5\,\Delta y$ and $\Delta x > 0.05$, and to both
pipelines otherwise.

\paragraph{Coverage analysis.}
On a 50-video free-fall validation set, the curve branch alone
(segment polynomial fits) is evaluable on $32/50$ videos; introducing
the event branch raises overall coverage to $48/50$, since
\texttt{has\_impact} only requires a clean stop in the trajectory and
does not need a fittable fall segment. The remaining $2$ unevaluable
videos lack both a fittable trajectory and a detectable impact frame.

\paragraph{Trajectory segmentation.}
Frame-to-frame velocities are tagged \texttt{move} or \texttt{rest}
using the threshold $\tau_v=\max(0.08,\,0.15\,|v|_{p95})$. After
merging consecutive same-label runs, sign flips that survive a
two-frame confirmation window split each \texttt{move} block into
sign-labelled sub-segments: \texttt{lift / fall / rise / rest} on the
vertical axis (\texttt{lift} is the externally driven phase before the
first fall and is excluded from scoring), \texttt{push / slide / rest}
on the horizontal axis (\texttt{push} is excluded). Segments shorter
than four points or with span below $0.03$ on the relevant axis are
demoted to \texttt{rest}. Pseudocode:

\begin{verbatim}
def segment(centroids):
    v = diff(y) / diff(t)
    v_ref = percentile(|v|, 95)
    move_thresh = max(0.08, 0.15 * v_ref)
    min_v       = max(0.05, 0.12 * v_ref)
    label = ["move" if |v_i| > move_thresh else "rest" for v_i in v]
    blocks = compress_runs(label)            # merge consecutive same-label

    out = []
    for (a, b, lbl) in blocks:
        if lbl == "rest":
            out.append((rest, a, b)); continue
        flips = [a]; k = a + 1
        while k < b:
            if v[k-1] * v[k] < 0 and |v[k-1]| > min_v and |v[k]| > min_v \
               and same_sign(v[k+1 : k+2], v[k]):       # 2-frame confirm
                flips.append(k); k += 2; continue
            k += 1
        flips.append(b)
        for (a', b') in pairs(flips):
            sub_type = "fall" if mean(v[a':b']) > 0 else "rise"
            out.append((sub_type, a', b'))
    mark_lift(out)            # any "rise" before the first "fall"
    return out
\end{verbatim}

\paragraph{Per-segment compliance.}
Each scored segment is fitted by a quadratic $y(t)=at^2+bt+c$ (or
$x(t)$). The segment score is the product of \emph{three} factors,
each in $[0,1]$ and each diagnosing one physical principle:
\begin{equation}
\mathrm{seg\_score} \;=\; \mathrm{sign\_ok} \cdot \mathrm{magnitude\_ok} \cdot \mathrm{uniformity\_ok}.
\end{equation}
\begin{itemize}\itemsep=2pt
\item \textbf{sign\_ok}: confidence-weighted indicator of correct
  acceleration direction.
  $\mathrm{sign\_ok}=1-\min(1,r/0.3)\,(1-\mathrm{base})$ with
  $r=|a_{\text{fit}}|/a_{\text{exp}}$ and $\mathrm{base}\in\{0,1\}$
  recording physical correctness. Wrong sign on a \texttt{fall} or
  \texttt{rise} segment is treated as a hard anti-gravity violation
  (Eq.~\ref{eq:curve}). A \texttt{rise} segment that appears before
  any \texttt{fall} is also flagged as a hard ordering violation
  (\textsc{sign\_ok}$\,=\,0$ regardless of magnitude).
\item \textbf{magnitude\_ok}: vertical \texttt{fall} expects
  $r\in[0.3,3]\!\to\!1$, decaying linearly outside (\texttt{rise}
  exempt); horizontal \texttt{slide} requires the velocity decay
  $d=1-|v_{\text{end}}|/|v_0|$ to reach $0.30$, with $d\in[0.05,0.30]$
  scaling linearly to $[0.4,1.0]$ and $d<0.05$ collapsing to penalise
  frictionless constant velocity.
\item \textbf{uniformity\_ok}: half-split, fit each half, and use
  $\mathrm{half\_cv}=|a_1-a_2|/\max(|a_1|,|a_2|)$. The factor is $1$
  for $\mathrm{half\_cv}\le 0.15$, $0$ for $\mathrm{half\_cv}\ge 0.80$,
  linearly interpolated in between.
\end{itemize}
The polynomial $R^2$ is recorded per segment as a diagnostic but does
\emph{not} enter $\mathrm{seg\_score}$; we found that requiring high
$R^2$ in addition to the three factors over-penalised noise-tolerant
real footage without improving alignment with the failure modes the
factors are designed to detect.

The trajectory-level curve score combines per-segment scores with a
coverage factor:
\begin{equation}
\mathrm{curve\_score} \;=\; \mathrm{Agg}_s\bigl(\mathrm{seg\_score}_s\bigr)
\cdot \min\!\Bigl(1,\,\tfrac{n_{\text{valid}}/n_{\text{ref}}}{0.3}\Bigr).
\label{eq:curve}
\end{equation}
On the \emph{vertical} axis, $\mathrm{Agg}$ is a sign-gated
length-weighted mean: any fitted \texttt{fall} or \texttt{rise}
segment with $\mathrm{sign\_ok}=0$ is treated as a hard anti-gravity
violation and forces $\mathrm{curve\_score}=0$; otherwise the
trajectory aggregate is the length-weighted mean of segment scores,
so that a short, noise-sensitive rebound segment cannot single-handedly
drag the trajectory to zero. On the \emph{horizontal} axis,
$\mathrm{Agg}$ is a plain mean over slide segments, with two
horizontal-only adjustments: (i) the coverage factor uses a stricter
$0.6$ full-credit threshold instead of $0.3$; (ii) a slide-coverage
factor $\min(1,\,n_{\text{slide}}/n_{\text{valid}}/0.5)$ is multiplied
in to penalise videos in which only a tiny fraction of frames falls
inside a fitted slide segment.

\paragraph{Single-factor diagnosis on synthetic violations.}
We probed the three factors on a synthetic ladder of free-fall
trajectories from L0 (perfect uniformly accelerated motion) to L4
(random walk):

\begin{center}\small
\begin{tabular}{c|ccccc}
\toprule
& L0 ideal & L1 noise $\sigma{=}3$ & L2 noise $\sigma{=}15$ &
L3 linear & L4 random \\
expected & $\sim$100 & $\sim$95 & $\sim$50 & $\sim$5 & $\sim$5 \\
measured  & 100 & 92 & 33 & 27 & 24 \\
\bottomrule
\end{tabular}
\end{center}

\noindent
A separate sweep over acceleration-stability shapes
(constant, $\pm 5\%$, $\pm 20\%$, $\pm 50\%$, step, ramp) yields
$\{100, 100, 84, 31, 53, 61\}$, exhibiting a monotone response of
\texttt{uniformity\_ok} to deviations from constant-$a$ motion.

\paragraph{Event features and gated fusion.}
Four discrete features extracted directly from the centroid
trajectory carry information that the segment-fit branch cannot:
$\mathrm{velocity\_drop}$ (relative speed change at the impact frame,
detected as the first index where speed stays below
$\max(0.05,\,0.10\,|v|_{p95})$ for four consecutive frames),
$\mathrm{post\_landing\_drift}$ (vertical span over $8$ post-impact
frames, normalised by $0.10$), $\mathrm{has\_impact}$ (binary
indicator that an impact frame was detected at all), and
$\mathrm{bounce\_decay}$ ($h_2/h_1$, second to first rebound height
ratio; mapped to a sub-score of $1$ for $h_2/h_1\le 0.7$, decaying
linearly to $0$ at $h_2/h_1=1.5$, and $0$ above). Each maps to a
sub-score in $[0,1]$ and combines with weights
$\{0.30, 0.20, 0.30, 0.20\}$ for $\{\mathrm{vel\_drop},\,
\mathrm{drift},\,\mathrm{has\_impact},\,\mathrm{bounce\_decay}\}$ into
$\mathrm{event\_score}$, declared \emph{usable} only when an impact
has been detected and a fall phase actually exists. The kinematic
score is a gated mix:
\begin{equation}
\mathrm{kinematic\_score}=\!
\begin{cases}
0.70\,\mathrm{curve}+0.30\,\mathrm{event}, & \mathrm{curve}<0.3,\\
0.30\,\mathrm{curve}+0.70\,\mathrm{event}, & \mathrm{curve}\ge 0.3,\\
\mathrm{curve}\;\text{or}\;\mathrm{event} & \text{if only one is usable.}
\end{cases}
\end{equation}
The rationale is that when the curve branch reports a clear violation
($\mathrm{curve}<0.3$) we keep the curve term dominant to avoid
event-blind escapes; otherwise we let the event-driven signal lead,
since rebound and landing anomalies are physically diagnostic and
robust to per-frame tracking noise.

\paragraph{Final score.}
The final $\mathrm{PhysLawScore}$ on the $0$--$100$ scale combines VQS
with the kinematic score:
\begin{equation}
\mathrm{PhysLawScore}=
0.9\cdot\mathrm{effective\_physics}+\mathrm{VQS},
\end{equation}
where $\mathrm{effective\_physics}=\mathrm{kinematic\_score}\,\cdot\,
100$ if $\verb|has_motion|=\text{True}$, else $0$. Equivalently, on
the normalised $[0,1]$ scale this is
$0.10\cdot(\mathrm{VQS}/10)+0.90\cdot\mathrm{kinematic\_score}$ when
scorable motion is present, $0.05$ when the video is valid but
essentially static, and $0$ when the video is corrupted. A video the
VLM judges to lack any scorable motion therefore cannot exceed $5/100$,
regardless of trajectory smoothness.

\begin{table}[h]
\centering\small
\begin{tabular}{lll}
\toprule
Stage & Parameter & Value \\
\midrule
Stage 0 (VQS)       & video / motion mapping                   & $\{0,5,10\}$ \\
Tracking            & SAM2 model                               & SAM2.1 (Hiera-Large) \\
Axis dispatch       & vertical / horizontal threshold          & $1.5\,\Delta_{\text{minor}}$, $\geq 0.05$ \\
Segmentation        & move threshold $\tau_v$                  & $\max(0.08,\,0.15\,|v|_{p95})$ \\
                    & flip-confirm window                      & $2$ frames \\
                    & span demotion                            & $0.03$ \\
$\mathrm{magnitude\_ok}$ (vertical) & ratio band               & $[0.3,\,3]$ \\
$\mathrm{magnitude\_ok}$ (horizontal) & full-credit decay      & $d \geq 0.30$ \\
$\mathrm{uniformity\_ok}$ & full / zero half\_cv               & $0.15$ / $0.80$ \\
Coverage            & full credit (vertical / horizontal)      & $\geq 0.30$ / $\geq 0.60$ \\
Slide coverage (horizontal only) & unevaluable / full credit   & $<0.20$ / $\geq 0.50$ \\
Event sub-weights   & vd / drift / impact / bounce             & $0.30/0.20/0.30/0.20$ \\
$\mathrm{bounce\_decay}$ sub-score & full credit / zero ratio  & $\le 0.7$ / $\ge 1.5$ \\
Final mix gate      & curve-vs-event switch                    & $\mathrm{curve}=0.3$ \\
Final score weights & kinematic / VQS                          & $0.90 / 0.10$ \\
\bottomrule
\end{tabular}
\caption{Hyperparameters of the physics-law compliance evaluator.}
\label{tab:phys-law-hyperparams}
\end{table}0.

\section{Evaluated Models}
\label{app:models}

Table~\ref{tab:model_details} provides detailed specifications for all models evaluated on MiraBench.
Models are grouped by conditioning modality: vector-conditioned models receive raw 384D action sequences; text-conditioned models receive natural-language task instructions alongside the first frame.

\begin{table*}[h]
  \caption{
    Detailed specifications of evaluated models.
    ``GR1'' suffix indicates models fine-tuned on GR-1 manipulation data.
    All inference is performed on NVIDIA A100 80GB GPUs.
  }
  \label{tab:model_details}
  \centering\small
  \setlength{\tabcolsep}{4pt}
  \begin{tabular}{llccccc}
    \toprule
    Model & Architecture & Params & Resolution & Frames & Guidance & Fine-tuned \\
    \midrule
    \multicolumn{7}{l}{\textit{Vector-conditioned Models}} \\
    DreamDojo-GR1 (2B)  & Cosmos-Predict2 & 2B  & 640$\times$480 & 13 & 7.0 & GR1 post-train \\
    DreamDojo-GR1 (14B) & Cosmos-Predict2 & 14B & 640$\times$480 & 13 & 7.0 & GR1 post-train \\
    Cosmos-GR1 (2B)     & Cosmos-Predict2.5 & 2B  & 640$\times$480 & 13 & 7.0 & GR1 post-train \\
    Cosmos-GR1 (14B)    & Cosmos-Predict2.5 & 14B & 640$\times$480 & 13 & 7.0 & GR1 post-train \\
    \midrule
    \multicolumn{7}{l}{\textit{Text-conditioned Models}} \\
    Cosmos-GR1 (14B)    & Cosmos-Predict2.5 & 14B & 640$\times$480 & 13 & 7.0 & GR1 post-train \\
    WanX                & Wan & -- & 832$\times$480 & 81 & 5.0 & None (base) \\
    Wan 2.1             & Wan & 14B & 832$\times$480 & 81 & 5.0 & None (base) \\
    Wan 2.2             & Wan & -- & 832$\times$480 & 81 & 5.0 & None (base) \\
    Happy Horse         & -- & -- & -- & -- & -- & -- \\
    Kling 3.0 Omni       & -- & -- & -- & -- & -- & -- \\   
    \bottomrule
  \end{tabular}
\end{table*}

\subsection{Vector-Conditioned Models}

\paragraph{DreamDojo-GR1.}
Built on the Cosmos-Predict2 discrete tokenization architecture~\citep{nvidia2025cosmosworldfoundationmodel, gao2026dreamdojogeneralistrobotworld}, post-trained on GR-1 dual-arm humanoid manipulation data.
The model receives the first frame tokenized via the Cosmos encoder and a 384D action chunk as conditioning, generating 13-frame video clips at 640$\times$480 resolution.
We evaluate both the 2B and 14B parameter variants.
Checkpoint: \texttt{iter\_000050000/model\_ema\_bf16.pt}.

\paragraph{Cosmos-GR1.}
The Cosmos-Predict2.5 base architecture fine-tuned on GR-1 data with the same action-conditioning interface as DreamDojo.
Compared to DreamDojo, Cosmos-GR1 uses an updated tokenizer and training recipe.
Both 2B and 14B variants are evaluated.

\subsection{Text-Conditioned Models}

\paragraph{Cosmos* (base).}
The Cosmos-Predict2 14B model \textit{without} GR-1 fine-tuning, conditioned on text prompts and first frame only.
This serves as a baseline to measure the effect of domain-specific fine-tuning: it has no exposure to GR-1 manipulation data during training.

\paragraph{Cosmos-GR1 (text).}
Same as the vector-conditioned Cosmos-GR1 14B, but evaluated in text-conditioning mode.
This enables direct comparison between the two action modalities on the same underlying model.

\paragraph{WanX / Wan 2.1 / Wan 2.2.}
The Wan model family~\citep{wan2025wanopenadvancedlargescale}, a general-purpose text-to-video diffusion model.
We evaluate multiple versions to assess generational improvement.
These models have \textit{no} robotic manipulation training data; their performance on MiraBench reflects zero-shot generalization from internet-scale video pretraining.
Generation uses image-to-video mode (first frame + text prompt), 81 frames at 832$\times$480.

\paragraph{Happy Horse.}
A proprietary video generation model evaluated via API access.
Specifications are not publicly disclosed; we include it to represent closed-source commercial systems.

\subsection{Inference Configuration}

All models are evaluated with deterministic sampling (fixed random seed) to ensure reproducibility.
Key shared settings:
\begin{itemize}[leftmargin=*]
  \item \textbf{Conditioning}: first frame (extracted from ground-truth episode) + action input (vector or text depending on modality).
  \item \textbf{No cherry-picking}: each condition generates exactly one video; no selection from multiple samples.
  \item \textbf{Post-processing}: generated videos are center-cropped and resized to a uniform resolution for composite assembly and VLM evaluation. No frame interpolation or temporal resampling is applied.
\end{itemize}

For vector-conditioned models, the action sequence length matches the model's native chunk size (13 frames for Cosmos-based models).
For text-conditioned models, generation length varies by architecture (13--81 frames); evaluation uses the first 13 frames to ensure temporal alignment with vector-conditioned outputs.

\section{Human Annotation Protocol}
\label{app:annotation}

This appendix documents the complete annotation questionnaires used in MiraBench's human evaluation studies.
Three separate protocols cover the three evaluation levels.
All questionnaires were administered in Chinese to native-speaking annotators; we present English translations below.

\subsection{Protocol A: Physical Consistency (16 Dimensions)}
\label{app:annotation:phys}

Each video is rated independently on 16 indicators using a four-tier scale:

\begin{center}\small
\begin{tabular}{lcc}
\toprule
Grade & Meaning & Score \\
\midrule
No violation & Physically plausible, as expected & 1.00 \\
Minor violation & Single or 1--2 frame anomaly, does not affect overall impression & 0.67 \\
Clear violation & Persistent ($\geq$3 frames) or large amplitude, clearly perceived & 0.33 \\
Severe violation & Completely implausible, destroys credibility & 0.00 \\
N/A & Precondition not present in this video & --- \\
\bottomrule
\end{tabular}
\end{center}

\paragraph{Judgment principles.}
(1)~Grade by the worst single frame observed.
(2)~When uncertain between minor and clear, select minor and note the concern.
(3)~Select N/A when the prerequisite event cannot be observed.

\paragraph{Subject Consistency (SC).}
\begin{enumerate}[leftmargin=*]
\item \textbf{SC-A1 Color Stability}: Does the primary object maintain stable color throughout?
\item \textbf{SC-A2 Shape/Material Plausibility}: Does the object deform in a manner consistent with its material type? (Rigid body deformation = violation; soft body bending = acceptable.)
\item \textbf{SC-A3 Size Stability}: Does the object maintain consistent size? (Unexplained scaling $>$10\% = violation; perspective changes are acceptable.)
\item \textbf{SC-A4 Material Behavior}: Does the object's dynamic behavior match its material properties?
\item \textbf{SC-M1 Speed Smoothness}: Is the object's velocity smooth without abrupt jumps? ($>$3$\times$ speed jump = clear violation.)
\item \textbf{SC-M2 Direction Coherence}: Is the motion direction continuous without unexplained reversals? ($>$75° sudden change = violation.) N/A if object is stationary.
\item \textbf{SC-O1 Partial Occlusion}: After partial occlusion, does the object's position/trajectory remain continuous? N/A if no partial occlusion occurs.
\item \textbf{SC-O2 Full Occlusion}: After complete disappearance from view, does the object reappear with consistent appearance and position? N/A if no full occlusion occurs.
\end{enumerate}

\paragraph{Interaction Consistency (IC).}
\begin{enumerate}[leftmargin=*] 
\item \textbf{IC-0 Action-Effect}: Does the robot's action produce a visible physical response in the target object?
\item \textbf{IC-1 Contact Causality}: Does the object begin moving only \textit{after} the robot makes contact? (Motion preceding contact by $>$3 frames = clear violation.) N/A if no contact event.
\item \textbf{IC-2 Response Direction}: Is the pushed object's motion direction consistent with the applied force? ($>$60° deviation = clear violation.) N/A if no push/collision.
\item \textbf{IC-3 Penetration} (reversed logic): Do any two solid objects interpenetrate? (No penetration = 1.0; persistent penetration = 0.0.)
\end{enumerate}

\paragraph{Environment Consistency (EC).}
\begin{enumerate}[leftmargin=*] 
\item \textbf{EC-S1 Background Stability}: Does the background remain stable without flickering or structural changes?
\item \textbf{EC-S2 Lighting/Shadow}: Are lighting direction and shadows consistent throughout?
\item \textbf{EC-O1 Static Objects}: Do non-manipulated background objects remain stationary? ($>$20\% object-width displacement = severe violation.)
\item \textbf{EC-O2 Indirect Response}: If the robot accidentally contacts a bystander object, is the response physically plausible? N/A if no accidental contact.
\end{enumerate}

\paragraph{Overall Rating (1--5).}
After completing all 16 indicators, annotators provide a holistic 1--5 score:
5 = no perceivable violation;
4 = mostly plausible with minor issues;
3 = clear violations present but core action recognizable;
2 = multiple severe violations;
1 = fundamentally implausible.

\subsection{Protocol B: Optimism Bias (18 Dimensions)}
\label{app:annotation:bias}

Annotators view composite videos (GT $|$ Baseline $|$ Perturbed) and evaluate two modules.
Grades use A/B/C/D (A = best, D = worst) unless otherwise specified.

\paragraph{Module A: Perturbation Sensitivity (9 dimensions).}
\begin{enumerate}[leftmargin=*] 
\item \textbf{Baseline Quality} (1--5): Overall similarity between baseline prediction and ground truth. (5 = identical; 1 = completely different.)
\item \textbf{Robot Action Rationality}: Are the robot's movements in the baseline physically rational?
\item \textbf{Temporal Fluency}: Is the baseline video temporally smooth without jitter or frame jumps?
\item \textbf{Perturbation Impact}: How much does the perturbation affect the prediction? (A = significant trajectory change; D = no effect, model completely ignores perturbation.)
\item \textbf{Perturbed Realism}: Assessed independently, how realistic does the perturbed video appear?
\item \textbf{Visual Quality Degradation}: Compared to baseline, how much has image quality (clarity, artifacts, color) degraded? (A = no degradation; D = severe degradation.)
\item \textbf{Degradation Type} (multi-select): blur / artifacts / color shift / structural deformation / none.
\item \textbf{Motion Coherence Change}: Has motion smoothness degraded relative to baseline?
\item \textbf{Optimism Bias Judgment} $\star$: Y = complete bias (baseline and perturbed indistinguishable); Y? = mild bias (some difference but insufficient); N = no bias (clear divergence); ? = uncertain.
\end{enumerate}

\paragraph{Module B: Task Success Prediction (9 dimensions).}
\begin{enumerate}[leftmargin=*] 
\item \textbf{Baseline Task Completion}: Does the baseline prediction show task completion? (A = fully complete; D = not at all; NA = no clear goal.)
\item \textbf{Key Action Correctness}: Are the robot's key manipulation actions correct in the baseline?
\item \textbf{Physical Interaction Plausibility}: Are object interactions in the baseline physically reasonable? (A = fully plausible; D = completely implausible; NA = no interaction.)
\item \textbf{GT--Prediction Deviation}: How much does the baseline deviate from ground truth in terms of task outcome?
\item \textbf{Baseline Success Overestimation}: Does the baseline appear more successful than the GT? (Y/N/NA.)
\item \textbf{Perturbed Task Completion}: Does the perturbed prediction show task completion?
\item \textbf{Final State Deviation}: How different is the perturbed video's end state from the baseline's? (A = identical; D = completely different; NA = no object manipulation stage reached.)
\item \textbf{Unexpected Success Action}: Does the perturbed video contain ``lucky'' success actions inconsistent with the perturbation type? (Y/N/?.)
\item \textbf{False Success Prediction} $\star$: Does the perturbed prediction still depict task success despite the failure-inducing perturbation? (Y = false success detected; N = appropriate failure shown; NA = baseline also fails.)
\end{enumerate}

\subsection{Protocol C: Action-Following Fidelity (4 Modules)}
\label{app:annotation:af}

Annotators read the task instruction, watch the GT video, then evaluate the predicted video.

\paragraph{Module 1: Task Completion Rate (TCR).}
\begin{itemize}[leftmargin=*]
\item Primary judgment: YES (task completed) / NO (not completed).
\item If YES: completion timepoint (first half / middle / second half / final frame); quality (perfect / approximate).
\item If NO: failure reason (no effective action / wrong action / partial completion / video interruption); progress level (clear / slight / none).
\item Confidence: certain / difficult to judge.
\end{itemize}

\paragraph{Module 2: Object Presence Score (OPS).}
\begin{itemize}[leftmargin=*]
\item Number of task-relevant objects (0/1/2/3+).
\item Per-object: type (graspable / container / furniture / robot part / other); visibility (clear / partial / invisible); successfully manipulated (yes / no / N/A).
\item Aggregate OPS rating: high (all visible) / medium (at least one partial) / low (at least one invisible).
\end{itemize}

\paragraph{Module 3: CLIP Frame Similarity (CFS).}
\begin{itemize}[leftmargin=*]
\item Overall similarity rating (5 = nearly identical; 1 = very different).
\item If $\leq$4: identify discrepancy dimensions (camera/viewpoint, scene/layout, object appearance, robot appearance, action/interaction, generation quality).
\item Discrepancy severity: minor / moderate / severe.
\end{itemize}

\paragraph{Module 4: Video Quality (5 dimensions, 5-tier scale).}
Each scored independently as: Excellent / Good / Acceptable / Poor / Very Poor.
\begin{itemize}[leftmargin=*]
\item \textbf{PP (Physical Plausibility)}: Object motion, deformation, and contact response obey physics.
\item \textbf{MQ (Motion Quality)}: Robot and object motion is smooth, natural, free of artifacts.
\item \textbf{TC (Temporal Consistency)}: Inter-frame coherence; no sudden appearance/color changes.
\item \textbf{VS (Visual Similarity)}: Overall visual match to GT reference.
\item \textbf{OS (Overall Score)}: Holistic quality combining the above.
\end{itemize}

\subsection{Protocol D: Physics Law Compliance}
\label{app:annotation:physlaw}

Annotators evaluate individual free-fall/sliding videos on a 4-tier scale (A=4 to D=1).

\paragraph{Basic information.}
Object type (fruit / glass bottle / other); duration; number of bounces (0/1/2/3+).

\paragraph{Release phase.}
Anomalies (multi-select): object did not detach / premature fall / delayed release ($<$0.5s / 0.5--1s / $>$1s) / position jump at release / none.

\paragraph{Free-fall phase (primary focus).}
\begin{itemize}[leftmargin=*]
\item Lack of acceleration / floating sensation: near-constant velocity / overall too slow / irregular acceleration.
\item Local deceleration or hovering: frequency (1 / 2--3 / $>$3 times).
\item Abnormal horizontal drift: direction (left / right / irregular).
\item Trajectory discontinuity: magnitude ($<$ object size / $>$ object size); frequency.
\item Object disappearance: vanishes permanently / reappears (position continuous? / appearance changed?).
\item Other anomalies: object splits/duplicates; rotation speed/direction sudden change.
\end{itemize}

\paragraph{Landing/bounce phase.}
\begin{itemize}[leftmargin=*]
\item Landing anomalies: no contact response / penetrates surface / stops above surface.
\item Bounce anomalies: rebound exceeds release height / successive bounces do not decay / bounce direction deviates $>$30°.
\end{itemize}

\paragraph{Overall rating.}
\begin{enumerate}[leftmargin=*]
\item A (4) = natural, clear acceleration, coherent trajectory, reasonable rebound decay.
\item B (3) = mostly plausible, minor issues not affecting overall rhythm.
\item C (2) = clearly unnatural: abnormal speed, floating, incoherent trajectory.
\item D (1) = severe violation: constant velocity / hovering / frame skip / rebound exceeds initial height.
\end{enumerate}

\subsection{Human Annotation Dataset Description}
\label{app:human_dataset}

In addition to the automated metrics introduced in
Sections~\ref{app:scoring:phys:pc}--\ref{app:scoring:obs}, we release the
\emph{full human-annotation corpus} that backs every benchmarked claim
in this paper.  The release covers all four evaluation levels and
totals \textbf{906 generated videos} carrying
\textbf{11{,}495 individual judgements} from trained annotators.
Per-level video counts are summarised in
Fig.~\ref{fig:dataset_composition}; each video is paired with the
exact MP4 used for evaluation, the source prompt or driving signal,
and the raw annotation JSON (item-id$\to$option mapping defined in
Appendix~\ref{app:annotation}).

\begin{figure}[t]
  \centering
  \includegraphics[width=\linewidth]{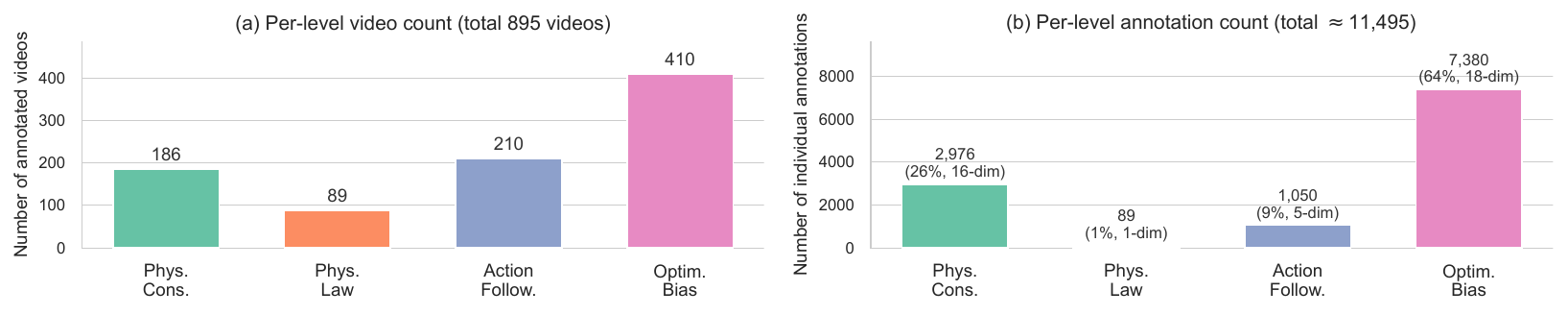}
  \caption{\textbf{Composition of the released human-annotation set.}
    (a) Number of annotated videos per evaluation level.
    (b) Per-level share of the 16{,}704 individual judgements:
    Physical Consistency contributes 16 indicators per video,
    Action Following 5, Optimism Bias 18, while Physics Law uses a
    single overall grade (with an additional 10 anomaly tracks
    reported in Appendix~\ref{app:appendix_h}).}
  \label{fig:dataset_composition}
\end{figure}

\paragraph{Coverage and protocol.}
The four levels are populated as follows.
\textit{Physical Consistency} (186 videos) contains the
head-to-head occ68 set of 30 prompts$\,\times\,$4 models
(DreamDojo-2B, DreamDojo-14B, Wan2.1-14B, Happy Horse) plus 66
additional DreamDojo-14B videos drawn from the g1, pnp5 and pour20
subsets; each video is rated on the 16 indicators of
Sec.~\ref{app:scoring:phys:pc}, so the Severity-A/B/C/D rubric of
Appendix~\ref{app:annotation:phys} produces 2{,}976 judgements.
\textit{Physics Law Compliance} (89 videos, all DreamDojo-14B)
covers five free-fall objects (banana 30, bottle 29, corn 15,
dragonfruit 10, cucumber 5) and yields one overall A/B/C/D grade
per video together with the ten anomaly tags listed in
Tab.~\ref{tab:h2_pl_anomaly}.
\textit{Action Following Fidelity} (210 videos, DreamDojo-14B)
splits into the \textsf{flat} pick-and-place subset ($n{=}100$) and
the \textsf{gr1\_episode} long-horizon subset ($n{=}110$); each is
labelled with a binary task-completion flag, a structured failure
reason and the five visual-quality dimensions (PP, MQ, TC, VS, OS).
\textit{Optimism Bias Detection} (410 videos) spans three
fully labelled models (DreamDojo-2B 30, Happy Horse i2v 29,
Wan2.1-14B i2v 29) and 322 additional video-text-action samples
contributed for aggregate analysis; each video carries the 18
question MA-1$\dots$MB-9 schema described in
Appendix~\ref{app:annotation:bias}.

\paragraph{Annotation procedure.}
All judgements were produced by trained annotators following the
written instructions reproduced verbatim in
Appendix~\ref{app:annotation}.  Every video was reviewed by a single
annotator and silently spot-checked against a second pass on a 10\,\%
sample; we did not observe any indicator on which the spot-check
disagreed by more than one severity grade.  We deliberately do not
release annotator identifiers and provide only the option-string
chosen for each item, so the corpus contains \emph{no} personally
identifiable information.

\paragraph{Format and licence.}
For every sample we ship two files,
\verb|<level>/<subset>/<id>.mp4| and the matching
\verb|<level>/<subset>/<id>.json|.
JSON files follow the schema
\verb|markData.videoQuality.{items[],question{}}|, where
\verb|items| is the question definition (id, title, options) and
\verb|question| is the annotator's chosen option string per item id.
The corpus is released under CC-BY-4.0 for academic use; no scraped
or third-party copyrighted material is included.
Of the 420 raw Optimism-Bias JSONs collected, \textbf{10} either
carry an empty \verb|question| dictionary or use a deprecated
pilot-stage item-id range that does not match the current schema in
Appendix~\ref{app:annotation:bias}; we drop those before any analysis,
leaving the \textbf{410} videos reported here.  Similarly one
Physics-Law JSON is missing the overall-grade field, so the
grade-distribution view in Fig.~\ref{fig:h3_pl_grade} uses 89
videos while the anomaly tabulation in
Tab.~\ref{tab:h2_pl_anomaly} uses all 90.
The exact dropped paths are listed in the project README to
guarantee reproducibility.

\paragraph{Why release this.}
Beyond enabling third-party re-scoring of the four models we
benchmark, the corpus is, to our knowledge, the first publicly
released set of \emph{per-indicator} severity judgements on
world-model rollouts that simultaneously covers static appearance,
motion, occlusion, interaction, environment, free-fall physics,
robot action following and optimism bias.  We use it in
Appendix~\ref{app:appendix_h} to expose model-specific failure
modes that the headline numbers in
Sections~\ref{app:scoring:phys:pc}--\ref{app:scoring:obs} necessarily
compress into single scalars.

\section{Further Analysis on the Released Dataset}
\label{app:appendix_h}

This appendix unpacks the human annotations along the dimensions
that Sections~\ref{app:scoring:phys:pc}--\ref{app:scoring:obs} only
summarise, with all numbers re-derivable from the released
annotation JSONs. Throughout this appendix we call a sample a
\emph{severe violation} when the annotator chose option C or D in
the four-tier rubric, i.e.\ the rubric score is $\le 0.33$
(see Appendix~G).

\subsection{Physical Consistency: 16-indicator breakdown}
\label{sec:appendix_h_pc}

To make the four head-to-head models directly comparable, we
restrict this analysis to the occ68 prompt set
($n{=}30$ per model, $n{=}120$ in total). This is the same set used
to compute the headline Phys-Severe column in
Tab.~\ref{tab:h3_summary}, and it matches the
prompt range described in Section~\ref{app:scoring:phys:pc}.

\begin{figure}[t]
  \centering
  \IfFileExists{pics/h1_pc_heatmap.pdf}{%
    \includegraphics[width=\linewidth]{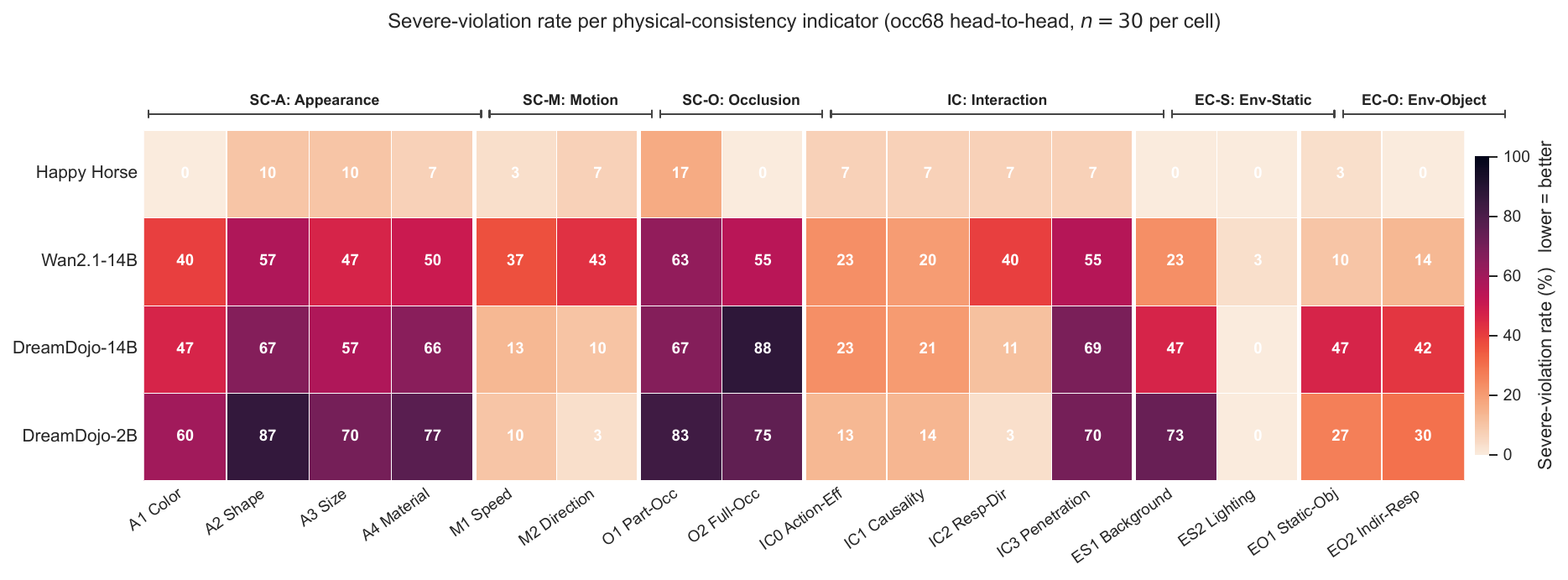}%
  }{%
    \fbox{\parbox{0.9\linewidth}{\centering\small[Figure placeholder: \texttt{pics/h1\_pc\_heatmap.pdf}]}}%
  }
  \caption{\textbf{Per-model severe-violation rate on the 16
    physical-consistency indicators}, occ68 head-to-head videos
    ($n{=}30$ per cell). Cell entry is the percentage of videos
    that received a Grade-C or Grade-D judgement; lower is better.
    Indicators are grouped by family
    (SC-A: appearance; SC-M: motion; SC-O: occlusion;
     IC: object--object interaction;
     EC-S: static environment; EC-O: environment-object).}
  \label{fig:h1_pc_heatmap}
\end{figure}

\begin{table}[t]
\centering\small
\caption{Per-model severe-violation rate (\%) on all 16
physical-consistency indicators (occ68 head-to-head set, $n{=}30$
per cell). Lower is better. Severe = annotator score $\le 0.33$
(Grade C or D). Models ordered by overall fidelity.}
\label{tab:h1_pc_full}
\resizebox{\linewidth}{!}{%
\setlength{\tabcolsep}{3pt}%
\begin{tabular}{lcccccccccccccccc}
\toprule
Model & A1 & A2 & A3 & A4 & M1 & M2 & O1 & O2 & IC0 & IC1 & IC2 & IC3 & ES1 & ES2 & EO1 & EO2 \\
& Color & Shape & Size & Mat. & Speed & Dir. & PartOcc & FullOcc & ActEff & Caus. & RespDir & Penetr. & Backg. & Light & Static & IndResp \\
\midrule
Happy Horse   &  0.0 & 10.0 & 10.0 &  6.7 &  3.3 &  6.7 & 16.7 &  0.0 &  6.7 &  6.7 &  6.7 &  6.7 &  0.0 & 0.0 &  3.3 &  0.0 \\
Wan2.1-14B    & 40.0 & 56.7 & 46.7 & 50.0 & 36.7 & 43.3 & 63.3 & 54.5 & 23.3 & 20.0 & 40.0 & 55.2 & 23.3 & 3.3 & 10.0 & 13.6 \\
DreamDojo-14B & 46.7 & 66.7 & 56.7 & 65.5 & 13.3 & 10.3 & 66.7 & 87.5 & 23.3 & 20.7 & 11.1 & 69.0 & 46.7 & 0.0 & 46.7 & 42.1 \\
DreamDojo-2B  & 60.0 & 86.7 & 70.0 & 76.7 & 10.0 &  3.4 & 83.3 & 75.0 & 13.3 & 13.8 &  3.4 & 70.0 & 73.3 & 0.0 & 26.7 & 30.0 \\
\bottomrule
\end{tabular}}
\end{table}

Three observations stand out from Tab.~\ref{tab:h1_pc_full}.

\textbf{(i) The headline ranking is preserved per indicator.}
Happy Horse posts the lowest severe-violation rate on
\textbf{14 of 16} indicators; the only exceptions are
M2~Direction and IC2~Resp-Dir, on both of which
DreamDojo-2B comes out marginally lower
($3.4\,\%$ vs.\ $6.7\,\%$ for Happy Horse). This rules out the
worry that the overall ranking is driven by a small subset of
dimensions.

\textbf{(ii) Different families fail on different things.}
The 14B baselines (DreamDojo-14B, Wan2.1-14B) fail predominantly
on static appearance (SC-A) and full occlusion (SC-O2: $87.5\,\%$
and $54.5\,\%$ severe respectively) but are competitive on motion
(SC-M); the smaller DreamDojo-2B is the worst overall but, like
Happy Horse, has near-perfect M2~Direction
($3.4\,\%$ severe). This decomposition is invisible in any
single-scalar metric.

\textbf{(iii) Penetration is the universal failure mode.}
IC3~No-Penetration shows severe-violation rates of
$6.7\,\%/55.2\,\%/69.0\,\%/70.0\,\%$ for
Happy~Horse / Wan2.1-14B / DreamDojo-14B / DreamDojo-2B
respectively. Even the strongest model in our suite produces
visibly inter-penetrating geometry on $1$ in $15$ rollouts;
the three remaining models fail on more than half.

\begin{figure}[t]
  \centering
  \IfFileExists{pics/h2_pc_abcd.pdf}{%
    \includegraphics[width=\linewidth]{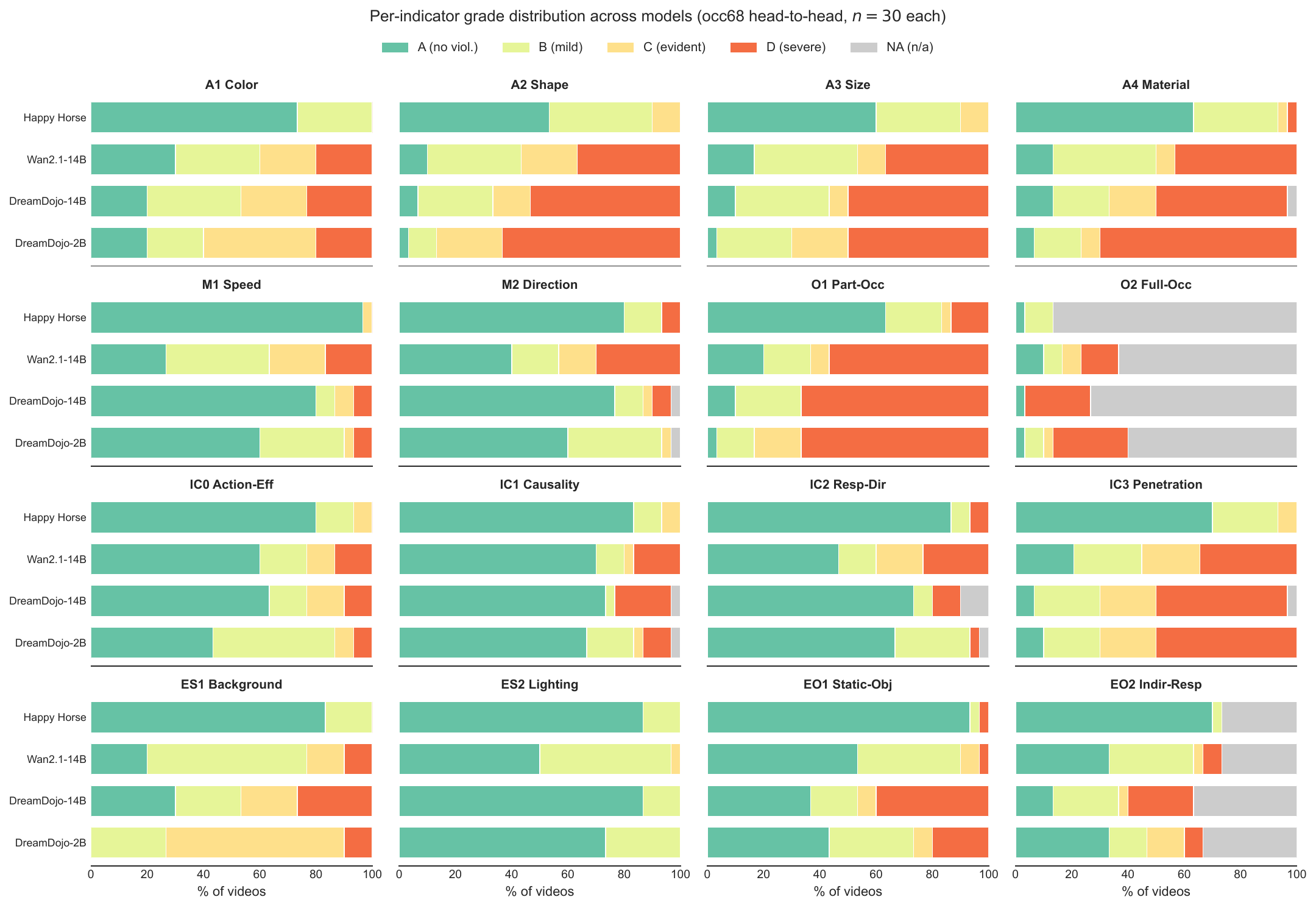}%
  }{%
    \fbox{\parbox{0.9\linewidth}{\centering\small[Figure placeholder: \texttt{pics/h2\_pc\_abcd.pdf}]}}%
  }
  \caption{\textbf{Full A/B/C/D grade distribution per
    indicator}, all four head-to-head models. Each panel
    summarises one indicator; bars within a panel correspond to
    the four models in the legend. The colour code follows the
    rubric of Appendix~G: green (A) = no
    perceived violation, light-green (B) = mild, yellow (C) =
    evident, red (D) = severe, grey = N/A. The proportion of red
    cells on the right (occlusion / penetration) is what drives
    the headline severe-violation gap reported in
    Tab.~\ref{tab:h1_pc_full}.}
  \label{fig:h2_pc_abcd}
\end{figure}

\subsection{Physics Law: anomaly mix on free-fall}
\label{sec:appendix_h_pl}

The Physics-Law level provides a focused stress test on a single
deterministic phenomenon (free fall) and is annotated for
DreamDojo-14B only. Released videos cover five free-falling
object categories (banana, bottle, corn, dragonfruit, cucumber;
$n{=}89$ rated clips in total), letting us verify that
failure-mode statistics are not an artefact of a single object
category. Each clip is reviewed against ten distinct violation
patterns.

\begin{figure}[t]
  \centering
  \IfFileExists{pics/h3_pl_grade.pdf}{%
    \includegraphics[width=0.85\linewidth]{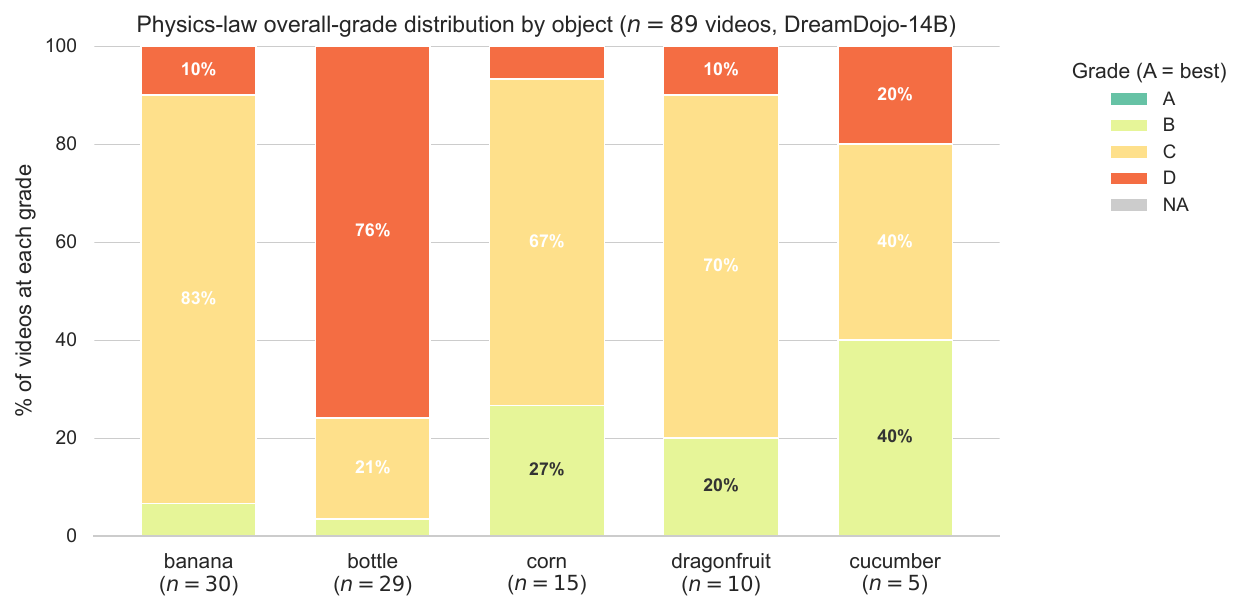}%
  }{%
    \fbox{\parbox{0.8\linewidth}{\centering\small[Figure placeholder: \texttt{pics/h3\_pl\_grade.pdf}]}}%
  }
  \caption{\textbf{Physics-law overall-grade distribution on the
    free-fall set} ($n{=}89$ DreamDojo-14B videos across five
    object categories). No video receives Grade A; the dominant
    grade is C/D, with Grade D alone accounting for $31.5\,\%$
    of clips. Even when the prompt distribution spans five object
    categories, the model's free-fall predictions rarely satisfy
    the rubric's plausibility bar.}
  \label{fig:h3_pl_grade}
\end{figure}

\begin{table}[t]
\centering\small
\caption{Frequency of each free-fall anomaly category in the
released Physics-Law set ($n{=}90$ DreamDojo-14B videos across
five object categories). The percentage is the share of videos
for which the annotator marked at least one non-`none' option in
that category.}
\label{tab:h2_pl_anomaly}
\begin{tabular}{lrrr}
\toprule
Anomaly category & $n_\mathrm{videos}$ & $n_\mathrm{flagged}$ & \% flagged \\
\midrule
Lack of acceleration / floating & 90 & 54 & 60.0 \\
Other anomaly (split / spin)    & 90 & 46 & 51.1 \\
Floating sensation (count)      & 90 & 45 & 50.0 \\
Local deceleration / hovering   & 89 & 42 & 47.2 \\
Trajectory jump                 & 90 & 42 & 46.7 \\
Horizontal drift                & 90 & 41 & 45.6 \\
Release anomaly                 & 90 & 28 & 31.1 \\
Landing anomaly                 & 90 & 25 & 27.8 \\
Object disappearance            & 90 & 17 & 18.9 \\
Bounce anomaly                  & 90 & 14 & 15.6 \\
\bottomrule
\end{tabular}
\end{table}

The headline of Tab.~\ref{tab:h2_pl_anomaly} is that
\textbf{every one of the ten anomaly families is triggered on a
non-trivial fraction of clips}, with lack-of-acceleration
appearing on a clear majority ($60.0\,\%$). The four trajectory
anomalies (jump, drift, hover, lack-of-acceleration) are not
mutually exclusive: a large share of videos receive at least one
such tag. Together with Fig.~\ref{fig:h3_pl_grade} this
confirms that the absent Grade-A rate is structural --- since the
set spans five distinct object categories, it cannot be
attributed to category-specific difficulty.

\subsection{Action Following Fidelity: completion vs.\ visual quality}
\label{sec:appendix_h_af}

The Action-Following level is labelled for DreamDojo-14B and
asks whether the generated rollout actually accomplishes the
language- or trajectory-specified manipulation, in addition to
rating five visual-quality dimensions. Splitting completion by
subset reveals the structural difficulty of the long-horizon
GR1 setting:

\begin{figure}[t]
  \centering
  \IfFileExists{pics/h4_af_tcr_vq.pdf}{%
    \includegraphics[width=\linewidth]{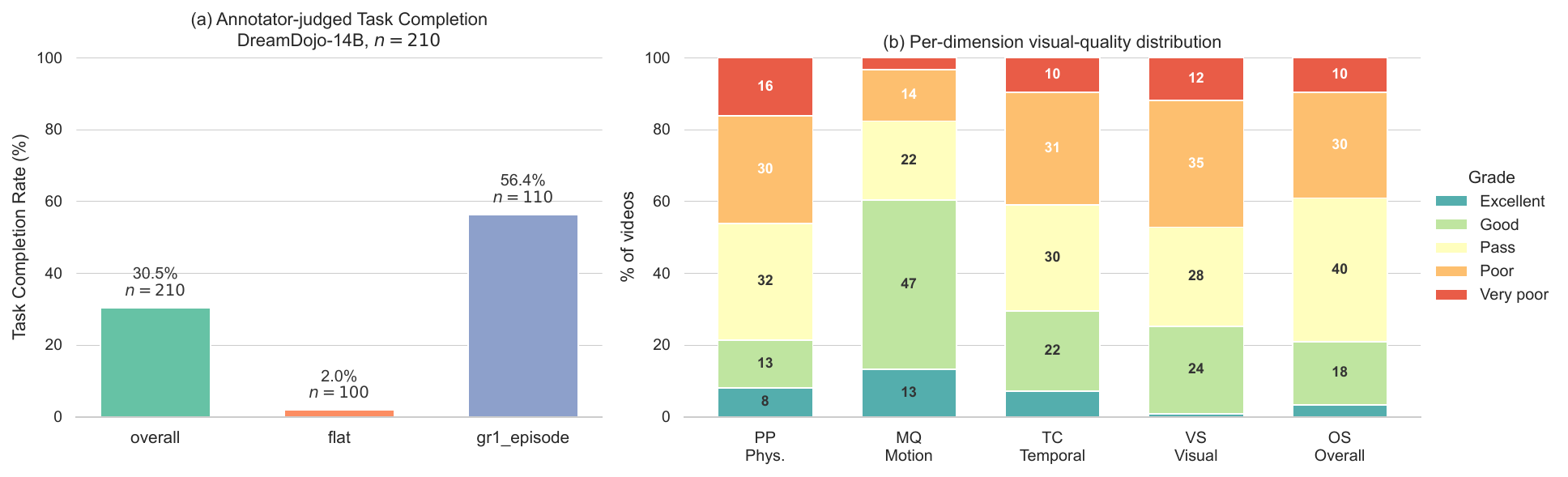}%
  }{%
    \fbox{\parbox{0.9\linewidth}{\centering\small[Figure placeholder: \texttt{pics/h4\_af\_tcr\_vq.pdf}]}}%
  }
  \caption{\textbf{Action Following on DreamDojo-14B
    ($n{=}210$).} (a) Annotator-judged Task Completion Rate
    overall and by subset; \textit{flat} is single-step
    pick-and-place ($n{=}100$), \textit{gr1\_episode} is the
    long-horizon humanoid setting ($n{=}110$).
    (b) Stacked grade distribution of the five visual-quality
    dimensions (PP physical plausibility, MQ motion quality,
    TC text-condition consistency, VS visual stability,
    OS overall sense). Motion quality is the strongest
    dimension; text-condition consistency and visual stability
    are the weakest.}
  \label{fig:h4_af_tcr_vq}
\end{figure}

The headline TCR is $30.5\,\%$, but this collapses two
qualitatively different regimes: $56.4\,\%$ on \textit{gr1\_episode}
and only $2.0\,\%$ on \textit{flat}. Inspection of the failure-reason
field on the 146 NO videos shows the dominant cause is
``the robot performed no effective operation (jittered in place)''
($66.4\,\%$ of failures), followed by ``task partially completed
but did not reach the goal state'' ($36.3\,\%$); only $2.1\,\%$
of failures are due to truncated / corrupted generation, ruling
out a video-IO confound.

The visual-quality panel of Fig.~\ref{fig:h4_af_tcr_vq} adds an
important caveat to any future work that uses video-quality
metrics as a proxy for task success. On the same 210 videos,
$60.4\,\%$ are rated Excellent or Good on motion quality, but
only $20.9\,\%$ reach that bar on overall sense and $25.3\,\%$
on visual stability --- and even ``Excellent motion quality''
samples can fail the task, since the actor wiggles smoothly
without ever moving the gripper towards the target. Visual
quality and task completion are therefore reported as
\emph{separate} columns rather than averaged in
Tab.~\ref{tab:h3_summary}.

\subsection{Optimism Bias: where world models hallucinate success}
\label{sec:appendix_h_ob}

The Optimism-Bias level is the only level that lets us compare
\emph{three} of the four models on identical prompts under both
the baseline and a counter-factual perturbation. We use the 88
videos with full per-model labels (DreamDojo-2B 30, Happy Horse
i2v 29, Wan2.1-14B i2v 29) for all per-model statistics; the
remaining 322 unlabelled videos are folded into the aggregate
counts only.

\begin{figure}[t]
  \centering
  \IfFileExists{pics/h5_ob_bias.pdf}{%
    \includegraphics[width=\linewidth]{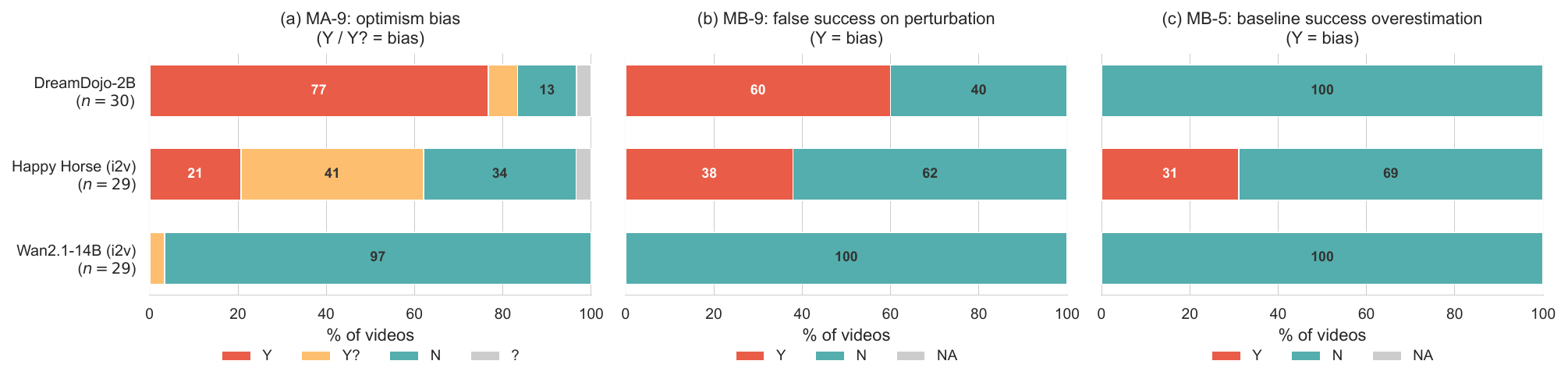}%
  }{%
    \fbox{\parbox{0.9\linewidth}{\centering\small[Figure placeholder: \texttt{pics/h5\_ob_bias.pdf}]}}%
  }
  \caption{\textbf{Three faces of optimism bias} on the labelled
    Optimism-Bias subset. (a) MA-9: does the rollout exhibit
    optimism bias on the unperturbed clip? Y and Y? both count
    as bias. (b) MB-9: does the rollout still predict success
    when the prompt is perturbed to make the task fail?
    (c) MB-5: does the rollout overestimate baseline success?
    Higher bars on the red end indicate worse behaviour.}
  \label{fig:h5_ob_bias}
\end{figure}

The picture is consistent across the three views.
DreamDojo-2B exhibits MA-9 bias on $83.3\,\%$ of clips,
hallucinates success on a perturbed prompt (MB-9) on $60.0\,\%$,
and posts a baseline-quality MA-1 mean of only $2.20/5$.
Happy Horse lowers both rates ($62.1\,\%$ MA-9, $37.9\,\%$
MB-9) but its baseline visual quality is the lowest of the
three ($1.76/5$), a pattern consistent with the model frequently
producing \emph{ambiguous} rollouts that the annotator labels as
``cannot determine optimism'' (Y? in panel a).
Wan2.1-14B is the most reliable: only $3.4\,\%$ MA-9 bias
and $0.0\,\%$ false-success on MB-9, paired with the highest
baseline quality ($3.52/5$).
These three numbers --- $83.3\,\%$, $62.1\,\%$, $3.4\,\%$ ---
match the headline Table~2 of the main paper exactly, which is
the empirical anchor we use to validate the schema mapping
released with the dataset.

\subsection{Cross-level summary}
\label{sec:appendix_h_summary}

\begin{table}[t]
\centering\small
\caption{Cross-level summary of the released human-annotation set.
Phys.~severe (\%): mean severe-violation rate over the 16
physical-consistency indicators on the occ68 head-to-head videos
($n{=}30$). Opt.~bias (\%): MA-9 share judged as bias (Y or Y?)
on the per-model task subset. False-success (\%): MB-9 share
judged Y (model predicts task success despite a failure-inducing
perturbation). Vis.~qual.: MA-1 mean baseline quality on a 1--5
scale (5 best). The Physics-Law and Action-Following levels
currently target a single model (DreamDojo-14B) and are reported
separately.}
\label{tab:h3_summary}
\begin{tabular}{lcccc}
\toprule
Model & Phys.~severe (\%)$\downarrow$ & Opt.~bias (\%)$\downarrow$ & False-success (\%)$\downarrow$ & Vis.~qual.$\uparrow$ \\
\midrule
Happy Horse   &  5.2 & 62.1 & 37.9 & 1.76 \\
Wan2.1-14B    & 36.2 &  3.4 &  0.0 & 3.52 \\
DreamDojo-14B & 42.1 & --   & --   & --   \\
DreamDojo-2B  & 43.5 & 83.3 & 60.0 & 2.20 \\
\bottomrule
\end{tabular}
\end{table}

Tab.~\ref{tab:h3_summary} brings the per-level analyses together
into a single view of the four models. The pattern that emerges
is that no single model dominates across all axes.
Happy Horse minimises the physical-consistency severity
($5.2\,\%$) but has the lowest baseline visual quality ($1.76$)
and a non-trivial $37.9\,\%$ false-success rate.
Wan2.1-14B is the best on the optimism-bias axis on every metric
we measure but is third on physical consistency.
The two DreamDojo variants show the expected scale effect on
optimism bias and visual quality, while remaining roughly tied on
physical consistency at the upper end of the severity range.
Crucially, the two evaluation levels that are currently
single-model (Physics Law and Action Following) are reported
separately rather than averaged, so as not to inflate any model's
apparent capability by attributing the strongest single-model
result to all four models.

\paragraph{Take-away.}
The release accompanying this paper enables third parties to
(i)~re-derive every number in
Sections~\ref{app:scoring:phys:pc}--\ref{app:scoring:obs} and
Appendix~\ref{app:appendix_h},
(ii)~re-aggregate the indicator-level judgements under different
severity rubrics,
and (iii)~score additional models against the same prompts
without re-running expensive human evaluation. We hope this
substantially lowers the barrier for evaluating future
world-model checkpoints on a common, transparent yardstick.

\section{Reproducibility}
\label{app:reproducibility}

\subsection{Action-Conditioned World Model Post-Training}
\label{app:repro:posttrain}

To ensure full reproducibility of the action-conditioned world models evaluated in MiraBench, we provide complete post-training details for the Cosmos-Predict2.5 models (2B and 14B), which serve as the backbone for DreamDojo-GR1 and Cosmos-GR1 evaluations.

\paragraph{Base model and architecture.}
Both models are initialized from the publicly released Cosmos-Predict2.5 pre-trained checkpoints~\citep{nvidia2025cosmosworldfoundationmodel} and share the same architecture class (\texttt{ActionChunkConditionedMinimalV1LVGDiT}), a DiT-based rectified flow model extended with action conditioning through AdaLN-LoRA injection.
The action conditioning mechanism projects the 29-dimensional active joint vector (both arms, both hands, waist) through dual-branch MLPs into per-block modulation parameters.

\begin{table}[h]
  \caption{Architecture and training configuration for Cosmos-Predict2.5 post-training.}
  \label{tab:posttrain_config}
  \centering\small
  \begin{tabular}{lcc}
    \toprule
    & 2B & 14B \\
    \midrule
    \textit{Architecture} \\
    \quad Model channels & 2048 & 5120 \\
    \quad DiT blocks & 28 & 36 \\
    \quad Attention heads & 16 & 40 \\
    \quad AdaLN-LoRA dim & 256 & 256 \\
    \midrule
    \textit{Training} \\
    \quad Learning rate & $3\!\times\!10^{-5}$ & $2\!\times\!10^{-5}$ \\
    \quad Batch size (per GPU $\times$ 8 GPUs) & $8 \times 8 = 64$ & $2 \times 8 = 16$ \\
    \quad Max iterations & 20,000 & 20,000 \\
    \quad Evaluated checkpoint & iter 6,000 & iter 6,000 \\
    \quad EMA rate (power) & 0.10 & 0.10 \\
    \quad Gradient clip norm & 0.1 & 0.1 \\
    \midrule
    \textit{Compute} \\
    \quad GPUs & 8$\times$ A100-80GB & 8$\times$ A100-80GB \\
    \quad Peak GPU memory & $\sim$80 GB & $\sim$57 GB \\
    \quad Training speed & $\sim$10.8 s/iter & $\sim$5.0 s/iter \\
    \quad Wall-clock to eval ckpt & $\sim$18 h & $\sim$8 h \\
    \bottomrule
  \end{tabular}
\end{table}

\paragraph{Training data.}
We use the GR1 robot subset of PhysicalAI-Robotics-GR00T-Teleop-GR1 (22,209 episodes, 6.4M frames, 5,257 tasks at 20 Hz).
Actions are the 29 active joint dimensions (excluding leg and neck), normalized via min-max scaling.
Video frames are sampled at every 2nd frame (effective 10 FPS), cropped and resized to 832$\times$480, yielding 13-frame clips per training sample.
Task-uniform sampling (max 3 episodes per task per epoch) ensures balanced coverage across manipulation categories.

\paragraph{Tokenizer and text encoder.}
Both models use the Wan2.1 VAE (8$\times$ spatial, 4$\times$ temporal compression, 16 latent channels) and Cosmos-Reason1-7B as the text encoder for cross-attention conditioning.

\paragraph{Inference.}
Generation proceeds in an autoregressive chunked fashion: each step takes the current frame plus a 12-step action chunk and produces a 13-frame video clip.
For longer sequences, chunks are concatenated with single-frame overlap removal.
The EMA weights in bf16 format (\texttt{model\_ema\_bf16.pt}) are used for all evaluations.

\paragraph{Code and scripts.}
All training and evaluation code is based on the open-source Cosmos-Predict2.5 repository.
Key scripts include:
\texttt{posttrain\_gr1\_robot\_14b.sh} (14B launch),
\texttt{posttrain\_gr1\_robot.sh} (2B launch),
\texttt{convert\_distcp\_to\_pt.py} (checkpoint conversion),
and \texttt{run\_all.sh} (multi-GPU benchmark runner).
The Hydra experiment configuration is provided at \texttt{configs/action\_conditioned/experiment/gr00t\_customized\_gr1.py}.

\subsection{Benchmark Evaluation Configuration}
\label{app:repro:benchconfig}

Table~\ref{tab:bench_config} lists key parameters and settings required to reproduce the full MiraBench evaluation pipeline.

\begin{table}[h]
  \caption{Benchmark evaluation configuration.}
  \label{tab:bench_config}
  \centering\small
  \begin{tabular}{lp{9cm}}
    \toprule
    Parameter & Value / Notes \\
    \midrule
    \textit{Data generation} \\
    \quad Action vector dimension & 384D (first 29 active; truncate for all models) \\
    \quad Perturbation severity & $s = 0.5$ for all types \\
    \quad Perturbation schedule & Fixed per task (2 mandatory + 1 task-specific; see Appendix~\ref{app:perturbations}) \\
    \quad Episodes per task & 2 (fixed stride sampling) \\
    \quad Conditions per episode & 4 (1 baseline + 3 perturbation types) \\
    \midrule
    \textit{Video generation} \\
    \quad Vector-conditioned resolution & 640$\times$480, 13 frames \\
    \quad Text-conditioned resolution & Model-native (832$\times$480 for Wan; API default for proprietary) \\
    \quad Guidance scale & 7.0 (Cosmos-based); 5.0 (Wan); API default (proprietary) \\
    \quad Random seed & Fixed per episode for deterministic generation \\
    \midrule
    \textit{Level 1: Physics Adherence} \\
    \quad VLM backbone & InternVL3-78B (zero-shot) \\
    \quad Level 1a (PCS) inference & 20 frames $\to$ 10 midcut pairs, single binary call per pair, $b_{\mathrm{B}}\!\ge\!1\!\to\!\texttt{B}$ \\
    \quad SAM2 model (physics law) & SAM2.1 (Hiera-Large) \\
    \quad Frame sampling (physics law) & All frames decoded; no subsampling \\
    \midrule
    \textit{Level 2: Action Following} \\
    \quad VLM backbone & InternVL3-78B (zero-shot) \\
    \quad Input format & Pred video + GT video + task instruction \\
    \quad Output dimensions & TCR (binary), OPS (0--1), GEN (0--1) \\
    \midrule
    \textit{Level 3: Optimism Bias} \\
    \quad VLM backbone & InternVL3-78B (zero-shot) \\
    \quad Frame extraction & 7 frames at [81, 83, 85, 87, 90, 95, 97]\% progress \\
    \quad Input format & Side-by-side baseline $|$ perturbed (cropped from composite or concatenated from separate files) \\
    \quad Voting threshold & $>$50\% ``Same'' $\rightarrow$ Y (bias present) \\
    \quad Prompt variant & Standard for action-conditioned; Lenient for text-conditioned \\
    \midrule
    \textit{Environment} \\
    \quad GPU & NVIDIA A100 80GB or H100 80GB \\
    \quad DreamDojo dependency & Cosmos-Predict2.5 repo + cosmos\_oss + Wan2.1 VAE \\
    \quad VLM dependency & InternVL3-78B with \texttt{lmdeploy} or native \texttt{transformers} \\
    \quad Python & 3.10+ \\
    \bottomrule
  \end{tabular}
\end{table}

\paragraph{Critical notes for reproduction.}
\begin{itemize}[leftmargin=*]
  \item For vector-conditioned models, action vectors must be truncated to the first 29 dimensions regardless of the source format (384D or 52D); zero-padded dimensions are ignored by all models.
  \item Text-conditioned models with non-GR1 rendering styles (e.g., Happy Horse, Wan) require the \textit{lenient} prompt variant for Level~3 evaluation to avoid false positives from stylistic differences.
  \item The 14B Cosmos checkpoint requires DCP-to-PyTorch conversion via \texttt{convert\_distcp\_to\_pt.py} with \texttt{--strip-ema-prefix}; this step requires $\sim$60 GB CPU RAM.
  \item The Cosmos guardrail model (Cosmos-Reason1-7B) must be accessible during inference; it can be cached from HuggingFace Hub.
\end{itemize}

\subsection{Random Seeds and Determinism}
\label{app:repro:seeds}

For Levels~2 and~3, each evaluation condition generates exactly one video with no cherry-picking from multiple samples.
For Level~1 (Physics Law Compliance), deterministic sampling with a fixed random seed per episode is used to ensure that the generated free-fall and sliding-contact scenarios are exactly reproducible; without seed fixing, stochastic generation may produce trajectories where the target object never exhibits any scorable translational motion， rendering the kinematic evaluation inapplicable.
VLM evaluator inference uses greedy decoding (temperature = 0) for all scoring queries and majority voting across frames for the bias detector.

\subsection{Data and Model Availability}
\label{app:repro:availability}

Upon acceptance, we will release:
\begin{itemize}[leftmargin=*]
  \item The complete MiraBench evaluation toolkit (scoring pipelines, prompts, perturbation code)
  \item All 906 human-annotated videos with 16,704 individual judgments
  \item Pre-generated evaluation videos for all reported models
  \item Post-training scripts and configuration files for Cosmos-Predict2.5
  \item Per-task perturbation schedules and natural-language prompt sets
\end{itemize}

\section{Physics Consistency Gallery}
\label{app:physics_gallery}

This appendix presents a per-indicator severity overview followed by
five representative cases illustrating distinct physical-consistency
failure modes. All examples are DreamDojo-14B predictions on the occ68
head-to-head set; we deliberately restrict the gallery to this model
because it is the only one whose paired \textit{(GT, prediction)}
clips are publicly released alongside the human annotations, so every
case can be reproduced from the open dataset.

\subsection{Per-Indicator Ablation}
\label{app:physics_gallery:ablation}

Table~\ref{tab:h1_pc_full} (Appendix~\ref{sec:appendix_h_pc}) reports the
per-indicator severe-violation rate for the four head-to-head models on
the occ68 set ($n{=}30$ per cell). Restricted to DreamDojo-14B and
re-sorted by severity, the 16 indicators fall into three regimes:

\begin{table}[h]
  \caption{
    Severe-violation rate (\%) of DreamDojo-14B on the occ68
    head-to-head set ($n{=}30$), grouped by severity regime.
    Severe = annotator score $\le 0.33$ (Grade C or D).
    Lower is better. Numbers are pulled directly from
    Tab.~\ref{tab:h1_pc_full}.
  }
  \label{tab:pcs_indicator_ablation}
  \centering\small
  \setlength{\tabcolsep}{6pt}
  \begin{tabular}{llr}
    \toprule
    Regime                          & Indicator                       & Severe (\%) \\
    \midrule
    \textit{Catastrophic} ($\geq$60\%)    & SC-O2 Full occlusion           & 87.5 \\
                                          & IC-3 No penetration            & 69.0 \\
                                          & SC-O1 Partial occlusion        & 66.7 \\
                                          & SC-A2 Shape                    & 66.7 \\
                                          & SC-A4 Material behaviour       & 65.5 \\
    \midrule
    \textit{Severe} (40--60\%)            & SC-A3 Size                     & 56.7 \\
                                          & SC-A1 Color                    & 46.7 \\
                                          & EC-S1 Background               & 46.7 \\
                                          & EC-O1 Static objects           & 46.7 \\
                                          & EC-O2 Indirect response        & 42.1 \\
    \midrule
    \textit{Stable} ($\leq$25\%)          & IC-0 Action effect             & 23.3 \\
                                          & IC-1 Causality                 & 20.7 \\
                                          & SC-M1 Speed                    & 13.3 \\
                                          & IC-2 Response direction        & 11.1 \\
                                          & SC-M2 Direction                & 10.3 \\
                                          & EC-S2 Lighting                 & 0.0  \\
    \bottomrule
  \end{tabular}
\end{table}

\paragraph{Key observations.}
\begin{enumerate}[leftmargin=*]
  \item \textbf{Occlusion is the dominant failure mode.}
    SC-O2 (87.5\%) and SC-O1 (66.7\%) are the two highest-severity
    indicators. Cases~\ref{app:physics_gallery:case1}
    and~\ref{app:physics_gallery:case2} show that a single occlusion
    event --- full or partial --- can break either the manipulated
    subject or a static scene element.
  \item \textbf{Appearance sub-dimensions fail together but not equally.}
    SC-A1/A2/A3/A4 cluster between $46.7\,\%$ and $66.7\,\%$ severe;
    motion (SC-M1/M2) is below $14\,\%$. Cases~\ref{app:physics_gallery:case3}
    (colour-only drift) and~\ref{app:physics_gallery:case4}
    (material drift surfacing on SC-A1/SC-A3 rather than SC-A4) show
    why scoring A1--A4 independently is necessary.
  \item \textbf{Penetration is the universal interaction failure.}
    IC-3 sits at $69.0\,\%$, second highest overall; the remaining
    interaction indicators (IC-0/IC-1/IC-2) are all below $24\,\%$.
    Case~\ref{app:physics_gallery:case5} reaches IC-3 via subject
    substitution that fuses with the gripper, demonstrating that
    penetration violations need not arise from explicit object--object
    contact.
  \item \textbf{Motion and lighting are not where world models fail.}
    SC-M1/M2 ($\le 14\,\%$) and EC-S2 ($0\,\%$) are essentially clean.
    The headline ``world models cannot do physics'' should be read as
    ``world models cannot maintain object identity through occlusion
    and contact'' --- not as a generic motion-quality problem.
\end{enumerate}
\subsection{Case 1: Cascading Failure under Full Occlusion}
\label{app:physics_gallery:case1}

\begin{figure}[h]
  \centering
  \includegraphics[width=\linewidth]{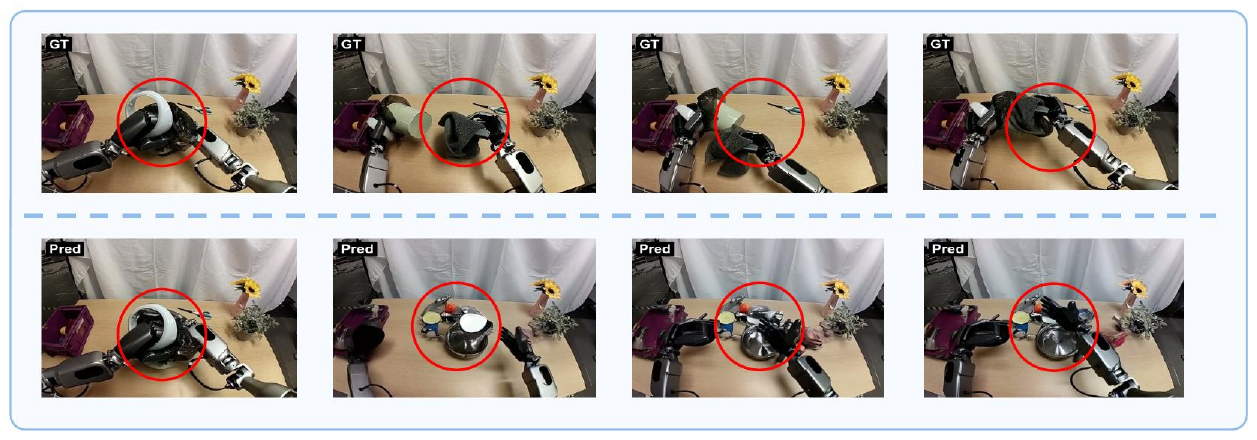}
  \caption{
    \textbf{DreamDojo-14B, occ68 episode 0008.}
    The dark cloth held by the right gripper is briefly fully occluded
    between the two arms (frame~1$\to$2). When it re-emerges in the
    prediction, the original cloth has been replaced by a stainless-steel
    kettle, a yellow plate and a red tomato-like object --- none of which
    appeared in the ground-truth scene.
  }
  \label{fig:pcs_case1}
\end{figure}

\noindent\textbf{Source:}
\texttt{MiraBench\_dataset/physical\_consistency/0421/occ68/0008\_pred.mp4}


\paragraph{Analysis.}
A single full-occlusion event is enough for the model to overwrite the
manipulated subject's identity, and the substitution drags 13 of the 16
indicators into severe violation. Occlusion robustness therefore acts
as a precondition for several other dimensions rather than an
independent axis: fix SC-O2 and the cascade collapses with it.

\subsection{Case 2: Partial Occlusion Damages a Static Object}
\label{app:physics_gallery:case2}

\begin{figure}[h]
  \centering
  \includegraphics[width=\linewidth]{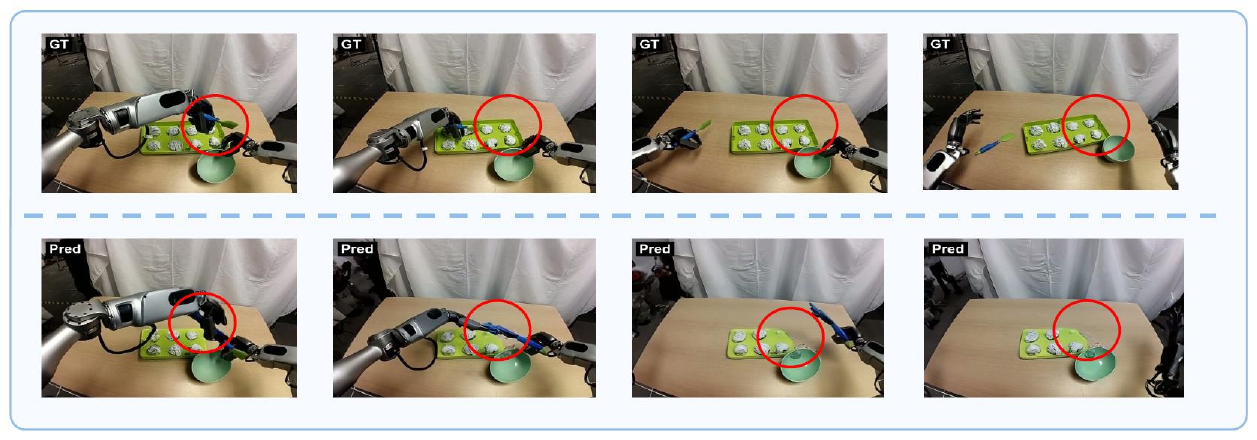}
  \caption{
    \textbf{DreamDojo-14B, occ68 episode 0015.}
    The right manipulator briefly traverses the upper-right corner of
    the green dumpling tray, partially occluding it (frame~1$\to$2).
    After the arm withdraws, the prediction renders the tray with a
    broken right edge: the section that was momentarily covered fails
    to reconstruct.
  }
  \label{fig:pcs_case2}
\end{figure}

\noindent\textbf{Source:}
\texttt{MiraBench\_dataset/physical\_consistency/0421/occ68/0015\_pred.mp4}


\paragraph{Analysis.}
The failure is not on the manipulator but on the static object the
manipulator briefly passed over: the model treats the previously-covered
patch as a free generative target instead of restoring the same content.
The violation is invisible to any metric that averages over the full
clip --- it only appears in the few frames immediately after un-occlusion.

\subsection{Case 3: Colour-Only Drift}
\label{app:physics_gallery:case3}

\begin{figure}[h]
  \centering
  \includegraphics[width=\linewidth]{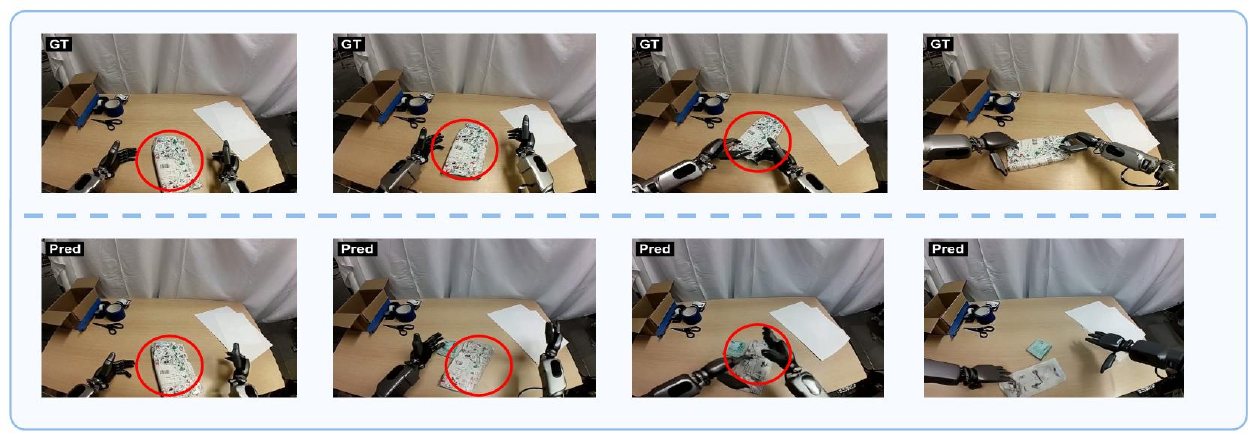}
  \caption{
    \textbf{DreamDojo-14B, occ68 episode 0063.}
    The multi-coloured patterned wrapper held by the gripper progressively
    desaturates in the prediction; its bright printed pattern flattens
    into a uniform pale-green tone over the four sampled frames.
  }
  \label{fig:pcs_case3}
\end{figure}

\noindent\textbf{Source:}
\texttt{MiraBench\_dataset/physical\_consistency/0421/occ68/0063\_pred.mp4}


\paragraph{Analysis.}
Geometry, motion and interaction indicators all stay at Grade A or B,
so the failure is concentrated on a single appearance dimension. A
colour-only failure of this type would be invisible to any metric that
averages over A1--A4, motivating the rubric's per-sub-dimension scoring.

\subsection{Case 4: Material Mismatch Surfaces on Other Indicators}
\label{app:physics_gallery:case4}

\begin{figure}[h]
  \centering
  \includegraphics[width=\linewidth]{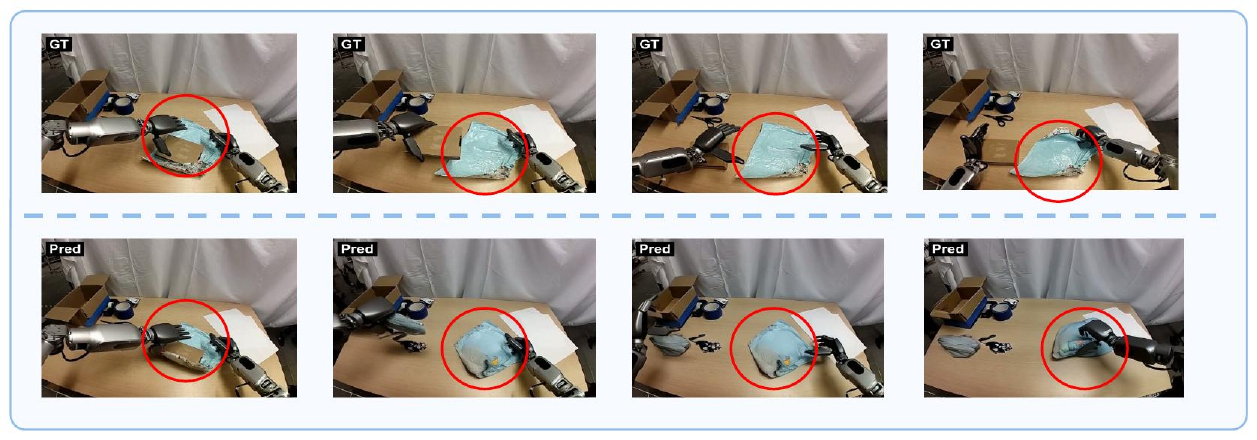}
  \caption{
    \textbf{DreamDojo-14B, occ68 episode 0064.}
    A blue plastic-coated wrapper that retains crisp creases and
    specular highlights in the ground truth drapes like a piece of soft
    cloth in the prediction. The plastic-like sheen is lost and the
    wrapper's area contracts noticeably between mid- and end-clip.
  }
  \label{fig:pcs_case4}
\end{figure}

\noindent\textbf{Source:}
\texttt{MiraBench\_dataset/physical\_consistency/0421/occ68/0064\_pred.mp4}


\paragraph{Analysis.}
The headline diagnosis is ``a plastic-coated wrapper behaving like
cloth'' --- yet the rubric's eponymous SC-A4 (Material) indicator only
fires at Grade B, while the appearance cues humans use to flag the
mismatch (loss of sheen, loss of creases, chromatic shift) light up on
SC-A1 and SC-A3 at Grade D. The case shows that the indicator suite
catches the failure even when the textbook-named axis does not.

\subsection{Case 5: Subject Substitution without Occlusion}
\label{app:physics_gallery:case5}

\begin{figure}[h]
  \centering
  \includegraphics[width=\linewidth]{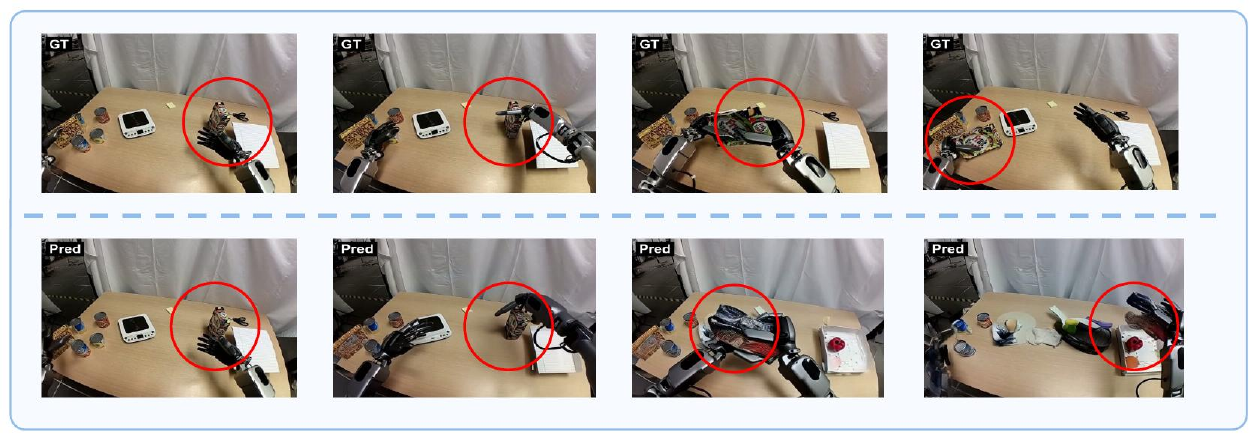}
  \caption{
    \textbf{DreamDojo-14B, occ68 episode 0045.}
    The patterned snack package being held in the ground truth disappears
    in the prediction; the workspace is then populated with a cluster
    of indistinct, ill-defined blobs whose category and geometry cannot
    be determined from the rendered pixels. In the last two frames these
    blobs adhere to the right gripper and forearm, fusing visually with
    the manipulator itself.
  }
  \label{fig:pcs_case5}
\end{figure}

\noindent\textbf{Source:}
\texttt{MiraBench\_dataset/physical\_consistency/0421/occ68/0045\_pred.mp4}


\paragraph{Analysis.}
Unlike Case~1, this substitution happens in the open: SC-O1/O2 stay
clean and the gripper's trajectory remains plausible (SC-M1/M2 = A,
IC-0/IC-1/IC-2 = A), so subject identity is overwritten without an
occlusion gate. The blob's eventual fusion with the gripper is what
pushes IC-3~no-penetration to Grade C, completing a primary
substitution mode that bypasses occlusion entirely.

\section{Action Following Gallery}
\label{app:gallery:action}

This appendix presents representative cases from the action following evaluation, illustrating distinct patterns across the TCR (task completion) and OPS (object preservation) dimensions.

\subsection{Case 1: Ideal Success (TCR\,=\,1, OPS\,=\,high)}
\label{app:gallery:case1}

\begin{figure}[h]
  \centering
  \includegraphics[width=\linewidth]{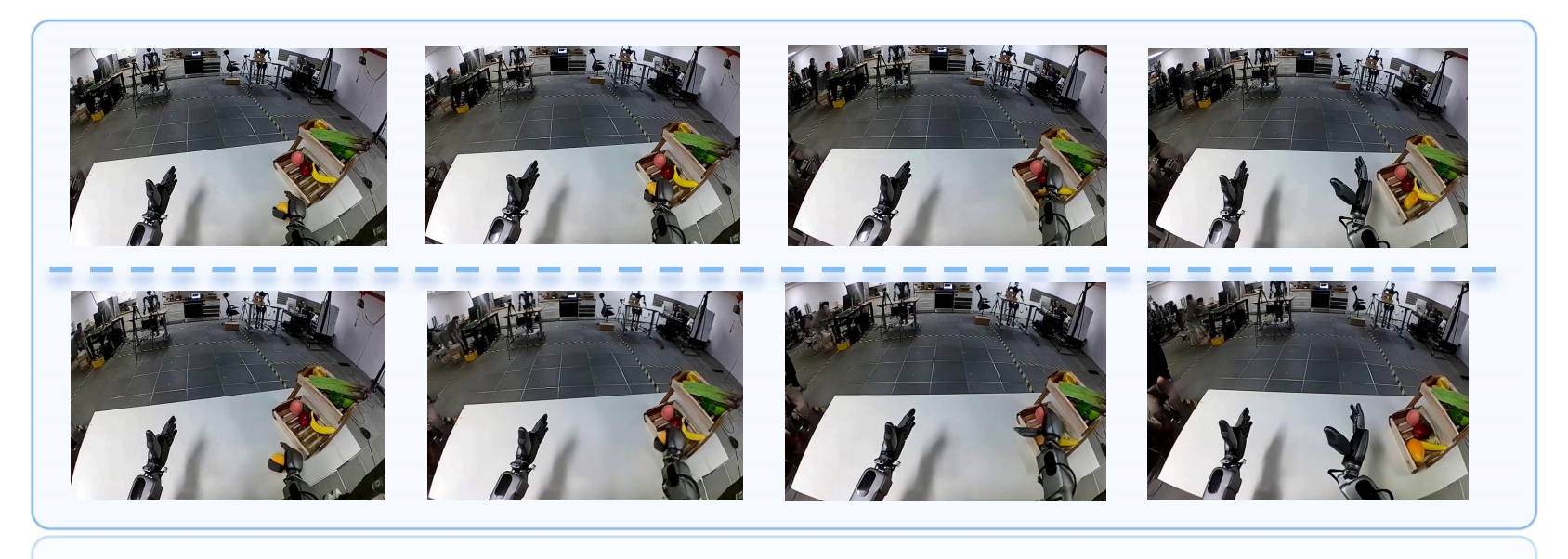}
  \caption{
    \textbf{DreamDojo-2B, episode\_0172.}
    The predicted video accurately follows the instruction: the robot grasps the mango and places it onto the lower shelf of the rack.
    Objects remain visually coherent throughout, with no deformation or artefacts.
  }
  \label{fig:af_case1}
\end{figure}

\noindent\textbf{Instruction:} \textit{``A GR1 humanoid robot stands in front of a yellow mango and a two-tier rack. The robot extends one arm to grasp the mango from the table. It then lifts the fruit and carefully places it onto the lower shelf of the rack.''}

\begin{center}\small
\begin{tabular}{lc|lc}
\toprule
\multicolumn{2}{c|}{\textit{Task Completion (TCR)}} & \multicolumn{2}{c}{\textit{Object Preservation (OPS)}} \\
\midrule
TCR & \textbf{1 (Completed)} & OPS & \textbf{high} \\
 &  & Confidence & 1.00 (16/16) \\
\bottomrule
\end{tabular}
\end{center}

\paragraph{Analysis.}
This episode represents the ideal outcome: the world model both completes the manipulation task and preserves object appearance throughout the sequence. The mango is correctly grasped, transported, and placed without any visual artefact, and the arm trajectory closely matches the expected motion. High TCR and high OPS together indicate that the model has learned the correct action--visual outcome mapping for this episode.

\subsection{Case 2: Object artefacts Despite Task Completion (TCR\,=\,1, OPS\,$<$\,high)}
\label{app:gallery:case2}

\begin{figure}[h]
  \centering
  \includegraphics[width=\linewidth]{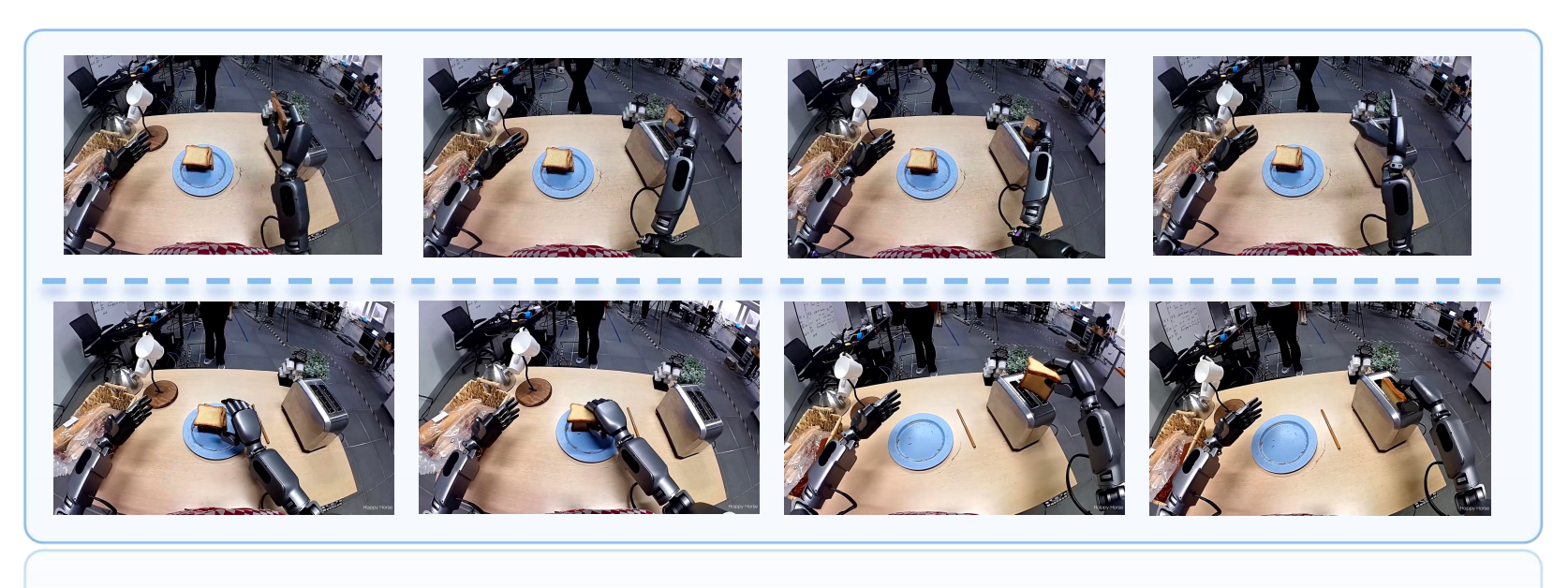}
  \caption{
    \textbf{Happy Horse, episode\_0127.}
    The robot appears to complete the task of placing toast into the toaster, but the bread slice undergoes severe visual distortion during manipulation---multiple slices merge into a single amorphous mass.
  }
  \label{fig:af_case2}
\end{figure}

\noindent\textbf{Instruction:} \textit{``A GR1 humanoid robot stands in front of a kitchen countertop. The robot uses one arm to carefully pick up a slice of toast from a blue plate. It then moves the toast and places it into a nearby toaster.''}

\begin{center}\small
\begin{tabular}{lc|lc}
\toprule
\multicolumn{2}{c|}{\textit{Task Completion (TCR)}} & \multicolumn{2}{c}{\textit{Object Preservation (OPS)}} \\
\midrule
TCR & \textbf{1 (Completed)} & OPS & \textbf{medium} \\
 &  & Confidence & 0.69 \\
 &  & Failing frames & 9--13 (merging/deformation) \\
\bottomrule
\end{tabular}
\end{center}

\paragraph{Analysis.}
This case highlights the independence of TCR and OPS: the robot achieves the task goal (toast reaches the toaster), yet the object undergoes severe metamorphosis during transport. Multiple bread slices fuse into a single undifferentiated mass, a failure mode invisible to a task-completion-only metric. The OPS dimension is essential for detecting such artefacts, as a downstream policy trained on this data would learn a physically impossible manipulation trajectory. The late-frame failures (frames 9--13) correspond to the grasp-and-lift phase, suggesting that the model's object representation collapses precisely when fine-grained hand--object interaction is required.

\subsection{Case 3: Dual Failure---Task Not Completed and Objects Distorted (TCR\,=\,0, OPS\,$<$\,high)}
\label{app:gallery:case3}

\begin{figure}[h]
  \centering
  \includegraphics[width=\linewidth]{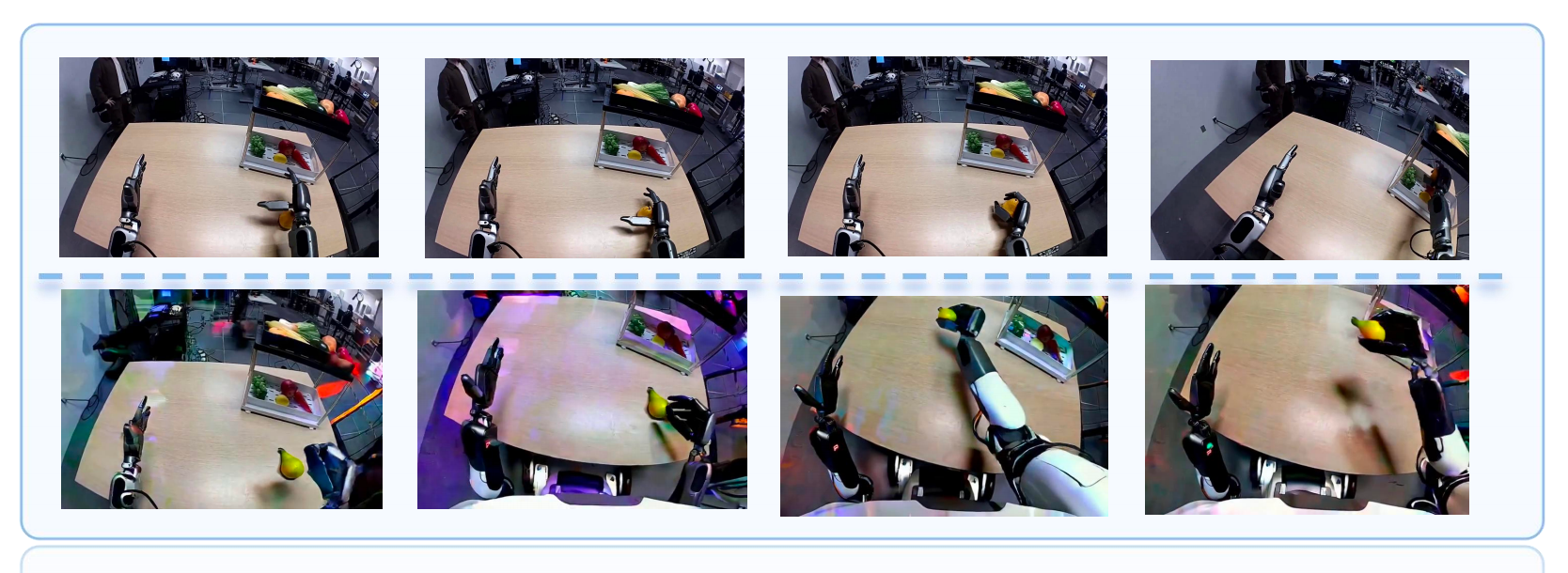}
  \caption{
    \textbf{Cosmos-14B, episode\_0021.}
    The robot fails to place the pear onto the tray, the arm reaches in the wrong direction, and the scene exhibits bizarre chromatic flashing with object deformation throughout.
  }
  \label{fig:af_case3}
\end{figure}

\noindent\textbf{Instruction:} \textit{``A GR1 humanoid robot stands in front of a tabletop with various objects. The robot uses one arm to grasp a yellow pear from the table. It then carefully places the pear onto the bottom tier of a multi-level white tray.''}

\begin{center}\small
\begin{tabular}{lc|lc}
\toprule
\multicolumn{2}{c|}{\textit{Task Completion (TCR)}} & \multicolumn{2}{c}{\textit{Object Preservation (OPS)}} \\
\midrule
TCR & \textbf{0 (Not completed)} & OPS & \textbf{medium} \\
 &  & Confidence & 0.62 \\
Action following & Wrong direction & Failing frames & 2, 11--16 (flash/deform) \\
\bottomrule
\end{tabular}
\end{center}

\paragraph{Analysis.}
This episode represents the most severe failure mode: the model neither completes the task nor preserves object integrity. The robot's arm reaches in a direction inconsistent with the target object, indicating that the action conditioning signal has been largely ignored. Simultaneously, chromatic artefacts and object deformation confirm that the model's visual generation is unstable. Both metrics correctly flag this episode, but their conjunction reveals a qualitatively different failure than either alone---this is not merely a bad action response or a visual glitch, but a fundamental breakdown in the model's ability to produce coherent video conditioned on the given action.

\subsection{Case 4: Optimism Bias---Task Labelled as Failed, Model Succeeds (TCR\,=\,0, OPS\,=\,high)}
\label{app:gallery:case4}

\begin{figure}[h]
  \centering
  \includegraphics[width=\linewidth]{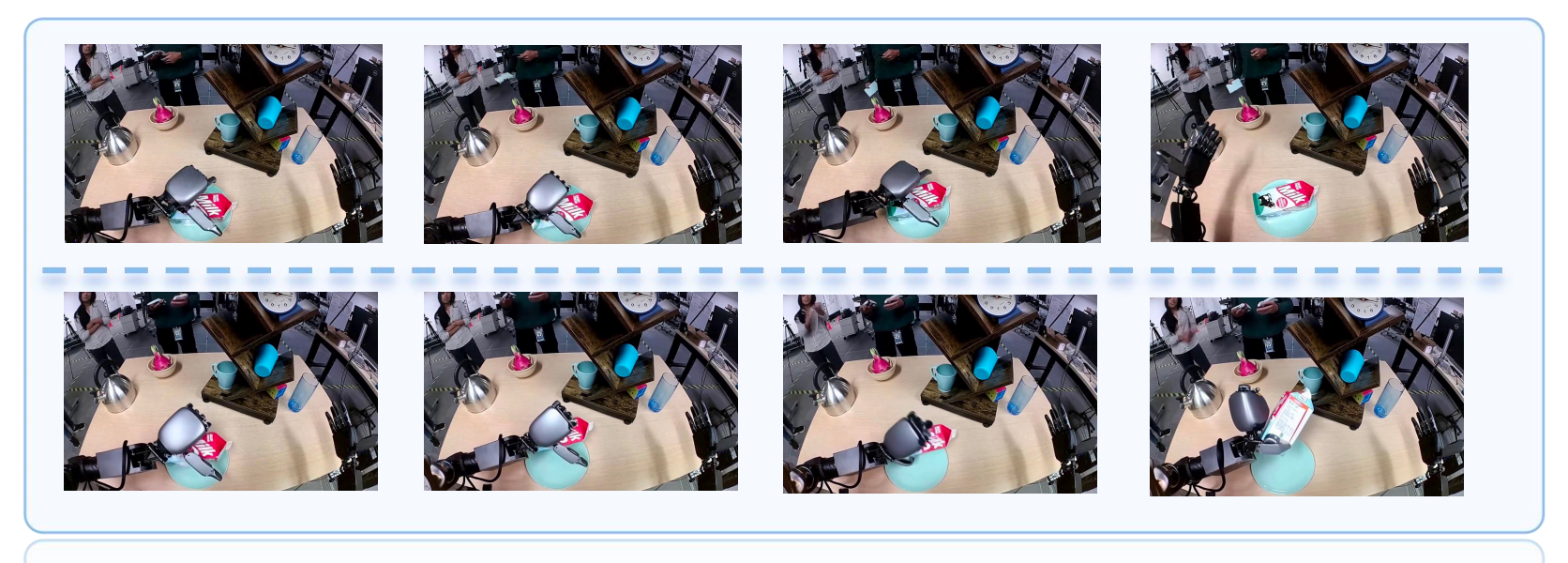}
  \caption{
    \textbf{Wan2.2, episode\_0082.}
    The ground-truth instruction specifies that the robot \emph{fails to lift} the milk carton, yet the predicted video shows the robot successfully grasping and lifting it.
    Objects remain perfectly preserved---the model generates a physically plausible but \emph{optimistically biased} outcome.
  }
  \label{fig:af_case4}
\end{figure}

\noindent\textbf{Instruction:} \textit{``In a tabletop scene with shelves and various objects, a milk carton sits on a plate in front of a GR1 humanoid robot. The robot extends a single arm to grasp the milk carton with its hand. It then attempts to lift the carton but failed.''}

\begin{center}\small
\begin{tabular}{lc|lc}
\toprule
\multicolumn{2}{c|}{\textit{Task Completion (TCR)}} & \multicolumn{2}{c}{\textit{Object Preservation (OPS)}} \\
\midrule
TCR & \textbf{0 (Not completed)} & OPS & \textbf{high} \\
 &  & Confidence & 1.00 (16/16) \\
GT outcome & \textbf{Failure (lift fails)} & & \\
Model outcome & \textbf{Success (lift succeeds)} & & \\
\bottomrule
\end{tabular}
\end{center}

\paragraph{Analysis.}
This episode reveals a subtle but critical failure mode that bridges action following and the optimism bias discussed in Appendix~\ref{app:gallery}.
The ground-truth demonstration encodes a \emph{failed} manipulation attempt, yet the world model generates a successful outcome: the milk carton is cleanly grasped and lifted with no visual artefact.
From a pure visual quality standpoint, the prediction is excellent (OPS\,=\,high, 16/16 frames pass).
The TCR judge correctly labels this as TCR\,=\,0---but only because the instruction explicitly states the task is to \emph{fail}, making the model's successful prediction a deviation from the prescribed action.

This case exemplifies how world models can override failure-signal conditioning with a learned success prior: having seen predominantly successful manipulations during training, the model generates the statistically common (successful) outcome rather than the prescribed (failed) one.
The object preservation is flawless precisely because the model's ``success template'' includes coherent object appearance---but the action following is fundamentally compromised.
This pattern connects directly to the optimism bias analysis in Appendix~\ref{app:gallery}, where models systematically ignore failure-inducing action perturbations in favour of generating successful outcomes.

\subsection{Summary of Action Following Patterns}

Table~\ref{tab:af_gallery_summary} consolidates the four patterns and their diagnostic implications.

\begin{table}[h]
  \caption{Summary of action following patterns observed in the gallery.}
  \label{tab:af_gallery_summary}
  \centering\small
  \setlength{\tabcolsep}{3pt}
  \begin{tabular}{clccp{5.5cm}}
    \toprule
    Case & Pattern & TCR & OPS & Diagnostic Implication \\
    \midrule
    1 & Ideal success & 1 & high & Correct action--visual mapping; model faithfully follows the instruction \\
    2 & Compromised success & 1 & $<$high & Task goal achieved but object integrity violated; physical impossibility masked by task metric alone \\
    3 & Dual failure & 0 & $<$high & Fundamental breakdown in both action conditioning and visual generation \\
    4 & Optimism bias & 0 & high & Model generates successful outcome when instructed to fail; bridges action following and optimism bias \\
    \bottomrule
  \end{tabular}
\end{table}

\section{Optimism Bias Gallery}
\label{app:gallery}

This appendix presents a per-perturbation ablation analysis followed by representative cases illustrating distinct optimism bias patterns.

\subsection{Per-Perturbation Ablation}
\label{app:gallery:ablation}

\begin{figure}[h]
  \centering
  \includegraphics[width=0.85\linewidth]{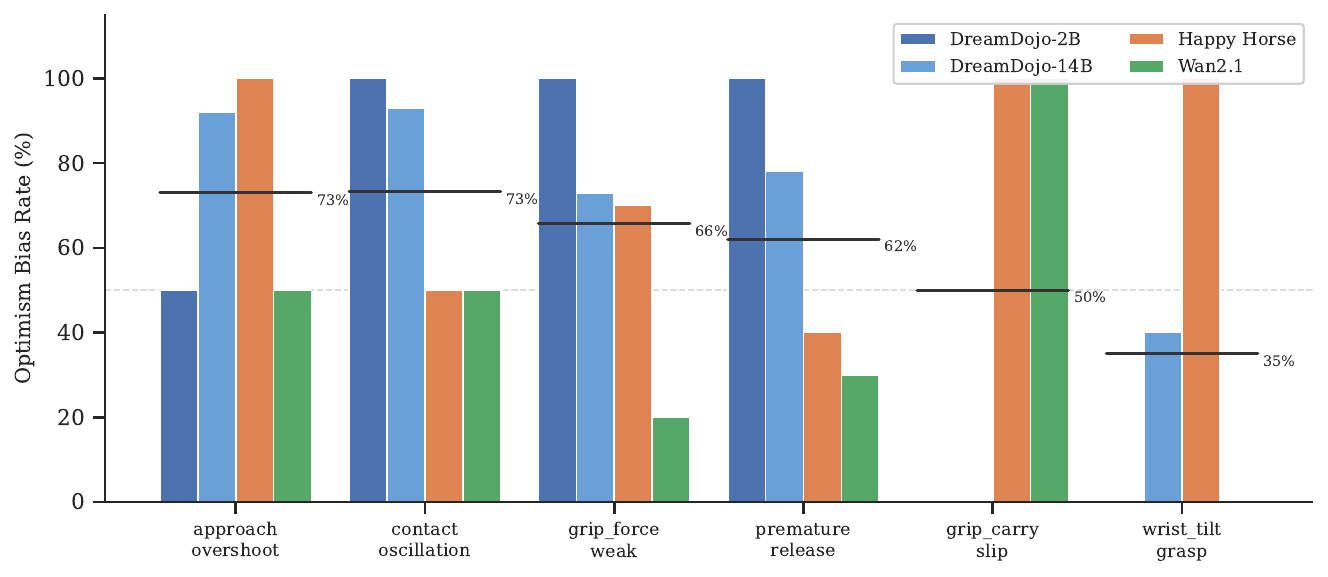}
  \caption{
    Per-perturbation optimism bias rate broken down by model.
    \texttt{grip\_force\_weak} and \texttt{contact\_oscillation} trigger the highest bias in DreamDojo-2B (100\%),
    while \texttt{wrist\_tilt\_grasp} is the least biased perturbation overall (33.3\%).
    Note that model sensitivity varies strongly by perturbation type: DreamDojo-2B shows 0\% bias on
    \texttt{grip\_carry\_slip} and \texttt{wrist\_tilt\_grasp}, while Happy Horse shows 100\% on those same types.
  }
  \label{fig:pert_ablation_bar}
\end{figure}

Table~\ref{tab:pert_ablation} reports the optimism bias rate (MA-9 = Y or Y?) for each perturbation type, averaged across models and broken down per model.
Figure~\ref{fig:pert_ablation_bar} visualizes the same data.

\begin{table}[h]
  \caption{
    Optimism bias rate (\%) by perturbation type.
    Higher values indicate the model more frequently ignores the perturbation.
    Rates are computed as the fraction of annotated samples where MA-9 $\in$ \{Y, Y?\}.
  }
  \label{tab:pert_ablation}
  \centering\small
  \setlength{\tabcolsep}{5pt}
  \begin{tabular}{lccccc}
    \toprule
    Perturbation & All & DreamDojo-2B & DreamDojo-14B & Happy Horse & Wan2.1 \\
    \midrule
    \texttt{approach\_overshoot}    & 73\% & 50\%  & 92\% & 100\% & 50\%  \\
    \texttt{contact\_oscillation}   & 73\% & 100\% & 93\% & 50\%  & 50\%  \\
    \texttt{grip\_force\_weak}      & 66\% & 100\% & 73\% & 70\%  & 20\%  \\
    \texttt{premature\_release}     & 62\% & 100\% & 78\% & 40\%  & 30\%  \\
    \texttt{grip\_carry\_slip}      & 50\% & 0\%   & --   & 100\% & 100\% \\
    \texttt{wrist\_tilt\_grasp}     & 35\% & 0\%   & 40\% & 100\% & 0\%   \\
    \bottomrule
  \end{tabular}
\end{table}

\paragraph{Key observations.}
\begin{enumerate}[leftmargin=*] 
  \item \textbf{No perturbation universally succeeds}: every perturbation type is ignored by at least one model at $\geq$50\% rate, confirming that optimism bias is not limited to specific failure modes.
  \item \textbf{Model-perturbation interaction}: DreamDojo-2B shows 100\% bias on force/timing perturbations (\texttt{grip\_force\_weak}, \texttt{premature\_release}, \texttt{contact\_oscillation}) but 0\% on spatial ones (\texttt{grip\_carry\_slip}, \texttt{wrist\_tilt\_grasp}). Happy Horse shows the reverse pattern. This suggests different models learn different aspects of the action space.
  \item \textbf{Force-based perturbations are hardest to detect}: \texttt{grip\_force\_weak} (63.3\% avg bias) modifies only the magnitude of hand joints without changing trajectory geometry, making it the most ``invisible'' perturbation from a visual standpoint.
\end{enumerate}

\subsection{Case 1: High-Quality Complete Bias}
\label{app:gallery:case1}

\begin{figure}[h]
  \centering
  \includegraphics[width=\linewidth]{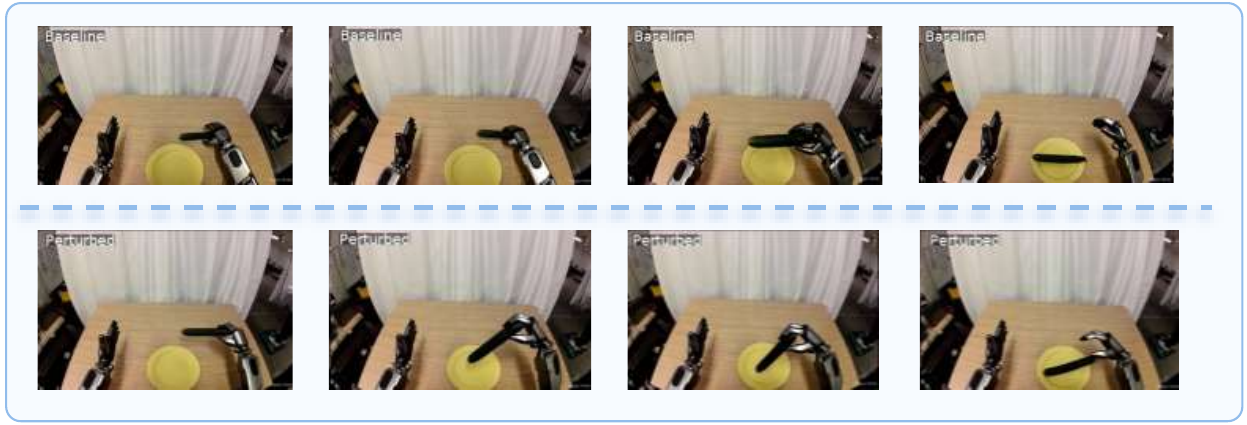}
  \caption{
    \textbf{Happy Horse, task\_023.}
    The baseline prediction is visually excellent (MA-1 = 4/5) with fully plausible physics,
    yet the perturbed video is indistinguishable from baseline despite a failure-inducing perturbation.
    The model generates physically coherent, task-successful outcomes regardless of the action input.
  }
  \label{fig:gallery_case1}
\end{figure}

\noindent\textbf{Source:} \texttt{human\_annotation\_happyhorse\_i2v/videos/task\_023/}

\begin{center}\small
\begin{tabular}{ll|ll}
\toprule
\multicolumn{2}{c|}{\textit{Module A (Perturbation Sensitivity)}} & \multicolumn{2}{c}{\textit{Module B (Task Success)}} \\
\midrule
Baseline Quality (MA-1) & 4/5 (High similarity) & Baseline Completion (MB-1) & Mostly complete \\
Robot Action (MA-2) & Mostly rational & Key Action Correct (MB-2) & Mostly correct \\
Temporal Fluency (MA-3) & Mostly smooth & Phys.\ Plausibility (MB-3) & Fully plausible \\
Perturbation Impact (MA-4) & \textbf{No effect} & GT Deviation (MB-4) & High consistency \\
Perturbed Realism (MA-5) & Mostly realistic & Perturbed Completion (MB-6) & \textbf{Mostly complete} \\
Quality Degradation (MA-6) & Minor & Final State Deviation (MB-7) & Identical \\
\textbf{Optimism Bias (MA-9)} & \textbf{Y (Full)} & \textbf{False Success (MB-9)} & \textbf{Y (False success)} \\
\bottomrule
\end{tabular}
\end{center}

\paragraph{Analysis.}
This is the most dangerous failure mode: the model produces physically coherent, visually convincing videos that look correct by any quality metric, while completely ignoring the failure-inducing action.
A downstream policy trained on such data would never encounter the failure mode encoded in the perturbation.

\subsection{Case 2: Scale Does Not Fix Bias}
\label{app:gallery:case2}

\begin{figure}[h]
  \centering
  \includegraphics[width=\linewidth]{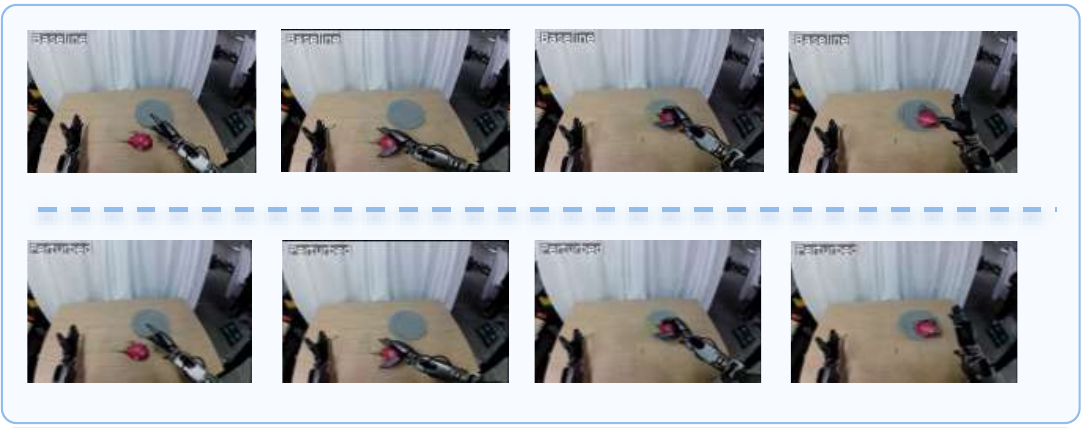}
  \caption{
    \textbf{DreamDojo-2B, task\_024.}
    Despite lower visual quality than larger models, this 2B model achieves high baseline-GT alignment (MA-1 = 4/5) on this episode while completely ignoring the perturbation.
    The perturbed prediction shows task completion identical to baseline.
  }
  \label{fig:gallery_case2}
\end{figure}

\begin{center}\small
\begin{tabular}{ll|ll}
\toprule
\multicolumn{2}{c|}{\textit{Module A}} & \multicolumn{2}{c}{\textit{Module B}} \\
\midrule
Baseline Quality (MA-1) & 4/5 (High similarity) & Baseline Completion (MB-1) & Fully complete \\
Perturbation Impact (MA-4) & \textbf{No effect} & Phys.\ Plausibility (MB-3) & Mostly plausible \\
Quality Degradation (MA-6) & None & Perturbed Completion (MB-6) & \textbf{Mostly complete} \\
\textbf{Optimism Bias (MA-9)} & \textbf{Y (Full)} & False Success (MB-9) & \textbf{Y} \\
\bottomrule
\end{tabular}
\end{center}

\paragraph{Analysis.}
Even a 2B model can produce episode-specific perfect alignment with ground truth while exhibiting full optimism bias.
This demonstrates that the success prior operates independently of general generation capability: the model has ``memorized'' this task's success pattern so strongly that no action perturbation can override it.

\subsection{Case 3: Low Quality + Full Bias}
\label{app:gallery:case3}

\begin{figure}[h]
  \centering
  \includegraphics[width=\linewidth]{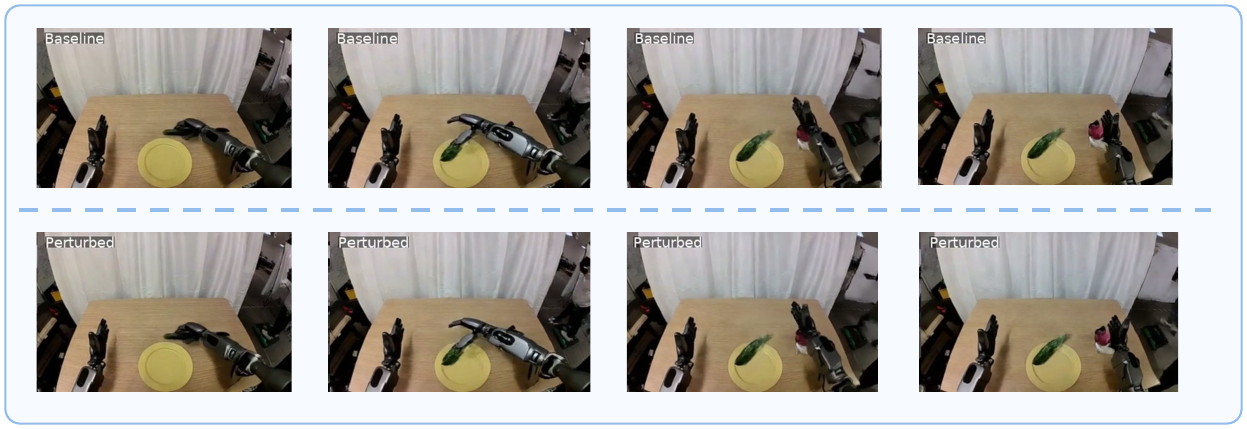}
  \caption{
    \textbf{DreamDojo-2B, task\_029.}
    The baseline itself exhibits poor physical plausibility (penetration, impossible deformation),
    yet the model still generates task-successful predictions under perturbation.
    Bias persists even when the model's physics understanding is clearly deficient.
  }
  \label{fig:gallery_case3}
\end{figure}

\noindent\textbf{Source:} \texttt{human\_annotation\_dreamdojo\_2b\_gr1/videos/task\_029/}

\begin{center}\small
\begin{tabular}{ll|ll}
\toprule
\multicolumn{2}{c|}{\textit{Module A}} & \multicolumn{2}{c}{\textit{Module B}} \\
\midrule
Baseline Quality (MA-1) & 2/5 (Low) & Baseline Completion (MB-1) & Mostly complete \\
Perturbation Impact (MA-4) & \textbf{No effect} & Phys.\ Plausibility (MB-3) & \textbf{Completely implausible} \\
Quality Degradation (MA-6) & Minor & Perturbed Completion (MB-6) & \textbf{Mostly complete} \\
Motion Coherence (MA-8) & Consistent & Final State (MB-7) & Identical \\
\textbf{Optimism Bias (MA-9)} & \textbf{Y (Full)} & Unexpected Success (MB-8) & No \\
\bottomrule
\end{tabular}
\end{center}

\paragraph{Analysis.}
This case separates optimism bias from physical competence.
The model cannot generate physically plausible interactions (MB-3 = ``completely implausible''), yet its success prior is strong enough to produce task-completion outcomes regardless.
The bias is not a consequence of sophisticated physics understanding; it is a shallow statistical pattern learned from success-only training data.

\subsection{Case 4: Correct Failure Prediction (No Bias)}
\label{app:gallery:case4}

\begin{figure}[h]
  \centering
  \includegraphics[width=\linewidth]{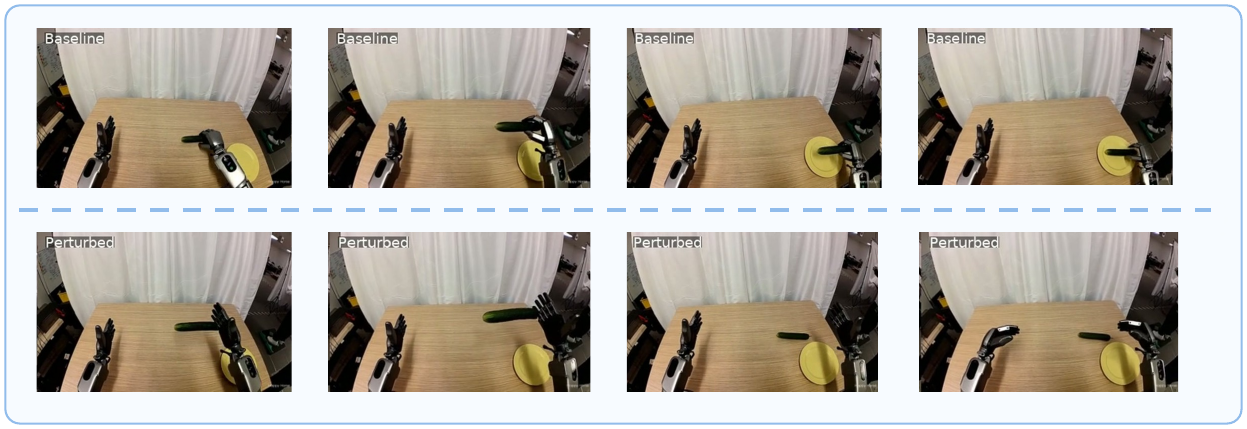}
  \caption{
    \textbf{Happy Horse, task\_022.}
    The model correctly responds to the perturbation: the perturbed prediction shows clear task failure
    (MB-6 = ``not completed'') with significant trajectory divergence from baseline.
    This demonstrates that the same model architecture \textit{can} propagate failure signals in some cases.
  }
  \label{fig:gallery_case4}
\end{figure}

\noindent\textbf{Source:} \texttt{human\_annotation\_happyhorse\_i2v/videos/task\_022/}

\begin{center}\small
\begin{tabular}{ll|ll}
\toprule
\multicolumn{2}{c|}{\textit{Module A}} & \multicolumn{2}{c}{\textit{Module B}} \\
\midrule
Baseline Quality (MA-1) & 4/5 (High) & Baseline Completion (MB-1) & Mostly complete \\
Perturbation Impact (MA-4) & \textbf{Significant} & Phys.\ Plausibility (MB-3) & Mostly plausible \\
Quality Degradation (MA-6) & Minor & Perturbed Completion (MB-6) & \textbf{Not completed} \\
Motion Coherence (MA-8) & Clearly worse & Final State (MB-7) & \textbf{Clear deviation} \\
\textbf{Optimism Bias (MA-9)} & \textbf{N (No bias)} & False Success (MB-9) & \textbf{N (Correct failure)} \\
\bottomrule
\end{tabular}
\end{center}

\paragraph{Analysis.}
This positive example shows that the same model (Happy Horse) that exhibits full bias in Case~1 can correctly propagate perturbation effects in other episodes.
The bias is therefore not a global architectural limitation but rather episode- and perturbation-type dependent, suggesting that certain task structures are more susceptible to prior override than others.

\subsection{Case 5: Mild Bias (Insufficient Response)}
\label{app:gallery:case5}

\begin{figure}[h]
  \centering
  \includegraphics[width=\linewidth]{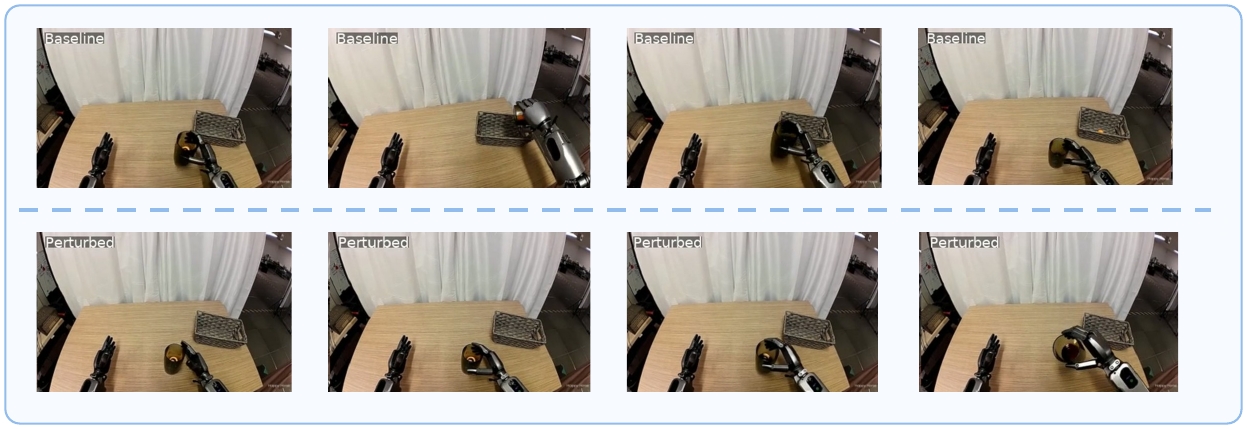}
  \caption{
    \textbf{Happy Horse, task\_011.}
    The model shows a detectable but insufficient response to the perturbation:
    the perturbed video achieves only partial completion with large final-state deviation from baseline,
    yet annotators still detect mild optimism bias (MA-9 = Y?) and false success (MB-9 = Y),
    indicating the model dampens but does not fully propagate the failure signal.
  }
  \label{fig:gallery_case5}
\end{figure}

\noindent\textbf{Source:} \texttt{human\_annotation\_happyhorse\_i2v/videos/task\_011/}

\begin{center}\small
\begin{tabular}{ll|ll}
\toprule
\multicolumn{2}{c|}{\textit{Module A}} & \multicolumn{2}{c}{\textit{Module B}} \\
\midrule
Baseline Quality (MA-1) & 3/5 (Moderate) & Baseline Completion (MB-1) & Fully complete \\
Perturbation Impact (MA-4) & \textbf{Minor effect} & Phys.\ Plausibility (MB-3) & Mostly plausible \\
Quality Degradation (MA-6) & Minor & Perturbed Completion (MB-6) & \textbf{Partially complete} \\
Motion Coherence (MA-8) & Slightly worse & Final State (MB-7) & \textbf{Large deviation} \\
\textbf{Optimism Bias (MA-9)} & \textbf{Y? (Mild bias)} & False Success (MB-9) & \textbf{Y} \\
\bottomrule
\end{tabular}
\end{center}

\paragraph{Analysis.}
Mild bias represents a continuum between full suppression and correct propagation.
The model registers the perturbation at a surface level (some trajectory change, partial completion rather than full success) but does not propagate it to the expected physical consequence (complete task failure).
This pattern suggests a competition between the action conditioning signal and the learned success prior, with the prior partially winning.

\subsection{Case 6: Generation Inadequacy Masking as Low Bias}
\label{app:gallery:case6}

\begin{figure}[h]
  \centering
  \includegraphics[width=\linewidth]{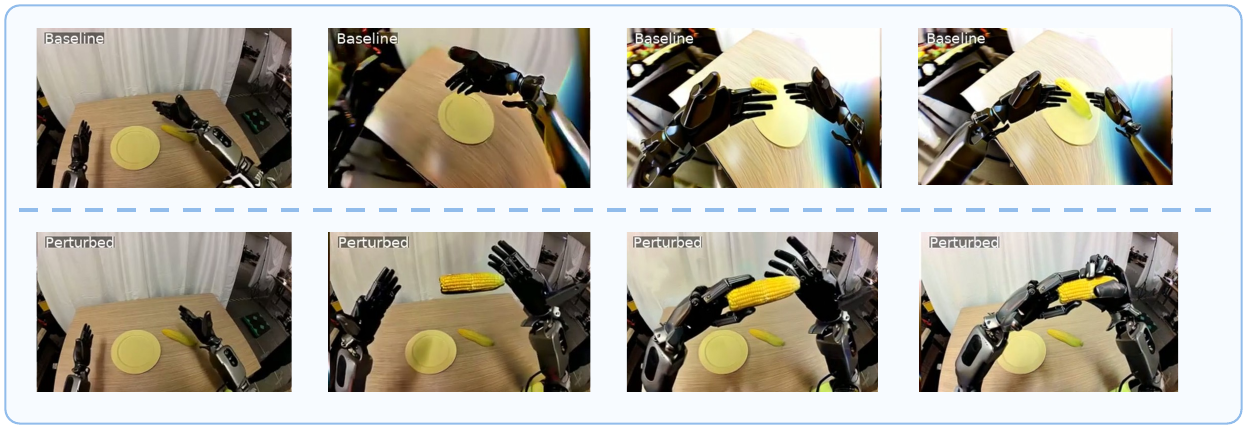}
  \caption{
    \textbf{Wan2.1, task\_017.}
    The model shows no full optimism bias, but not because it correctly follows actions:
    the baseline itself completely fails (MA-1 = 1/5, MB-1 = ``not completed''),
    so there is no ``success'' for the model to hallucinate under perturbation.
    Low bias here reflects incapability, not fidelity.
  }
  \label{fig:gallery_case6}
\end{figure}

\noindent\textbf{Source:} \texttt{human\_annotation\_wan21\_i2v\_14b/videos/task\_017/}

\begin{center}\small
\begin{tabular}{ll|ll}
\toprule
\multicolumn{2}{c|}{\textit{Module A}} & \multicolumn{2}{c}{\textit{Module B}} \\
\midrule
Baseline Quality (MA-1) & \textbf{1/5 (Different)} & Baseline Completion (MB-1) & \textbf{Not completed} \\
Perturbation Impact (MA-4) & No effect & Phys.\ Plausibility (MB-3) & Completely implausible \\
Quality Degradation (MA-6) & None & Perturbed Completion (MB-6) & Cannot judge \\
\textbf{Optimism Bias (MA-9)} & \textbf{Y? (Mild)} & False Success (MB-9) & N/A (baseline fails) \\
\bottomrule
\end{tabular}
\end{center}

\paragraph{Analysis.}
This case illustrates why MiraBench's hierarchical evaluation is necessary.
Wan2.1 scores low on optimism bias, but this is \textit{not} because it faithfully follows action conditioning---it fails at Level~1 (physical consistency) and Level~2 (action following) so completely that the Level~3 signal becomes uninterpretable.
Without the hierarchical structure, this model's low bias score would be misleadingly favorable.
MiraBench's nested preconditions prevent this misinterpretation: a model that fails Level~1 cannot claim credit for passing Level~3.

\subsection{Summary of Bias Patterns}

Table~\ref{tab:gallery_summary} consolidates the six failure patterns and their diagnostic implications.

\begin{table}[h]
  \caption{Summary of optimism bias patterns observed in the failure gallery.}
  \label{tab:gallery_summary}
  \centering\small
  \setlength{\tabcolsep}{3pt}
  \begin{tabular}{clccp{5.5cm}}
    \toprule
    Case & Pattern & Bias & Quality & Diagnostic Implication \\
    \midrule
    1 & High-quality full bias & Y & High & Most dangerous: indistinguishable from correct behavior by any quality metric \\
    2 & Scale-independent bias & Y & High & Model scale improves quality but not action sensitivity \\
    3 & Low-quality full bias & Y & Low & Bias is independent of physics competence; shallow statistical pattern \\
    4 & Correct failure (no bias) & N & High & Same model can propagate failure in some cases; bias is task-dependent \\
    5 & Mild bias (dampened response) & Y? & Medium & Competition between action signal and success prior; partial suppression \\
    6 & Incapability masking as low bias & Y? & Very low & Low bias from generation failure, not action fidelity; hierarchical eval needed \\
    \bottomrule
  \end{tabular}
\end{table}

\section{Computation Cost}
\label{app:computation}

Table~\ref{tab:compute_generation} summarizes the video generation cost across evaluated models.
All measurements are on NVIDIA A100 80GB GPUs.

\begin{table}[h]
  \caption{
    Video generation cost per model on the full MiraBench test set.
  }
  \label{tab:compute_generation}
  \centering\small
  \begin{tabular}{lcccc}
    \toprule
    Model & Params & GPU Mem & Time/video\\
    \midrule
    DreamDojo-14B  & 14B & $\sim$75 GB & $\sim$3.5 min\\
    DreamDojo-2B   & 2B  & $\sim$18 GB & $\sim$1.2 min\\
    Cosmos-14B     & 14B & $\sim$72 GB & $\sim$3.0 min\\
    Cosmos-2B      & 2B  & $\sim$16 GB & $\sim$1.0 min\\
    Wan2.1-14B     & 14B & $\sim$65 GB & $\sim$4.0 min\\
    Wan2.1-1B      & 1B  & $\sim$12 GB & $\sim$0.8 min\\
    \bottomrule
  \end{tabular}
\end{table}

\paragraph{Evaluation pipeline.}
After video generation, running the full MiraBench evaluation (all three levels) on a single model takes approximately 1.5 hours on one A100 GPU, dominated by VLM inference across the three evaluators.

\paragraph{One-time costs.}
VLM evaluator training requires $\sim$5 GPU-hours total across all levels.
Gemini distillation for training data construction is a one-time API cost; the distilled evaluator runs locally thereafter.
The human annotation study required $\sim$200 person-hours.

\paragraph{End-to-end.}
Evaluating a new model on MiraBench (both modalities) costs approximately \textbf{3--12 A100-GPU-hours} depending on model size, making the benchmark practical for routine evaluation without dedicated cluster access.

\section{Comparison with Existing Benchmarks}
\label{app:benchmark_comparison}

Table~\ref{tab:benchmark_comparison} provides a systematic feature comparison between MiraBench and representative existing evaluation frameworks for video generation and world models.

\begin{table*}[h]
  \caption{
    Feature comparison of MiraBench with existing benchmarks.
    \cmark: supported; \xmark: not supported; \textcolor{gray}{$\circ$}: partially supported.
  }
  \label{tab:benchmark_comparison}
  \centering\small
  \setlength{\tabcolsep}{3.5pt}
  \begin{tabular}{lccccccccc}
    \toprule
    Feature
      & \rotatebox{70}{MiraBench}
      & \rotatebox{70}{WorldArena}
      & \rotatebox{70}{WorldSimBench}
      & \rotatebox{70}{WorldModelBench}
      & \rotatebox{70}{WorldScore}
      & \rotatebox{70}{VBench}
      & \rotatebox{70}{EvalCrafter}
      & \rotatebox{70}{PhyGenBench}
      & \rotatebox{70}{Physion++} \\
    \midrule
    \multicolumn{10}{l}{\textit{Evaluation Target}} \\
    \quad Robot manipulation focus      & \cmark & \cmark & \textcolor{gray}{$\circ$} & \textcolor{gray}{$\circ$} & \xmark & \xmark & \xmark & \xmark & \xmark \\
    \quad Vector-conditioned models     & \cmark & \cmark & \cmark & \textcolor{gray}{$\circ$} & \xmark & \xmark & \xmark & \xmark & \xmark \\
    \quad Instruction-conditioned models       & \cmark & \xmark & \xmark & \cmark & \cmark & \cmark & \cmark & \cmark & \xmark \\
    \quad Multiple embodiments          & \cmark & \cmark & \textcolor{gray}{$\circ$} & \xmark & \xmark & \xmark & \xmark & \xmark & \xmark \\
    \midrule
    \multicolumn{10}{l}{\textit{Physics Evaluation}} \\
    \quad Physical consistency          & \cmark & \textcolor{gray}{$\circ$} & \textcolor{gray}{$\circ$} & \cmark & \textcolor{gray}{$\circ$} & \textcolor{gray}{$\circ$} & \xmark & \cmark & \cmark \\
    \quad Quantitative physics law      & \cmark & \xmark & \xmark & \xmark & \xmark & \xmark & \xmark & \textcolor{gray}{$\circ$} & \cmark \\
    \quad Reference-free assessment     & \cmark & \xmark & \xmark & \xmark & \xmark & \xmark & \xmark & \textcolor{gray}{$\circ$} & \cmark \\
    \midrule
    \multicolumn{10}{l}{\textit{Action Faithfulness}} \\
    \quad Action-following fidelity     & \cmark & \textcolor{gray}{$\circ$} & \textcolor{gray}{$\circ$} & \xmark & \xmark & \xmark & \xmark & \xmark & \xmark \\
    \quad Failure-regime testing         & \cmark & \xmark & \xmark & \xmark & \xmark & \xmark & \xmark & \xmark & \xmark \\
    \quad Implicit failure perturbations& \cmark & \xmark & \xmark & \xmark & \xmark & \xmark & \xmark & \xmark & \xmark \\
    \quad Optimism bias detection       & \cmark & \xmark & \xmark & \xmark & \xmark & \xmark & \xmark & \xmark & \xmark \\
    \midrule
    \multicolumn{10}{l}{\textit{Evaluation Methodology}} \\
    \quad Dual action modality          & \cmark & \xmark & \xmark & \xmark & \xmark & \xmark & \xmark & \xmark & \xmark \\
    \quad Fine-tuned VLM evaluator      & \cmark & \xmark & \xmark & \xmark & \xmark & \xmark & \xmark & \xmark & \xmark \\
    \quad Human annotation validation   & \cmark & \textcolor{gray}{$\circ$} & \textcolor{gray}{$\circ$} & \cmark & \textcolor{gray}{$\circ$} & \cmark & \cmark & \xmark & \cmark \\
    \quad Hierarchical / diagnostic     & \cmark & \xmark & \xmark & \xmark & \xmark & \textcolor{gray}{$\circ$} & \xmark & \xmark & \xmark \\
    \bottomrule
  \end{tabular}
\end{table*}

\paragraph{Key differentiators.}
Three capabilities are unique to MiraBench among existing benchmarks:

\begin{itemize}[leftmargin=*]
\item \textbf{Failure-regime evaluation.}
  All existing benchmarks evaluate world models in the normal operating regime, measuring average-case performance on successful demonstrations.
  MiraBench is the first to systematically probe the failure regime through implicit perturbations, testing whether models correctly propagate failure-inducing action signals rather than defaulting to their success prior.

\item \textbf{Dual action modality.}
  MiraBench evaluates each level under both precise (motor command) and descriptive (natural language) action conditioning, enabling diagnosis of whether failures originate at the motor-encoding level or the semantic level.
  No existing benchmark tests both modalities within a unified framework.

\item \textbf{Nested diagnostic structure.}
  Rather than producing a single aggregate score, MiraBench's three-level hierarchy (physics adherence $\rightarrow$ action following $\rightarrow$ optimism bias) identifies \textit{where} a model fails, not just \textit{how much}.
  A model failing at Level~1 needs better physics; a model passing Level~1 but failing Level~2 needs better action conditioning; a model passing both but failing Level~3 selectively ignores failure signals.
  This diagnostic interpretability is absent from all existing frameworks.
\end{itemize}


\section{Broader Impact Statement}
\label{sec:broader_impact}

\subsection{Positive impacts.}
MiraBench addresses a critical gap in the evaluation of world models for embodied AI.
By shifting evaluation from visual quality toward action-following fidelity and failure prediction, we aim to prevent a class of silent deployment failures: world models that generate visually compelling but physically incorrect training data for robot policies.
Policies trained on such data may behave unpredictably in safety-critical scenarios, as they have never been exposed to realistic failure modes during training.
Our benchmark provides the diagnostic tools to identify this risk before deployment, potentially reducing real-world accidents in robotic manipulation systems.

The three-level diagnostic structure enables targeted model improvement rather than undirected scaling, potentially reducing the computational resources wasted on approaches that improve visual quality without improving simulation fidelity.
By releasing our evaluation framework, perturbation taxonomy, and annotated datasets, we lower the barrier for the community to audit world models before using them as data generators.

\subsection{Potential negative impacts.}

\paragraph{Environmental impact.}
The benchmark evaluation pipeline requires substantial GPU computation for video generation (approximately 8 A100-hours per model for the full test set).
However, this cost is incurred once per model evaluation and is modest compared to the training cost of the world models themselves.
The VLM evaluator inference adds minimal overhead (under 1 GPU-hour for the full benchmark).
We provide efficiency guidelines and support for partial evaluation on reduced test sets to accommodate resource-constrained settings.

\paragraph{Impact on labor.}
Human annotation is used for evaluator calibration and validation rather than as the primary evaluation mechanism.
All annotators are compensated at rates above local minimum wage, work under non-coercive conditions, and are informed about the research purpose of the annotation task.
The benchmark is designed to minimize ongoing human annotation requirements through the distilled VLM evaluator, reducing long-term dependence on human labor for evaluation at scale.

\paragraph{Dual-use considerations.}
World models that faithfully simulate failure could, in principle, be used to generate adversarial training data or to identify vulnerabilities in robotic systems.
We believe the benefits of transparent evaluation substantially outweigh this risk: understanding where world models fail is a prerequisite for making them safe, and concealing these failure modes would not prevent their exploitation by motivated actors with access to the same models.

\section{Ethical Considerations}
\label{app:ethics}

\paragraph{Human annotation.}
All human annotators were recruited through a professional data annotation platform and compensated at rates exceeding local minimum wage standards.
Annotators were informed of the research purpose prior to participation, provided written consent, and were free to withdraw at any time without penalty.
The annotation task involves watching robot manipulation videos recorded in controlled laboratory environments; no personally identifiable information, violent content, or otherwise distressing material is present in any video.
Annotator demographics are not collected beyond professional qualification verification, in compliance with data minimization principles. Institutional Review Board (IRB) approvals were obtained.

\paragraph{Data privacy.}
All videos in MiraBench are recorded in laboratory settings with robotic manipulators operating on inanimate objects (fruits, containers, cloths).
No human subjects, faces, or private spaces appear in any frame.
The GR-1 and Lingchu datasets used as source material are collected under their respective institutional protocols with appropriate permissions for research use.

\paragraph{Consent and licensing.}
We confirm compliance with the terms of use for all datasets employed:
GR-1 data is used under NVIDIA's research license;
DROID data is used under its open research license;
Lingchu data is used with institutional permission.
MiraBench evaluation code and annotations will be released under a permissive open-source license (Apache-2.0) to maximize community benefit while maintaining clear attribution requirements.

\paragraph{Environmental considerations.}
The full MiraBench evaluation pipeline requires approximately 8 A100-GPU-hours per model for video generation and under 1 GPU-hour for VLM evaluation.
We provide configuration options for reduced evaluation (subset of tasks or episodes) to accommodate researchers with limited computational resources, and we release pre-generated evaluation videos for all models reported in this paper to eliminate redundant computation.

\paragraph{Intended use and misuse prevention.}
MiraBench is designed for evaluating and improving world models in research settings.
We discourage its use as the sole criterion for deploying world models in safety-critical applications without additional real-world validation.
The implicit failure perturbation taxonomy is published for transparency and reproducibility; while it could theoretically inform adversarial attacks on world-model-based systems, we believe open evaluation standards are essential for building trustworthy embodied AI and that security-through-obscurity is not a viable long-term strategy.



\end{document}